\documentclass[sigconf]{acmart}

\AtBeginDocument{%
  \providecommand\BibTeX{{%
    \normalfont B\kern-0.5em{\scshape i\kern-0.25em b}\kern-0.8em\TeX}}}

\copyrightyear{2024}
\acmYear{2024}
\setcopyright{acmlicensed}\acmConference[KDD '24]{Proceedings of the 30th ACM SIGKDD Conference on Knowledge Discovery and Data Mining}{August 25--29, 2024}{Barcelona, Spain}
\acmBooktitle{Proceedings of the 30th ACM SIGKDD Conference on Knowledge Discovery and Data Mining (KDD '24), August 25--29, 2024, Barcelona, Spain}
\acmDOI{10.1145/3637528.3671844}
\acmISBN{979-8-4007-0490-1/24/08}

\usepackage[utf8]{inputenc} 
\usepackage[T1]{fontenc}    
\usepackage{hyperref}       
\usepackage{amsfonts}       

\usepackage{enumitem}
\usepackage{amsmath}
\usepackage{algorithm}
\usepackage{algorithmicx}
\usepackage[noend]{algpseudocode}
\usepackage{makecell}
\usepackage{multirow}
\usepackage{multicol}
\usepackage{array}
\usepackage{subfigure}
\usepackage{capt-of}
\usepackage{wrapfig}
\usepackage{tikz}

\newcommand*\circled[1]{\tikz[baseline=(char.base)]{
            \node[shape=circle,draw,inner sep=0.5pt] (char) {#1};}}

\usepackage{cleveref}
\usepackage{caption}
\usepackage{stfloats}
\usepackage{float}
\usepackage[normalem]{ulem}
\useunder{\uline}{\ul}{}
\usepackage{balance}

\begin{document}

\title{MSPipe: Efficient Temporal GNN Training via\\ Staleness-Aware Pipeline
}


\author{Guangming Sheng}
\orcid{}
\affiliation{%
  \institution{The University of Hong Kong}
  \city{Hong Kong}
  \country{China}}
\email{gmsheng@connect.hku.hk}

\author{Junwei Su}
\authornote{Corresponding author}
\orcid{}
\affiliation{%
  \institution{The University of Hong Kong}
  \city{Hong Kong}
  \country{China}}
\email{junweisu@connect.hku.hk}

\author{Chao Huang}
\orcid{}
\affiliation{%
  \institution{The University of Hong Kong}
  \city{Hong Kong}
  \country{China}}
\email{chaohuang75@gmail.com}

\author{Chuan Wu}
\orcid{}
\affiliation{%
  \institution{The University of Hong Kong}
  \city{Hong Kong}
  \country{China}}
\email{cwu@cs.hku.hk}

\renewcommand{\shortauthors}{Guangming Sheng, Junwei Su, Chao Huang, and Chuan Wu}





\renewcommand{\shorttitle}{MSPipe: Efficient Temporal GNN Training via Staleness-Aware Pipeline}
\begin{abstract}
Memory-based Temporal Graph Neural Networks (MTGNNs) are a class of temporal graph neural networks that utilize a node memory module to capture and retain long-term temporal dependencies, leading to superior performance compared to memory-less counterparts. However, the iterative reading and updating process of the memory module in MTGNNs to obtain up-to-date information needs to follow the temporal dependencies. This introduces significant overhead and limits training throughput. Existing optimizations for static GNNs are not directly applicable to MTGNNs due to differences in training paradigm, model architecture, and the absence of a memory module. Moreover, these optimizations do not effectively address the challenges posed by temporal dependencies, making them ineffective for MTGNN training. In this paper, we propose MSPipe, a general and efficient framework for memory-based TGNNs that maximizes training throughput while maintaining model accuracy. Our design specifically addresses the unique challenges associated with fetching and updating node memory states in MTGNNs by integrating staleness into the memory module. However, simply introducing a predefined staleness bound in the memory module to break temporal dependencies may lead to suboptimal performance and lack of generalizability across different models and datasets. To overcome this, we introduce an online pipeline scheduling algorithm in MSPipe that strategically breaks temporal dependencies with minimal staleness and delays memory fetching to obtain fresher memory states. This is achieved without stalling the MTGNN training stage or causing resource contention. Additionally, we design a staleness mitigation mechanism to enhance training convergence and model accuracy. Furthermore, we provide convergence analysis and demonstrate that MSPipe maintains the same convergence rate as vanilla sampling-based GNN training. Experimental results show that MSPipe achieves up to 2.45$\times$ speed-up without sacrificing accuracy, making it a promising solution for efficient MTGNN training. The implementation of our paper can be found at the following link: \href{https://github.com/PeterSH6/MSPipe}{https://github.com/PeterSH6/MSPipe}.


\end{abstract}

\maketitle

\section{Introduction}

Many real-world graphs exhibit dynamic characteristics, with nodes and edges continuously evolving over time,
such as temporal social networks~\cite{rozenshtein2019mining, wei2015measuring} and temporal user-item graphs in recommendation systems~\cite{ma2020temporal, ye2020time}.
Previous attempts to model such dynamic systems have relied on static graph representations, which overlook their temporal nature~\cite{zhang2019ige+, zhang2020learning, nguyen2018continuous,su2022structure,su2024bg}. Recently, temporal graph neural networks (TGNNs) have been developed to address this limitation. TGNNs are designed to incorporate time-aware information, learning both structural and temporal dependencies. Consequently, TGNNs facilitate more accurate and comprehensive modeling of dynamic graphs~\cite{rossi2021tgn, wang2021apan, kumar2019jodie, su2023towardsER, trivedi2019dyrep, xu2020tgat,zhang2023tiger, cong2023we, sankar2020dysat,su2024pres,su2023towards}.

\begin{figure*}[htbp]
    \centering
    \includegraphics[width=0.77\linewidth]{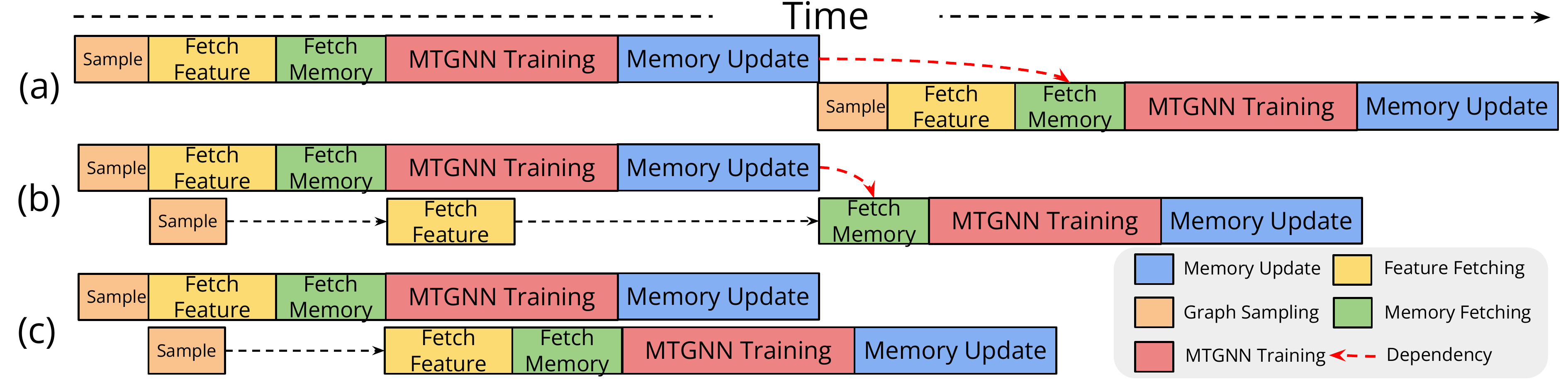}
    \vspace{-2mm}
    \caption{Memory-based TGNN training. (a) represents the general training scheme; (b) shows the pre-sampling and pre-fetching optimization;
     (c) is the case of breaking the temporal dependency, where the TGNN training stage is executed uninterruptedly.}
    \label{fig:pipeline}
    \vspace{-3mm}
\end{figure*}

Among the existing TGNN models, MTGNNs like TGN~\cite{rossi2021tgn}, APAN~\cite{wang2021apan}, JODIE~\cite{kumar2019jodie}, and TIGER~\cite{zhang2023tiger} have achieved state-of-the-art performance on various tasks, notably link prediction and node classification~\cite{dgb_neurips_D&B_2022}. Their success can be attributed to the node memory module, which stores time-aware representations, enabling the capture of intricate long-term information for each node. The training process of MTGNNs involves the following steps:
First, node memory states and node/edge features from sampled subgraphs are loaded and inputted into the MTGNN model. In the model, the \textit{message} module sequentially processes incoming events to generate message vectors. Subsequently, the \textit{memory} module utilizes these message vectors along with the previous memory states to generate new memory vectors. Then, the \textit{embedding} module combines the latest memory vectors with structural information to generate temporal embeddings for the vertices. At the end of each iteration, the updated memory states are written back to the memory module storage in the CPU's main memory, as illustrated in Figure~\ref{fig:pipeline}. \\\vspace{-0.12in}

\noindent \textbf{Significant Overhead of The Memory Module in TGNNs}.
Despite their impressive performance, training memory-based TGNNs at scale remains challenging due to the \textit{temporal dependency} induced by the memory module. This temporal dependency arises from the memory fetch and update operations across different iterations. Specifically, the latest memory state of a node cannot be fetched until the update of the node memory module in the previous iteration is completed. This dependency is illustrated by the red arrow in Figure~\ref{fig:pipeline}, indicating that subsequent iterations rely on the most recently updated node memory from previous iterations. The memory module functions as a recursive filtering mechanism, continually distilling and incorporating information from historical events into the memory states. Respecting this temporal dependency incurs significant overhead in memory-based TGNN training, accounting for approximately 36.1\% to 58.6\% of the execution time of one training iteration, depending on the specific models. However, preserving this temporal dependency is essential for maintaining the model's performance. Therefore, it's imperative to enhance the training throughput while effectively modeling the temporal dependency without compromising the model's accuracy. \\\vspace{-0.12in}

\noindent \textbf{Limitation of Static GNN Optimizations.}
There is a line of research~\cite{wan2022pipegcn, peng2022sancus, zheng2022bytegnn, kaler2022accelerating, gandhi2021p3} focused on optimizing the training of static GNNs. However, the temporal dependencies specific to MTGNNs, arising from the memory module, pose unique challenges. As a result, these works are inadequate for handling such temporal dependencies and are ineffective for MTGNN training. For instance, when applying pre-sample and pre-fetch optimizations from ByteGNN~\cite{zheng2022bytegnn} and SAILENT~\cite{kaler2022accelerating}, the memory fetching in the next training iteration must wait until the memory update in the current iteration is completed, as shown in Figure~\ref{fig:pipeline}(b). This waiting period diminishes training efficiency. Moreover, approaches like PipeGCN~\cite{wan2022pipegcn} and SAILENT~\cite{peng2022sancus} address the substantial communication overhead caused by inter-layer dependencies in multi-layer GNNs using the full graph training paradigm. However, these approaches may not be applicable to MTGNNs, which typically utilize a single layer and employ sample-based subgraph training. Hence, there is an urgent need for a general parallel execution framework enabling more efficient and scalable distributed MTGNN training.

To address these gaps, we introduce MSPipe, a general and efficient training system for memory-based TGNNs. MSPipe leverages a minimal staleness bound to accelerate MTGNN training while ensuring model convergence with theoretical guarantees. \\\vspace{-0.12in}

\noindent \textbf{Training Pipeline Formulation.}
To identify the bottlenecks in MTGNN training, we present a formulation for the MTGNN training pipeline. Through an analysis of initiation and completion times across various training stages, we decompose MTGNN training into distinct stages. This formulation enables a comprehensive analysis of training bottlenecks. Leveraging this formulation, we conduct a thorough profiling of distributed MTGNN training. Our analysis highlights the potential for optimizing the bottlenecks arising from the memory module and its temporal dependencies. \\\vspace{-0.12in}

\noindent \textbf{Tackling the Temporal Dependencies.} We propose two key designs to enhance training throughput while preserving model accuracy. \textbf{(1)} We break the temporal dependencies by introducing staleness in the memory module, as shown in Figure~\ref{fig:pipeline}(c). However, determining an appropriate staleness bound requires careful tuning and lacks generalizability across diverse MTGNN models and datasets. Setting a small bound may hinder system throughput, while a large bound can introduce errors in model training. To overcome this challenge, we design a minimal staleness algorithm that determines a precise staleness bound and effectively schedules the training pipeline accordingly. The resulting minimal staleness bound ensures uninterrupted execution of MTGNN training stages. Moreover, it allows for the retrieval of the node memory vectors that are as fresh as possible, effectively minimizing staleness errors. \textbf{(2)} To further improve the convergence of MSPipe, we propose a lightweight staleness mitigation method that leverages the node memory vectors of recently updated nodes with the highest similarity, which effectively reduces the staleness error. \\\vspace{-0.12in}

\noindent \textbf{Theoretical guarantees.} Although previous works have analyzed the convergence rate of static GNN training~\cite{chen2017stochastic, cong2020minimal, cong2021importance}, the consequences of violating temporal dependencies have not yet been explored. Therefore, we present an in-depth convergence analysis for our proposed methods, validating their effectiveness.

In summary, we make the following contributions in this paper:

$\bullet$ We propose a general formulation for the MTGNN training pipeline, allowing us to identify training bottlenecks arising from the memory module. Based on the formulation, MSPipe strategically determines a minimal staleness bound to ensure uninterrupted MTGNN training while minimizing staleness error, thereby maximizing training throughput with high accuracy.

$\bullet$ We propose a lightweight similarity-based staleness mitigation strategy to further improve the model convergence and accuracy.

$\bullet$ We provide a theoretical convergence analysis, demonstrating that MSPipe does not sacrifice convergence speed. The convergence rate of our method is the same as vanilla MTGNN training (without staleness).


$\bullet$ We evaluate the performance of MSPipe through extensive experiments. Our results demonstrate that MSPipe outperforms state-of-the-art MTGNN training frameworks, achieving up to 2.45$\times$ speed-up and 83.6\% scaling efficiency without accuracy loss.
\begin{figure}[t]
    \includegraphics[width=\linewidth]{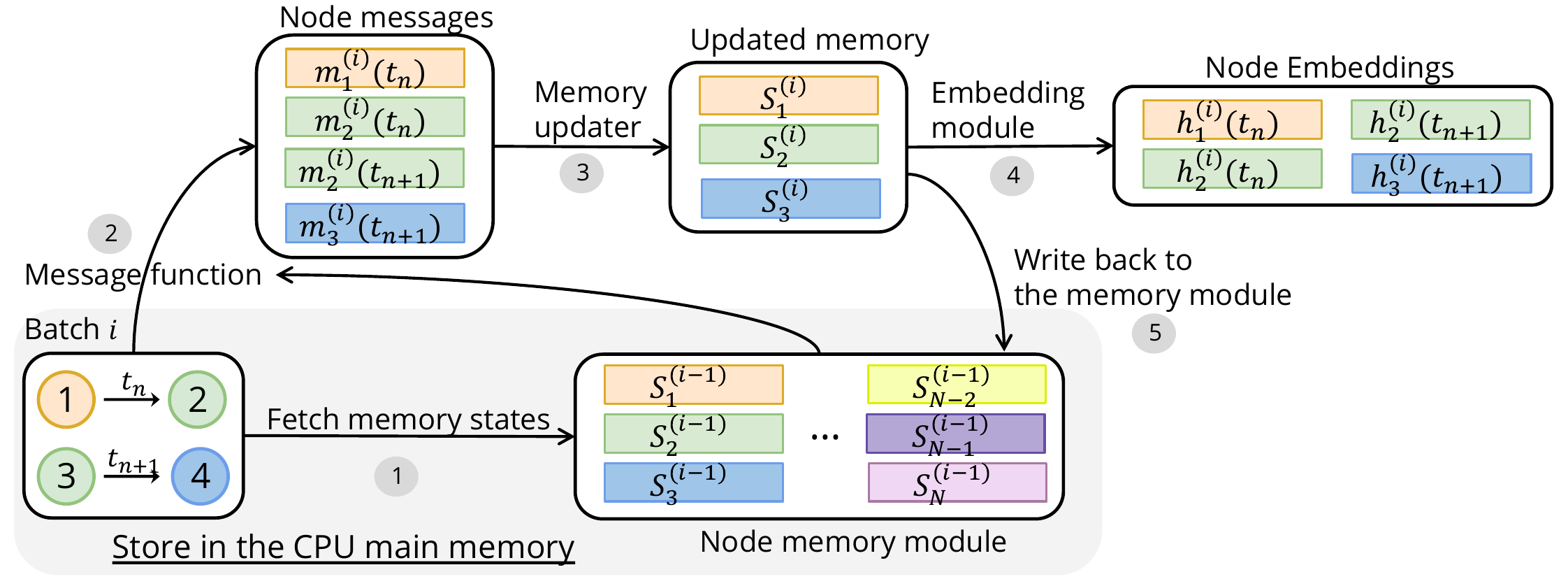}
    \vspace{-6mm}
    \caption{Memory-based TGNN Training Stages. The node memory states are stored in the CPU memory to ensure consistency among multiple training workers and reduce GPU memory contention. The MTGNN model is stored in the GPU.} 
    \vspace{-4mm}
    \label{fig:tgn}
\end{figure}
\section{Preliminary} \label{sec:preliminary}
\textbf{Dynamic Graphs}. We focus on event-based representation for dynamic graphs. A dynamic graph can be represented as $\mathcal G = (\mathcal V, \mathcal E)$, where $\mathcal V = \{1,..., N\}$ is the node set and $\mathcal E = \{y_{uv}(t)\}, u, v \in \mathcal V$  is the event sets~\cite{skarding2021foundations, rossi2021tgn, xu2020tgat}. The event set $\mathcal E$ represents a sequence of graph events $y_{uv}(t)$, indicating interactions between nodes $u$ and $v$ at timestamp $t \geq 0$.

\noindent \textbf{Temporal Graph Neural Network.}
Among the variety of TGNNs, memory-based TGNNs achieve superior accuracy in modeling temporal dynamics in graph-structured data~\cite{rossi2021tgn, wang2021apan, kumar2019jodie, dgb_neurips_D&B_2022, zhang2023tiger}. Memory-based TGNNs maintain a node memory vector $s_v$ for each node $v$ in a dynamic graph that memorizes long-term dependencies. The
memory update and training paradigms can be formulated as:
\begin{align}\label{eq:tgnn}
m^{(i)}_{v} &= \mathop{msg}\Big (s^{(i-1)}_{v}, s^{(i-1)}_{u}, y_{uv}(t), \Delta t \Big ) \notag \\
s^{(i)}_{v} &= \mathop{mem}\Big (s^{(i-1)}_{v}, m^{(i)}_{v} \Big )  \\
h^{(i)}_v &= \mathop{emb}\Big (s^{(i)}_{v}, s^{(i)}_{u}\ |\ u \in \mathcal{N}(v)\Big ) \notag
\end{align}
\noindent where $m^{(i)}_{v}$ represents a message generated by the graph event related to $v$ that occurs at training iteration $i$, $s^{(i)}_{v}$ is the memory states and $h^{(i)}_v$ is the embedding of node $v$ in iteration $i$ and $\Delta$t represents the time gap between the last updated time of the memory state $s^{(i-1)}_{v}$ of node $v$ and the occurrence time of the current graph event $y_{uv}(t)$. $\mathcal{N}(v)$ is the 1-hop temporal neighbours of nodes $v$. The message module $\mathop{msg}$ (e.g., MLP), memory update module $\mathop{mem}$ (e.g., RNN), and embedding module $\mathop{emb}$ (e.g., a single layer GAT) are all learnable components.
Note that all the operations described above collectively form the \textit{MTGNN training stage}, which is executed on the GPU. The updated memory vectors $s^{(i)}_{v}$ will be written back to the node memory storage in the CPU main memory. The detailed training workflow is illustrated in Figure~\ref{fig:tgn}.

\section{MSPipe framework}
We introduce MSPipe, a stall-free minimal-staleness scheduling system designed for MTGNN training (Figure~\ref{fig:pipeline}(c)). Our approach identifies the memory module as the bottleneck and leverages pipelining techniques across multiple iterations to accelerate training. We determine the minimal number of staleness iterations necessary to prevent pipeline stalling while ensuring the retrieval of the most up-to-date memory states.
However, incorporating the minimal staleness bound into the training pipeline introduces resource competition due to parallel execution. To mitigate this, we present a resource-aware online scheduling algorithm that controls the staleness bound and alleviates resource contention. Additionally, we propose a lightweight similarity-based memory update mechanism to further mitigate staleness errors and obtain fresher information.

\renewcommand\arraystretch{0.5}
\begin{table}[t]
    \caption{Training time breakdown of TGN model. 
    }
    \vspace{-3mm}
    \centering
    \resizebox{\linewidth}{!}{
    \begin{tabular}{llllll}
    \toprule
        \textbf{Dataset} & \textbf{Sample} & \textbf{\makecell{Fetch\\feature}} & \textbf{\makecell{Fetch\\memory}} & \textbf{\makecell{Train\\MTGNN}} & \textbf{\makecell{Update\\memory}} \\ 
    \midrule
        REDDIT~\cite{kumar2019jodie} & 9.5\% & 12.6\% & 5.7\% & 46.9\% & 25.3\%  \\
        WIKI~\cite{kumar2019jodie} & 6.6\% & 5.8\% & 5.8\% & 51.5\% & 30.3\%  \\ 
        MOOC~\cite{kumar2019jodie} & 9.7\% & 3.0\% & 2.5\% & 53.1\% & 31.7\%  \\ 
        LASTFM~\cite{kumar2019jodie} & 11.5\% & 9.1\% & 8.5\% & 43.0\% & 26.8\%  \\ 
        GDELT~\cite{zhou2022tgl} & 17.6\% & 12.8\% & 10.5\% & 37.5\% & 21.6\% \\ 
    \bottomrule
    \end{tabular}}
    \label{tab:breakdown}
\end{table}

\subsection{MSPipe mechanism} \label{sec:3_1}
\textbf{Significant memory operations overhead.} 
We consider a 5-stage abstraction of memory-based TGNN training, i.e., graph sampling, feature fetching, memory fetching, MTGNN training, and memory update. 
We conduct detailed profiling of the execution time of each stage, with time breakdown shown in Table \ref{tab:breakdown}.
Memory operations incur substantial overhead ranging from 36.1\% to 58.6\% of one iteration training time for different MTGNN models, while sampling and feature fetching do not, due to the 1-layer MTGNN structure.
In Figure \ref{fig:pipeline}(b), memory fetching depends on memory vectors updated at the end of the last iteration, and 
has to wait for the relatively long MTGNN training and memory updating to finish

\noindent \textbf{Pipline mechanism.} 
A natural design to accelerate the training process involves decoupling the temporal dependency between the memory update stage in one training iteration and the memory fetching stage in the subsequent iteration, by leveraging stale memory vectors in the latter. Figure~\ref{fig:pipeline}(c) provides an overview of the training pipeline, where computation (e.g., MTGNN training) is parallelized with fragmented I/O operations including feature fetching, memory fetching, and memory update. The advanced memory fetching stage introduces a certain degree of staleness to the node memory module, causing the MTGNN model to receive outdated input. Mathematically, MSPipe's training can be formulated as follows (modifications from Eqn.~\ref{eq:tgnn} are highlighted in blue):
\begin{align}\label{eq:stale_tgnn}
m^{(i)}_{v} &= \mathop{msg} \Big ({\color{blue}\tilde{s}^{(i-k)}_{v}}, {\color{blue}\tilde{s}^{(i-k)}_{u}}, y_{uv}(t), \Delta t \Big ) \notag \\ 
{\color{blue}\tilde{s}^{(i)}_v} &= \mathop{mem} \Big ({\color{blue}\tilde{s}^{(i-k)}_{v}}, m^{(i)}_{v} \Big)  \\
h^{(i)}_v &= \mathop{emb} \Big ({\color{blue}\tilde{s}^{(i)}_{v}, \ \tilde{s}^{(i)}_{u}}\ |\ u \in \mathcal{N}(v) \Big)  \notag
\end{align}

\noindent where $\tilde{s}^{(i)}_v$ represents the memory vector of node $v$ in training iteration $i$ updated based on stale memory vector in iteration $i-k$, and $h^{(i)}_v$ is the embedding of node $v$.
MSPipe uses the memory vector from $k$ iterations before the current iteration to generate messages and train the model. 

\begin{figure}[t]
    \centering
    \includegraphics[width=0.96\linewidth]{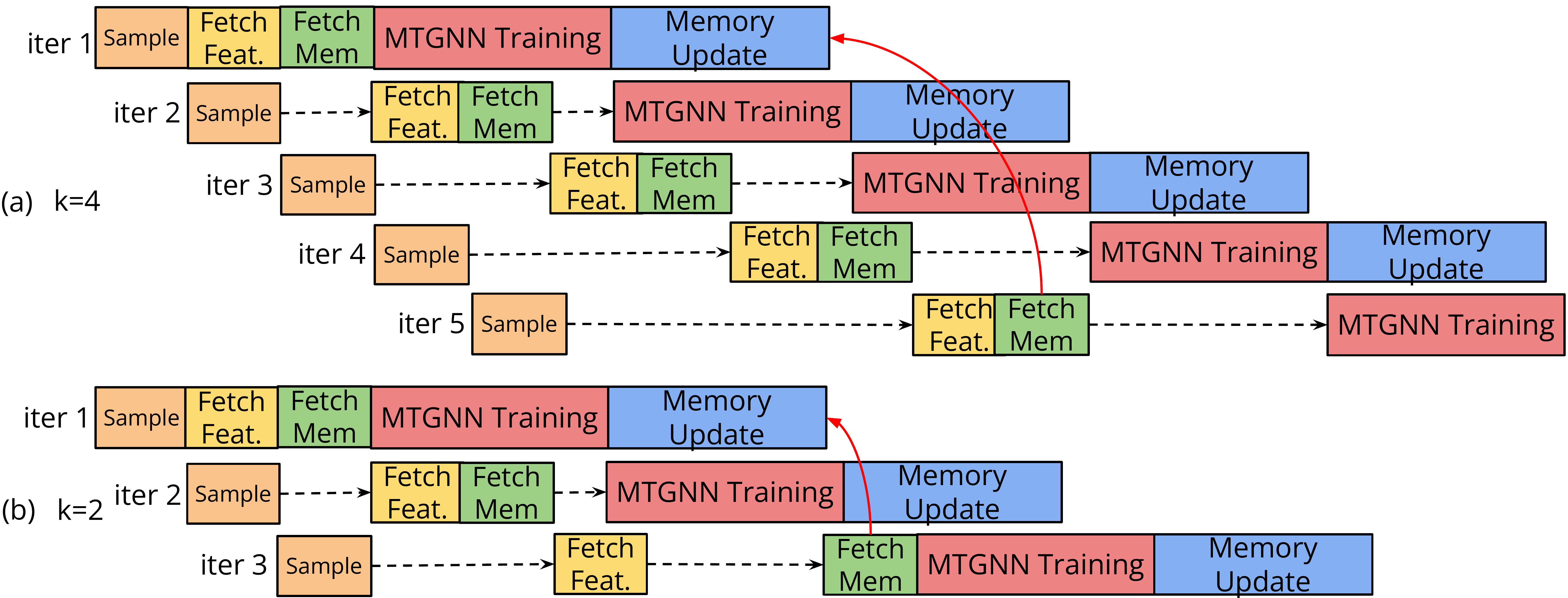}
    \caption{Pipeline execution. The dashed black arrow represents the bubble time. The red arrow denotes memory fetching to retrieve memory vectors updated $k$ iterations before.}
    \label{fig:k_stale}
\end{figure}

In the example pipeline in Figure~\ref{fig:k_stale}, we have staleness bound $k=2, 4$, indicating that MSPipe retrieves memory vectors updated two and four iterations before, respectively. Previous GNN frameworks~\cite{wan2022pipegcn, peng2022sancus} use a predefined staleness bound to address different dependencies.\textit{We argue that randomly selecting a staleness bound is inadequate. A small or large staleness bound may affect system performance or introduce errors in model training.} To support our argument, we conduct experiments on the LastFM dataset~\cite{kumar2019jodie}, training TGN model~\cite{rossi2021tgn}. As shown in Figure~\ref{fig:motivate_staleness}, applying the smallest staleness bound (e.g., $k=2$) leads to training throughput degradation, while employing a larger staleness bound (e.g., $k=4, 5$) impacts model accuracy. To address this, we introduce a pipeline scheduling policy that determines the minimal staleness bound that maximizes system throughput without affecting model convergence.

\subsection{Stall-free Minimal-staleness Pipeline}
\begin{figure}[t]
\begin{minipage}{.48\linewidth} 
\centering 
\includegraphics[width=.98\linewidth]{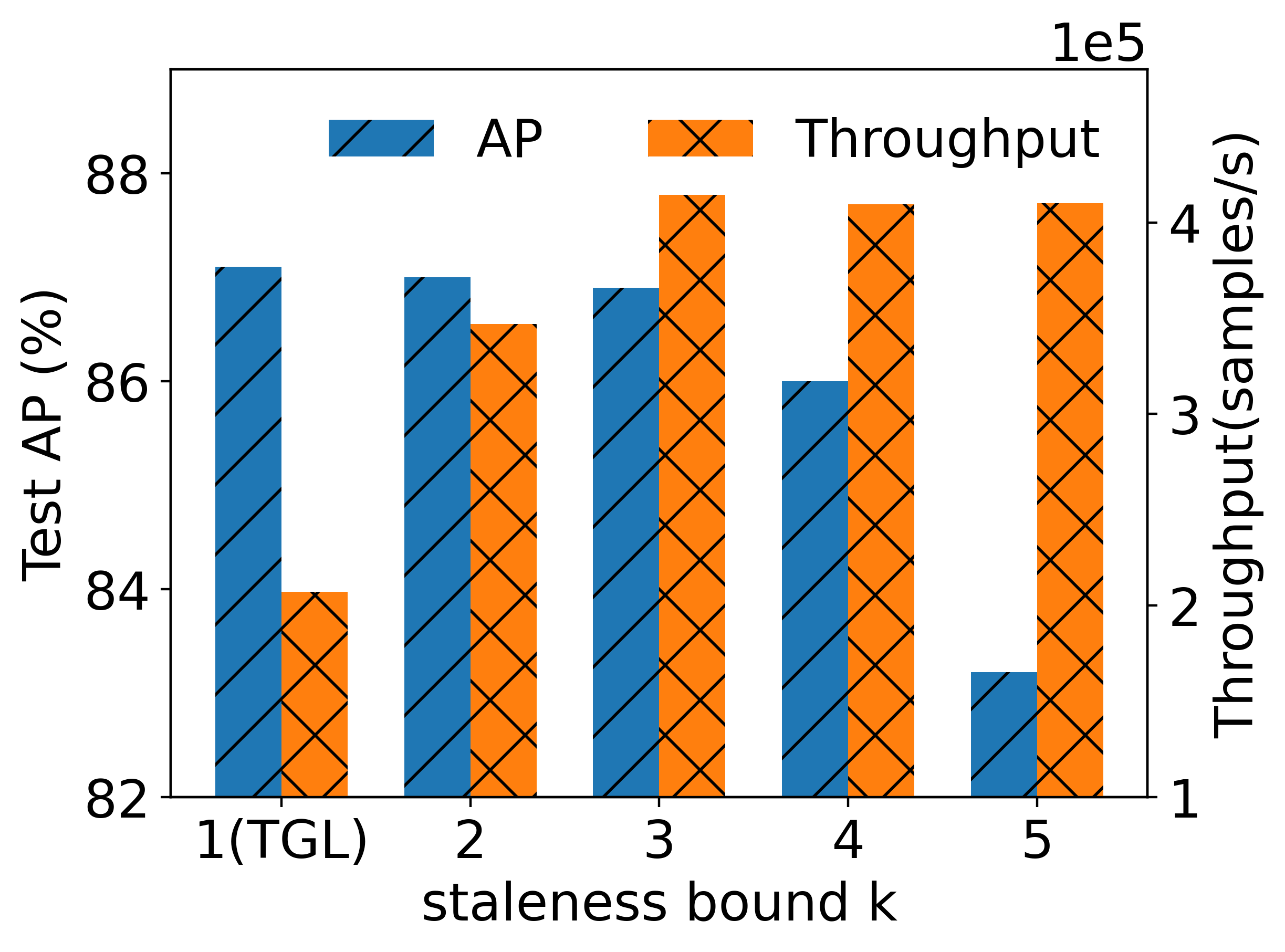}
\caption{Model accuracy and training throughput at different staleness bounds.} 
\label{fig:motivate_staleness}
\end{minipage}%
\hspace{2pt}
\begin{minipage}{.49\linewidth} 
\centering 
    \centering
    \includegraphics[width=\linewidth]{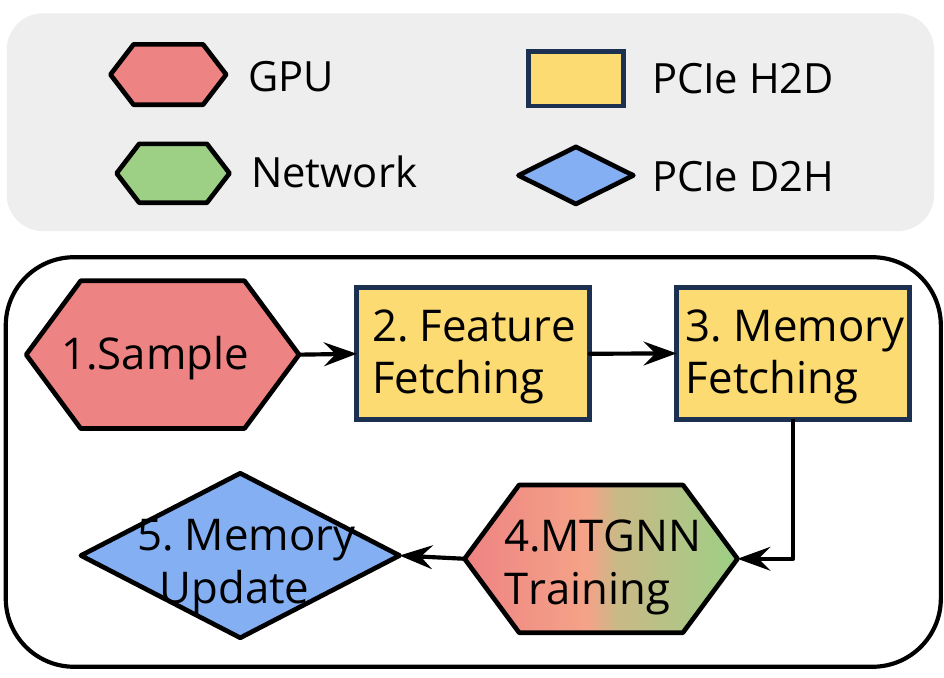}
\caption{Different resource requirements (by color/shape) of 5 training stages.} 
\label{fig:resource}
\end{minipage}
\end{figure}

To maximize MTGNN training throughput, our objective is to enable the GPU to seamlessly perform computation (i.e., MTGNN training stage) without waiting for data preparation, as depicted in Figure~\ref{fig:k_stale}(a). We seek to determine the minimal staleness bound $k$ and perform resource-aware online pipeline scheduling to avoid resource contention. This approach enables maximum speed-up without stalling the MTGNN training stage and ensures model convergence. 
To accurately model resource contention, we analyze the resource requirements of different stages. Figure~\ref{fig:resource} demonstrates that feature fetching and memory fetching contend for the copy engine and PCIe resources during the copy operation from host to device. However, no contention is encountered during the memory update stage, as it involves a copy operation from device to host~\cite{choquette2020nvidia}. Additionally, we adopt a GPU sampler with restricted GPU resource allocation to avoid competition with the MTGNN training stage.

\textbf{The start and end time modeling at different stages.} 
The problem of ensuring uninterrupted execution of the MTGNN training stage with minimal staleness can be transformed into determining the start time of each training stage. Therefore, it's essential to model the range of starting and ending times for different stages. Let $b_i^{(j)}$ and $e_i^{(j)}$ denote the start time and end time of stage $j$ in iteration $i$. The execution time of stage $j$, denoted as $\tau^{(j)}$, can be collected in a few iterations of profiling. The end time $e_i^{(j)}$ can be computed by adding execution time $\tau^{(j)}$ to the start time $b_i^{(j)}$, stated as $e_{i}^{(j)} = b_i^{(j)} + \tau^{(j)}$. There are three cases for computing $b_i^{(j)}$ to ensure sequential execution and avoid resource competition:

\noindent \textbf{1)} For the first stage, the sampler can initiate the sampling of a new batch immediately after the completion of the previous sample stage. This can be expressed as $b_i^{(1)} = e_{i-1}^{(1)} = b_{i-1}^{(1)} + \tau^{(1)}$.

\noindent \textbf{2)} In the second stage, feature fetching competes for PCIe and copy engine resources with memory fetching (stage 3) in the previous iteration. Hence, feature fetching cannot begin until both the memory fetching from the previous iteration and the sampling stage (stage 1) from the current iteration have been completed, as illustrated in Figure~\ref{fig:k_stale}. Consequently, the start time is determined as $b_i^{(2)} = \max{\Big \{e_i^{(1)}, e_{i-1}^{(3)}\Big \}} = \max{\Big \{b_i^{(1)} + \tau^{(1)}, b_{i-1}^{(3)} + \tau^{(3)} \Big \}}$. 

\noindent \textbf{3)} The remaining stages adhere to sequential execution order. Taking the MTGNN training stage as an example, it cannot commence until both the memory fetching from the current iteration and the same stage (i.e., MTGNN training) from the previous iteration have finished. The start time for these three stages are formulated as $b_i^{(j)}=\max{\Big \{e_i^{(j-1)}, e_{i-1}^{(j)}\Big \}} = \max{\Big \{b_i^{(j-1)} + \tau^{(j-1)}, b_{i-1}^{(j)} + \tau^{(j)} \Big \}}$.
By combining the above results, we obtain the following equations:
\vspace{-1mm}
\begin{align}
    \label{eq:start}
    \centering
    & b_i^{(j)} = \left\{
        \begin{aligned}
        & e_{i-1}^{(j)} &j = 1\\
        & \max{\Big \{ e_i^{(j-1)}, e_{i-1}^{(j+1)} \Big \}} & j = 2\\
        & \max{\Big \{ e_i^{(j-1)}, e_{i-1}^{(j)} \Big \}} &j \in [3, 5]\\
        \end{aligned}
        \right.& \\ \label{eq:end}
    & e_{i}^{(j)} = b_i^{(j)} + \tau^{(j)} \qquad \qquad \quad \ \ \, \, \, j \in [1, 5]
\end{align}

\textbf{Minimal-staleness bound.} 
Given the start and end time ranges of different stages, we observe a time gap between the start time of stage $b_i^{(j)}$ and the end time of the previous stage $e_i^{(j-1)}$, referring to the bubble time in Figure~\ref{fig:k_stale}. This motivates us to advance or delay the execution of a stage to obtain a fresher node memory state. To maximize training throughput with the least impact on model accuracy, our objective is to determine the minimal staleness bound $k_i$, ensuring that MSPipe fetches the most up-to-date memory vectors that are $k_i$ iterations prior to the current iteration $i$, without causing pipeline stalling. To tackle this optimization process, we must satisfy the following three constraints:

\begin{figure*}[t]
\begin{minipage}{.23\linewidth} 
\centering 
\includegraphics[width=0.9\linewidth]{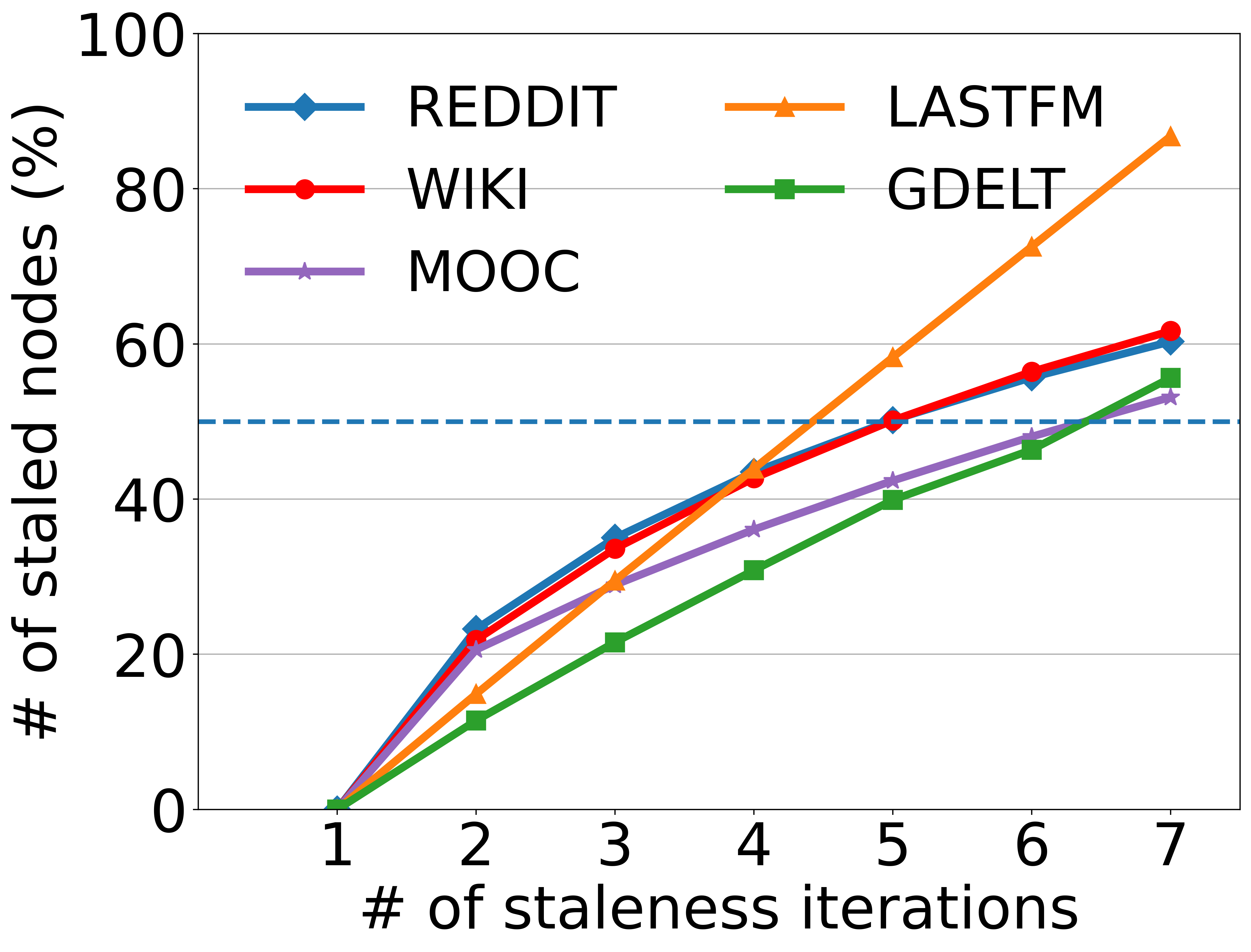}
\caption{Percentage of nodes that use staled memory vectors under different numbers of staleness iterations} 
\label{fig:overlap}
\end{minipage}%
\hspace{3pt}
\begin{minipage}{.5\linewidth} 
    \vspace{2mm}
    \centering
    \includegraphics[width=\linewidth]{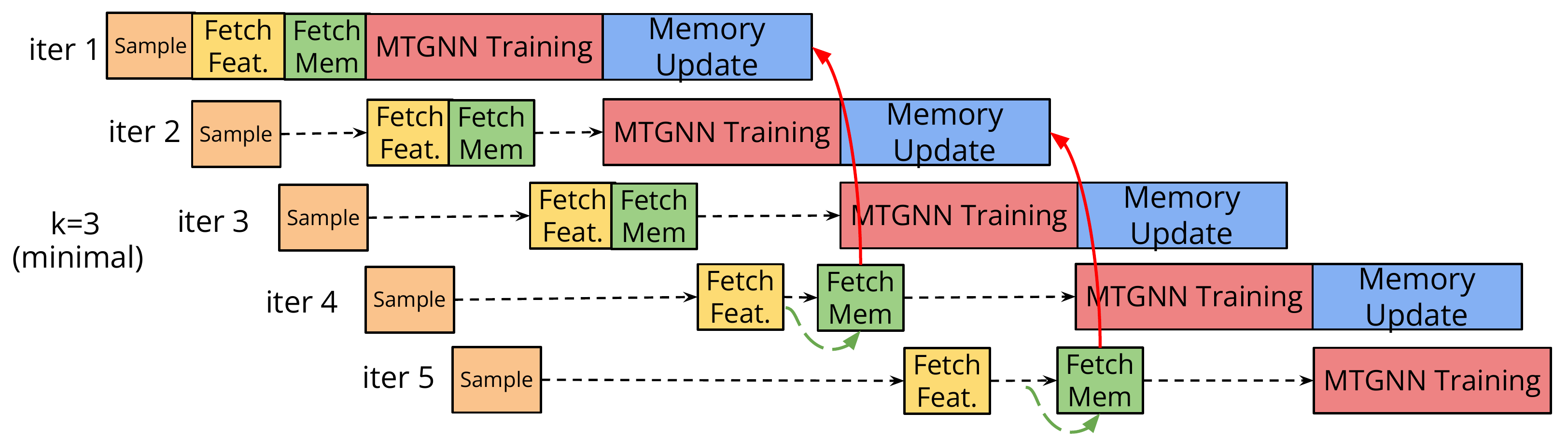}
    \vspace{-5mm}
    \caption{Resource-aware online schedule with minimal staleness bound is 3. The scheduler delays the memory fetch by utilizing the bubble time and avoids resource competence from different stages. The dashed green and black arrows represent the delay time and the bubble time respectively. The red arrow denotes fetching the memory states updated $k$ iterations before.}
    \label{fig:schedule}
\end{minipage}
\hspace{3pt}
\begin{minipage}{.23\linewidth} 
    \centering
    \includegraphics[width=0.9\linewidth]{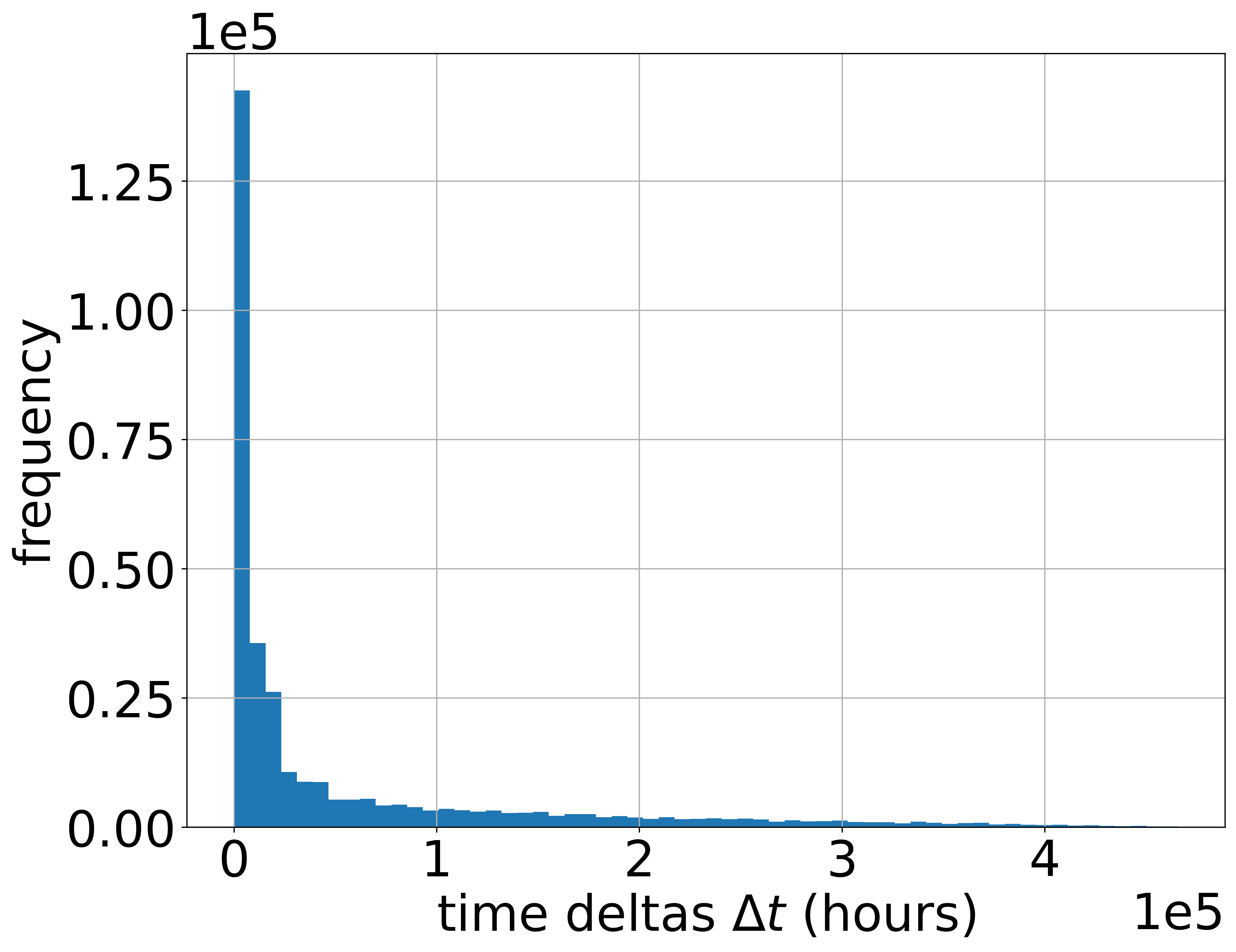}
    \caption{Distribution of $\Delta t$ in WIKI dataset. Other datasets follow a similar power-law distribution.
    }
    \label{fig:wiki}
\end{minipage}
\end{figure*}

\noindent \textbf{C1:} We ensure that memory updates for the $i-k_i$th iteration are completed before fetching memory states in the $i$th iteration, which can be expressed as $e_{i - k_i}^{(5)} \geq b_i^{(3)}$.

\noindent \textbf{C2:} To enable incessant execution of MTGNN training stages on the GPU, we should guarantee that delaying the memory fetching stage does not stall the subsequent MTGNN training stage. This condition can be formulated as $e_{i - k_i}^{(5)} \leq b_{i}^{(4)} - \tau^{(3)}$, where $b_{i}^{(4)} - \tau^{(3)}$ represents the delayed starting time of the memory fetching stage.

\noindent \textbf{C3:} We apply an upper bound $k_{\mathrm{max}}$ on the staleness bound based on a key observation: \textit{During each iteration, the memory module updates only a small subset of nodes' memory vectors.} Consequently, it is only the memory vectors of these specific nodes that become stale when they are fetched prior to the memory update stage. Figure~\ref{fig:overlap} demonstrates the increase in the percentage of stale nodes with larger staleness iterations. We select an upper bound $k_{\mathrm{max}}$ to ensure the percentage of stale nodes will not exceed 50\%.
Combining all above, we can formulate the following optimization problem:
\begin{gather*}
\begin{aligned}
     \text{minimize} & \quad k_i \\
     \text{subject to} & \quad e_{i - k_i}^{(5)} \geq b_i^{(3)}, \\
    & \quad e_{i - k_i}^{(5)} \leq b_{i}^{(4)} - \tau^{(3)}, \\
\end{aligned} \\
    1 \leq k_i < \min\{i, k_{\mathrm{max}}\}, i=1, \dots, E.
\end{gather*}
\noindent Here $E$ is the total number of iterations in an epoch. By iterating through each iteration, the above problem can be solved in $O(E)$.

\textbf{Resource-aware online pipeline schedule.} 
Once the minimal staleness iteration number $k_{i}$ has been determined, we can schedule the training pipeline by deciding the commencement time of each stage. This scheduling problem can be modeled as a variant of the ``bounded buffer problem'' in producer-consumer systems~\cite{mehmood2011implementation}.
Here, the buffer length corresponds to the number of staled iterations $k_i$, with the memory update stage acting as a slow consumer and the memory fetching stage as a fast producer. 
To ensure efficient training, the scheduler ensures that the training stages from different iterations do not compete for the same hardware resources and strictly adhere to a sequential execution order. By leveraging the minimal staleness iteration numbers $k_{i}$, the scheduler monitors the staleness state of each iteration and defers the memory fetching stage until the minimal staleness condition is satisfied, ensuring that subsequent MTGNN training stages are not impeded to maximize training throughput. This is achieved by effectively utilizing the bubble time to delay the memory fetching stage, as illustrated in Figure~\ref{fig:schedule}. The detailed pseudocode can be found in Appendix~\ref{sec:app_algo}.

\subsection{Similarity-based Staleness mitigation}
Nodes in the dynamic graph only update their memory states based on events directly involving them. Therefore, the nodes that are not involved in any graph events for a long duration will maintain stationary memory states, which would result in stale representations~\cite{rossi2021tgn, kumar2019jodie}. MSPipe may aggravate this problem although minimal staleness is introduced. To improve model convergence and accuracy with MSPipe, we further propose a staleness mitigation strategy by aggregating memory states of recently active nodes with the highest similarity, which are considered to have similar and fresher temporal representations, to update the stale memory of a node.  When node $v$'s memory has not been updated for time $\Delta t$, longer than a threshold $\gamma$, we update the stale memory of the node, $\tilde{s}^{(i-k_i)}_v$, by combining it with the averaged memory states of a set of most similar and active nodes $\Omega(v)$. An active node is defined to be the one whose memory is fresher than that of node $v$ and $\Delta t$ is smaller than $\gamma$. To measure the similarity between different nodes, we count their common neighbors which are reminiscent of the Jaccard similarity~\cite{leskovec2020mining}. We observe that $\Delta t$ follows a power-law distribution shown in Figure~\ref{fig:wiki}, which means that only a few $\Delta t$ values are much larger than the rest. We accordingly set $\gamma$ to $p$ quantile (e.g., 99\% quantile) of the $\Delta t$ distribution to reduce staleness errors. We apply the following memory staleness mitigation mechanism in the memory fetching stage:

\vspace{-2pt}
\begin{equation*}
    \hat{s}^{(i-k_i)}_{v} = \lambda \tilde{s}^{(i-k_i)}_{v}  + (1 - \lambda) \frac{\sum_{u \in \Omega(v)} \tilde{s}^{(i-k_i)}_{u}}{|\Omega(v)|} 
\end{equation*}
where $\hat{s}^{(i-k_i)}_{v}$ is the mitigated memory vector of node $v$ at iteration $i - k_i$, and $\lambda$ is a hyperparameter in $[0, 1]$. The mitigated memory vector will then be fed into the memory update function to generate new memory states for the node: 
\begin{equation*}
    \hat{s}^{(i)}_v = \mathop{mem}\Big (\hat{s}^{(i-k_i)}_{v}, m^{(i)}_{v} \Big )
 \end{equation*}

\section{Theoretical Analysis} \label{sec:theory}
We analyze the convergence guarantee and convergence rate of MSPipe with respect to our bounded node memory vector staleness. By carefully scheduling the pipeline and utilizing stale memory vectors, we demonstrate that our approach incurs negligible approximation errors that can be bounded. We provide a rigorous analysis of the convergence properties of our approach, which establishes the theoretical foundation for its effectiveness in practice.
\begin{theorem}[Convergent result, informal]\label{thm:convergent}
With a memory-based TGNN model, suppose that \textbf{1)} there is a bounded difference between the stale node memory vector $\Tilde{s}^{(i)}_{v}$ 
and the exact node memory vector $s^{(i)}_{v}$ with the staleness bound $\epsilon_s$, i.e., $\big \Vert \Tilde{s}^{(i)}_{v} - s^{(i)}_{v} \big \Vert_{F}  \leq \epsilon_s$ where $ \Vert \Vert_F$ is the Frobenius norm; \textbf{2)} the loss function $\mathcal{L}$ in MTGNN training is bounded below and $L$-smooth; and \textbf{3)} the gradient of the loss function $\mathcal{L}$ is $\rho$-Lipschitz continuous. Choose step size $\eta = \min \Big \{\frac{2}{L}, \frac{1}{\sqrt{t}}\Big \}$. There exists a constant $D > 0$ such that:
$$\min_{ 1 \leq t \leq T} \big \Vert \nabla \mathcal{L}(W_t) \big \Vert_F^2 \leq \big [2\mathcal{L}(W_{0}) - \mathcal{L}(W^\ast) + \rho D \big ] \frac{1}{\sqrt{T}},$$
where $W_{0}$, $W_t$ and $W^\ast$ are the initial, step-t and optimal model parameters, respectively. 
\end{theorem}
The formal version of Theorem~\ref{thm:convergent} along with its proof can be found in Appendix~\ref{sec:app_proof}.
Theorem~\ref{thm:convergent} indicates that the convergence rate of MSPipe is $O(T^{-\frac{1}{2}})$, which shows that our approach maintains the same convergence rate as vanilla sampling-based GNN training methods ($O(T^{-\frac{1}{2}})$ ~\cite{chen2017stochastic, cong2020minimal, cong2021importance}).


\section{Experiments} 
\label{sec:4}
We conduct experiments to evaluate the proposed framework MSPipe, targeting answering the following research questions:

$\bullet$ Can MSPipe outperform state-of-the-art baseline MTGNN training systems on different models and datasets? (Section~\ref{sec:exp_expedited})

$\bullet$ Can MSPipe maintain the model accuracy and preserve the convergence rate? (Section~\ref{sec:exp_expedited} and ~\ref{sec:exp_con_rate})

$\bullet$ How do the key designs in MSPipe contribute to its overall performance, and what is its sensitivity to hyperparameters? (Section~\ref{sec:exp_minimal} and ~\ref{sec:exp_mitigate})

$\bullet$ How are the memory footprint and GPU utilization when applying staleness in MSPipe?(Section~\ref{sec:exp_gpu_mem})

\renewcommand\arraystretch{0.88}
\begin{table*}[t]
\caption{AP of dynamic link prediction and speedup. The best and second-best results are emphasized in \textbf{bold} and \underline{underlined}. The AP difference smaller than 0.1\% is considered the same. 
The results are averaged over 3 trials with standard deviations.
}
\resizebox{\textwidth}{!}{
\begin{tabular}{cccccccccccc}
\toprule
\multirow{2}{*}{\textbf{Model}} & \multirow{2}{*}{\textbf{Dataset}} & \multicolumn{2}{c}{\textbf{REDDIT}} & \multicolumn{2}{c}{\textbf{WIKI}} & \multicolumn{2}{c}{\textbf{MOOC}} & \multicolumn{2}{c}{\textbf{LASTFM}} & \multicolumn{2}{c}{\textbf{GDELT}} \\ \cline{3-12} 
 &  & \textbf{\small{AP(\%)}} & \textbf{Speedup} & \textbf{\small{AP(\%)}} & \textbf{Speedup} & \textbf{\small{AP(\%)}} & \textbf{Speedup} & \textbf{\small{AP(\%)}} & \textbf{Speedup} & \textbf{\small{AP(\%)}} & \textbf{Speedup} \\ \midrule
\multirow{5}{*}{TGN} & TGL & \textbf{99.82(0.01)} & 1$\times$ & \textbf{99.43(0.03)} & 1$\times$ & \textbf{99.42(0.03)} & 1$\times$ & \underline{87.21(1.90)} & 1$\times$ & \textbf{98.23(0.05)} & 1$\times$ \\
 & Presample & \textbf{99.80(0.01)} & 1.16$\times$ & \textbf{99.43(0.03)} & 1.12$\times$ & \textbf{99.40(0.03)} & 1.16$\times$ & \underline{87.12(1.51)} & 1.36$\times$ & 98.18(0.05) & 1.32$\times$ \\
 & MSPipe & \textbf{99.81(0.02)} & \textbf{1.77$\times$} & 99.14(0.03) & \textbf{1.54$\times$} & \underline{99.32(0.03)} & \textbf{1.50$\times$} & 86.93(0.89) & \textbf{2.00$\times$} & \textbf{98.25(0.06)} & \textbf{2.36$\times$} \\ 
 & MSPipe-S & \textbf{99.82(0.01)} & \underline{1.72$\times$} & \underline{99.39(0.03)} & \underline{1.52$\times$} & \textbf{99.48(0.03)} & \underline{1.47$\times$} & \textbf{87.93(1.26)}& \underline{1.96$\times$} & \textbf{98.29(0.04)} & \underline{2.26$\times$} \\ 
 \midrule
\multirow{5}{*}{JODIE} & TGL & \textbf{99.63(0.02)} & 1$\times$ & \textbf{98.40(0.03)} & 1$\times$ & \textbf{98.64(0.01)} & 1$\times$ & \underline{73.04(2.89)} & 1$\times$ & 98.01(0.07) & 1$\times$ \\
 & Presample & \textbf{99.62(0.03)} & 1.10$\times$ & \textbf{98.41(0.03)} & 1.14$\times$ & \textbf{98.61(0.03)} & 1.09$\times$ & \underline{72.96(2.68)} & 1.37$\times$ & 98.04(0.05) & 1.73$\times$ \\
 & MSPipe & \textbf{99.62(0.02)} & \textbf{1.55$\times$} & \underline{97.24(0.02)} & \textbf{1.65$\times$} & \textbf{98.63(0.02)} & \textbf{1.50$\times$} & 71.7(2.84) & \textbf{1.87$\times$} & \underline{98.12(0.08)} & \textbf{2.28$\times$} \\ 
 & MSPipe-S & \textbf{99.63(0.02)} & \underline{1.50$\times$} & \underline{97.61(0.02)} & \underline{1.54$\times$} & \textbf{98.66(0.02)} & \underline{1.48$\times$} & \textbf{76.32(2.45)} & \underline{1.79$\times$} & \textbf{98.23(0.05)} & \underline{2.23$\times$} \\
 \midrule
\multirow{5}{*}{APAN} & TGL & \textbf{99.62(0.03)} & 1$\times$ & \textbf{98.01(0.03)} & 1$\times$ & \textbf{98.60(0.03)} & 1$\times$ & \underline{73.37(1.59)} & 1$\times$ & 95.80(0.02) & 1$\times$ \\
 & Presample & \textbf{99.65(0.02)} & 1.38$\times$ & \textbf{98.03(0.03)} & 1.06$\times$ & \textbf{98.62(0.03)} & 1.30$\times$ & 73.24(1.70) & 1.49$\times$ & 95.83(0.04) & 1.71$\times$ \\
 & MSPipe & \textbf{99.63(0.03)} & \textbf{2.03$\times$} & 96.43(0.04)& \textbf{1.78$\times$} & \underline{98.38(0.02)} & \textbf{1.91$\times$} & 72.41(1.21) & \textbf{2.37$\times$} & \underline{95.94(0.03)} & \textbf{2.45$\times$} \\
 & MSPipe-S & \textbf{99.64(0.03)} & \underline{1.96$\times$} & \underline{97.12(0.03)} & \underline{1.63$\times$} & \textbf{98.64(0.03)} & \underline{1.77$\times$} & \textbf{76.08(1.42)} & \underline{2.19$\times$} & \textbf{96.02(0.03)} & \underline{2.41$\times$} \\
 \bottomrule
\end{tabular}
}
\label{tab:train}
\end{table*}

\renewcommand\arraystretch{0.5}
\begin{table}[t]
\small
\centering
\caption{The detailed statistics of the datasets. $|d_v|$ and $|d_e|$ show the dimensions of node features and edge features.}
\begin{tabular}{llllllll}
\toprule
Dataset & $|V|$ & $|E|$ & $|d_v|$ & $|d_e|$ & Duration \\
\midrule
Reddit~\cite{kumar2019jodie} & 10,984 & 672,447 & 0 & 172 & 1 month \\
WIKI~\cite{kumar2019jodie} & 9,227 & 157,474 & 0 & 172 & 1 month \\
MOOC~\cite{kumar2019jodie} & 7,144 & 411,749 & 0 & 128 & 17 months \\
LastFM~\cite{kumar2019jodie} & 1,980 & 1,293,103 & 0 & 128 & 1 month \\
GDELT~\cite{zhou2022tgl} & 16,682 & 191,290,882 & 413 & 186 & 5 years \\
\bottomrule
\end{tabular}
\label{tab:datasets}
\end{table}

\subsection{Experiment settings}
\textbf{Testbed.} The main experiments are conducted on a machine equipped with two 64-core AMD EPYC CPUs, 512GB DRAM, and four NVIDIA A100 GPUs (40GB), and the scalability experiments are conducted on two of such machines with 100Gbps interconnect bandwidth. 

\textbf{Datasets and Models.} We evaluate MSPipe on five temporal datasets: REDDIT, WIKI, MOOC, LASTFM~\cite{kumar2019jodie} and a large dataset GDELT~\cite{zhou2022tgl}. Table~\ref{tab:datasets} summarizes the statistics of the temporal datasets. On each dataset, we use the same 70\%-15\%-15\% chronological train/validation/test set split as in previous works~\cite{xu2020tgat, rossi2021tgn}. 
We train 3 state-of-the-art memory-based TGNN models, JODIE~\cite{kumar2019jodie}, TGN~\cite{rossi2021tgn} and APAN~\cite{wang2021apan}. The implementations of TGN, JODIE, and APAN are modified from TGL~\cite{zhou2022tgl} which was optimized by TGL to achieve better accuracy than their original versions.

\textbf{Baselines.} We adopt \textbf{TGL}~\cite{zhou2022tgl}, a state-of-the-art MTGNN training system, as the synchronous MTGNN training baseline. 
We also implement the \textbf{Presample} (with pre-fetching features) mechanism similar to SAILENT~\cite{kaler2022accelerating} on TGL as a stricter baseline, which provides a parallel sampling and feature fetching scheme by executing them in advance. 
We implement \textbf{MSPipe} on PyTorch ~\cite{paszke2019pytorch} and DGL~\cite{wang2019deep}, supporting both single-machine multi-GPU and multi-machine distributed MTGNN training. \textbf{MSPipe-S} is MSPipe with staleness mitigation from similar neighbors with $\lambda$ set to 0.95. Noted that MSPipe does not enable the staleness mitigation by default. The implementation details of MSPipe can be found in Appendix~\ref{sec:app_impl}.

\textbf{Training settings.} To ensure a fair comparison, we used the same default hyperparameters as TGL, including a learning rate of 0.0001, a local batch size of 600 (4000 for the GDELT dataset), and hidden dimensions and memory dimensions of 100. We train each dataset for 100 epochs, except for GDELT, which was trained in 10 epochs. We sampled the 10 most recent 1-hop neighbors for all datasets and constructed mini-batches with an equal number of positive and negative node pairs for sampling and subgraph construction during training and evaluation. The experiments are conducted under the transductive learning setting and we use average precision for evaluation metrics. 
For a more comprehensive analysis of various batch sizes, we provide detailed experiments in Appendix~\ref{sec:app_batch_size_sens}.

\subsection{Expedited Training While Maintaining Accuracy} \label{sec:exp_expedited}
The results in Table~\ref{tab:train} show that MSPipe improves the training throughput while maintaining high model accuracy. AP in the table stands for average model precision evaluated on the test set.

\begin{figure}[t]
\vspace{-1mm}
\subfigure[LASTFM]{
\begin{minipage}[t]{0.49\linewidth}
\includegraphics[width=0.90\linewidth]{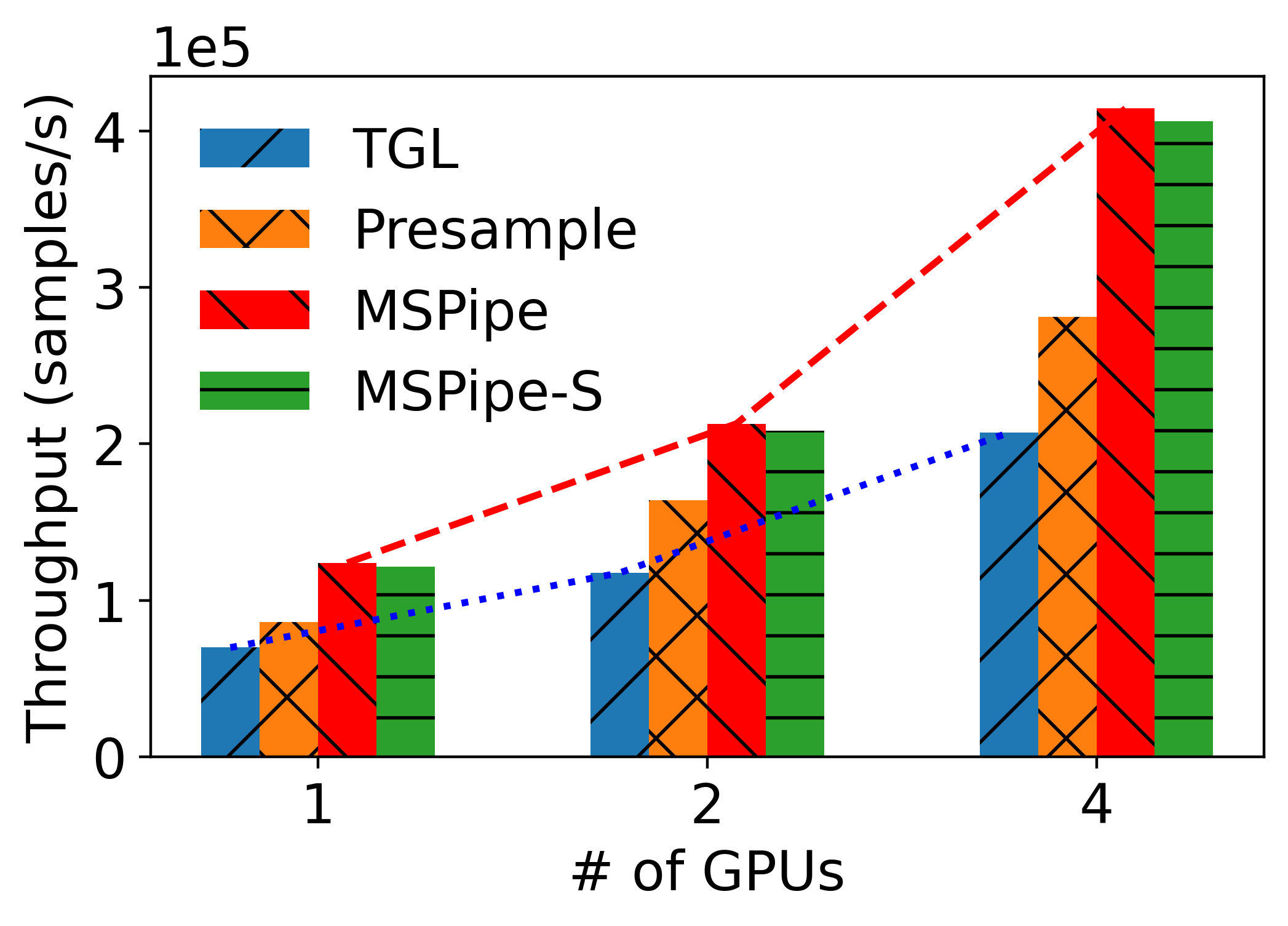}
\end{minipage}
}%
\subfigure[GDELT]{
\begin{minipage}[t]{0.49\linewidth}
\includegraphics[width=0.90\linewidth]{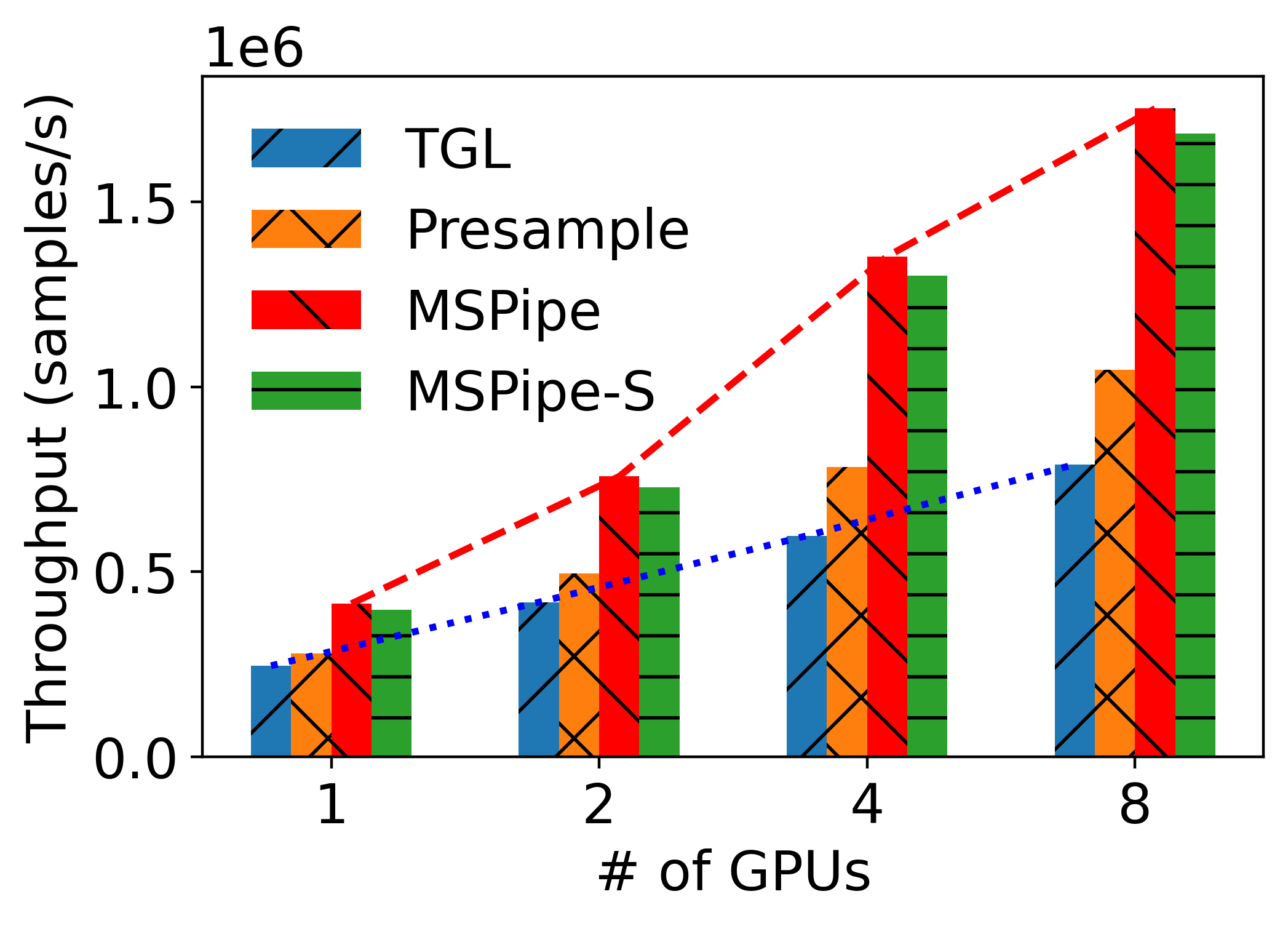}
\label{fig:scale_gdelt}
\end{minipage}
}%
\vspace{-4mm}
\caption{Scalability of training TGN.}
\label{fig:scalability}
\vspace{-4mm}
\end{figure}

\textbf{Training Throughput.} 
We observe that MSPipe is 1.50$\times$ to 2.45$\times$ faster than TGL, and achieves up to 104\% speed-up as compared to the Presample mechanism. MSPipe obtains the best speed-up on GDELT, which can be attributed to the relatively smaller proportion of execution time devoted to the MTGNN training stage compared to other datasets (as shown in Table~\ref{tab:breakdown}). This is mainly because MSPipe effectively addresses the primary bottlenecks in MTGNN training by breaking temporal dependencies between iterations and ensuring uninterrupted progression of the MTGNN training stage, thereby enabling seamless overlap with other stages. Consequently, the total training time is predominantly determined by the uninterrupted MTGNN training stage. Notably, a smaller MTGNN training stage results in a larger speed-up, further contributing to the superior performance of MSPipe.

\textbf{Model Accuracy.} 
MSPipe without staleness mitigation can already achieve comparable test average precision with TGL on all datasets, with a marginal degradation ranging from 0 to 1.6\%. This can be attributed to the minimal staleness mechanism and proper pipeline scheduling in MSPipe.

\textbf{Staleness Mitigation.} 
With the proposed staleness mitigation mechanism, MSPipe-S consistently achieves higher average precision than MSPipe across all models and datasets. Notably, MSPipe-S achieves the same test accuracy as TGL on REDDIT and MOOC datasets, while surpassing TGL's model performance on LastFM and GDELT datasets. MSPipe-S introduces a minimal overhead of only 3.73\% on average for the staleness mitigation process. This demonstrates the efficiency of the proposed mechanism in effectively mitigating staleness while maintaining high-performance. 

\textbf{Scalability.} Figure~\ref{fig:scalability} presents the training throughput with different numbers of GPUs on LastFM and GDELT datasets.
MSPipe achieves not only consistent speed-up but also
up to 83.6\% scaling efficiency on a machine, which is computed as the ratio of the speed-up achieved by using 4 GPUs to the ideal speed-up,
outperforming other baselines. We also scale TGN training on GDELT to two machines with eight GPUs in Figure~\ref{fig:scale_gdelt}. Without explicit optimization for inter-machine communication, MSPipe still outperforms the baselines and exhibits better scalability.

\textbf{GPU sampler Analysis.} Although MSPipe utilizes a GPU sampler for faster sampling, we found that our sampler is 24.3\% faster than TGL's CPU sampler for 1-hop most recent sampling, which accounts for only 3.6\% of the total training time as shown in Table~\ref{tab:gpu_sample} in Appendix~\ref{sec:app_gpu_sample}. Therefore, the performance gain is primarily attributed to our pipeline mechanism and resource-aware minimal staleness schedule but not to the acceleration of the sampler.

\subsection{Preserving Convergence Rate} \label{sec:exp_con_rate}
To validate that MSPipe can maintain the same convergence rate as vanilla sampling-based GNN training without applying staleness ($O(T^{-\frac{1}{2}}$)), we compare the training curves of all models on all datasets in Figure~\ref{fig:converge} (the complete results can be found in
Appendix~\ref{sec:app_convergence_result}). 
We observe that MSPipe's training curves largely overlap with those of vanilla methods (TGL and Presample), verifying our theoretical results in Section~\ref{sec:theory}. With staleness mitigation, MSPipe-S can achieve even better and more steady convergence (e.g., on WIKI and LastFM) than others.

\begin{figure}[t]
\subfigure[WIKI]{
\begin{minipage}[t]{0.49\linewidth}
\includegraphics[width=0.95\linewidth]{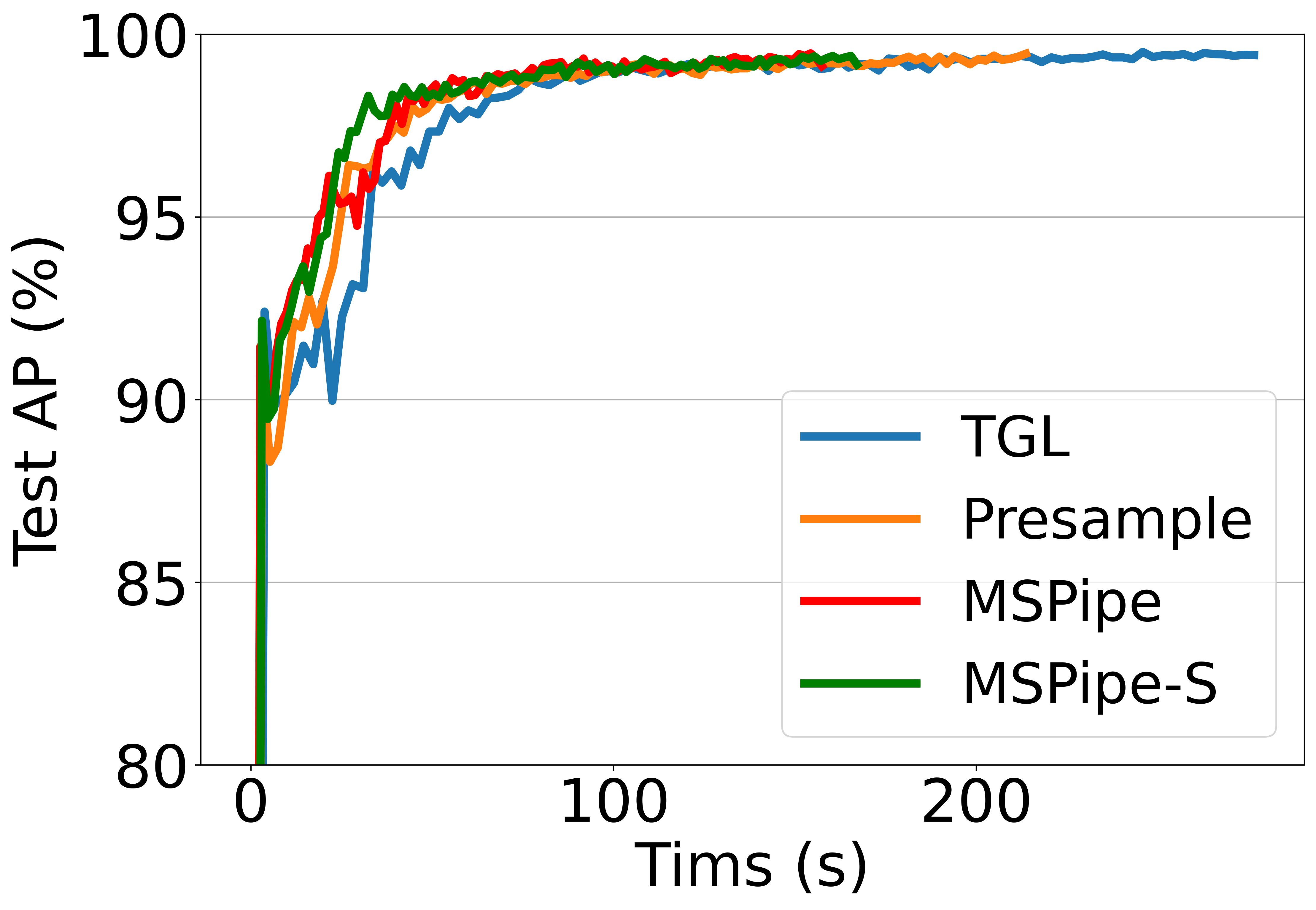}
\end{minipage}%
}%
\subfigure[LASTFM]{
\begin{minipage}[t]{0.49\linewidth}
\includegraphics[width=0.95\linewidth]{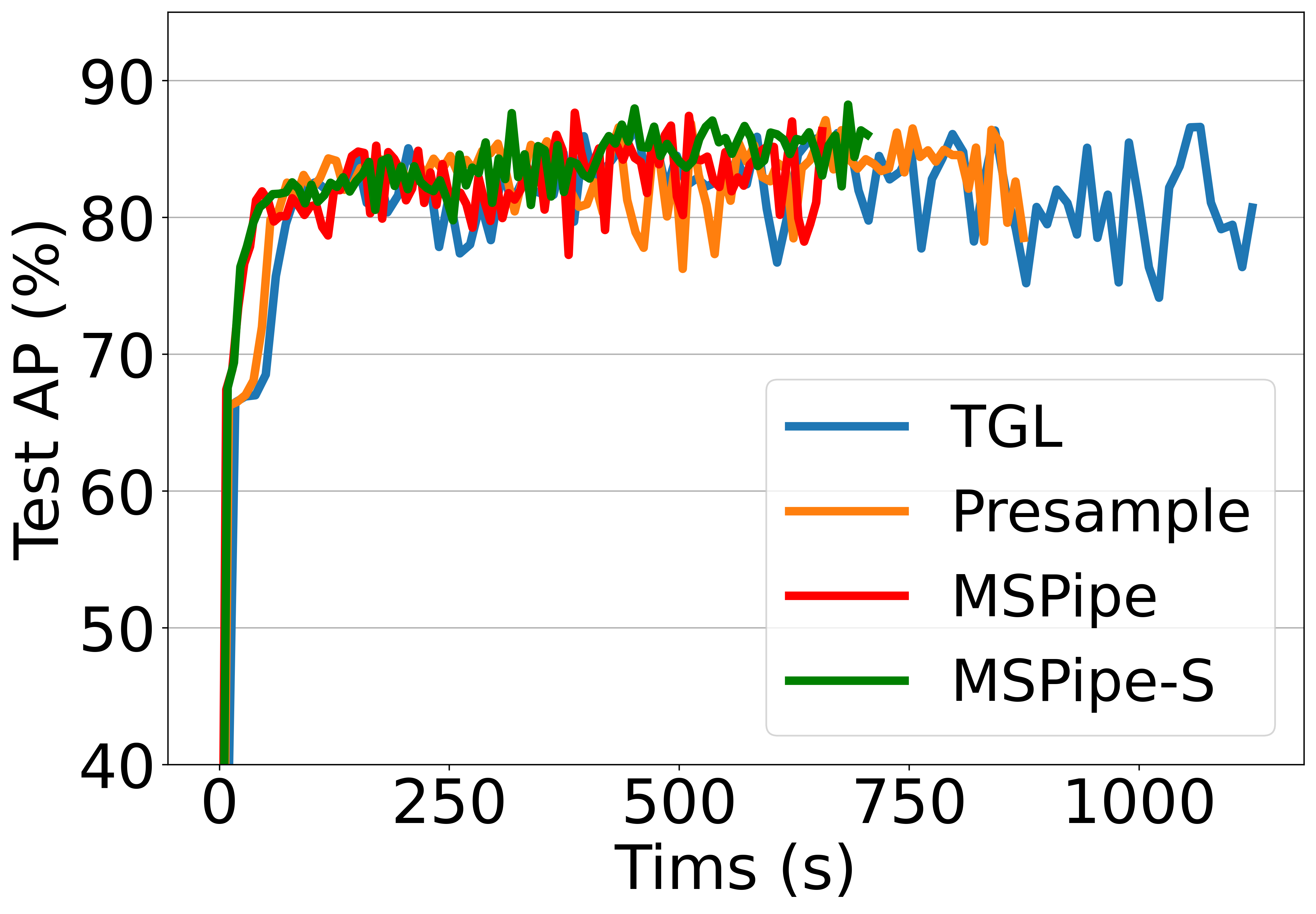}
\end{minipage}
}%
\vspace{-4mm}
\caption{Convergence of TGN training. x-axis is the wall-clock training time, and y-axis is the test average precision.}
\label{fig:converge}
\vspace{-4mm}
\end{figure}

\begin{figure}[t]
\subfigure[MOOC]{
\begin{minipage}[t]{.48\linewidth} 
\includegraphics[width=\linewidth]{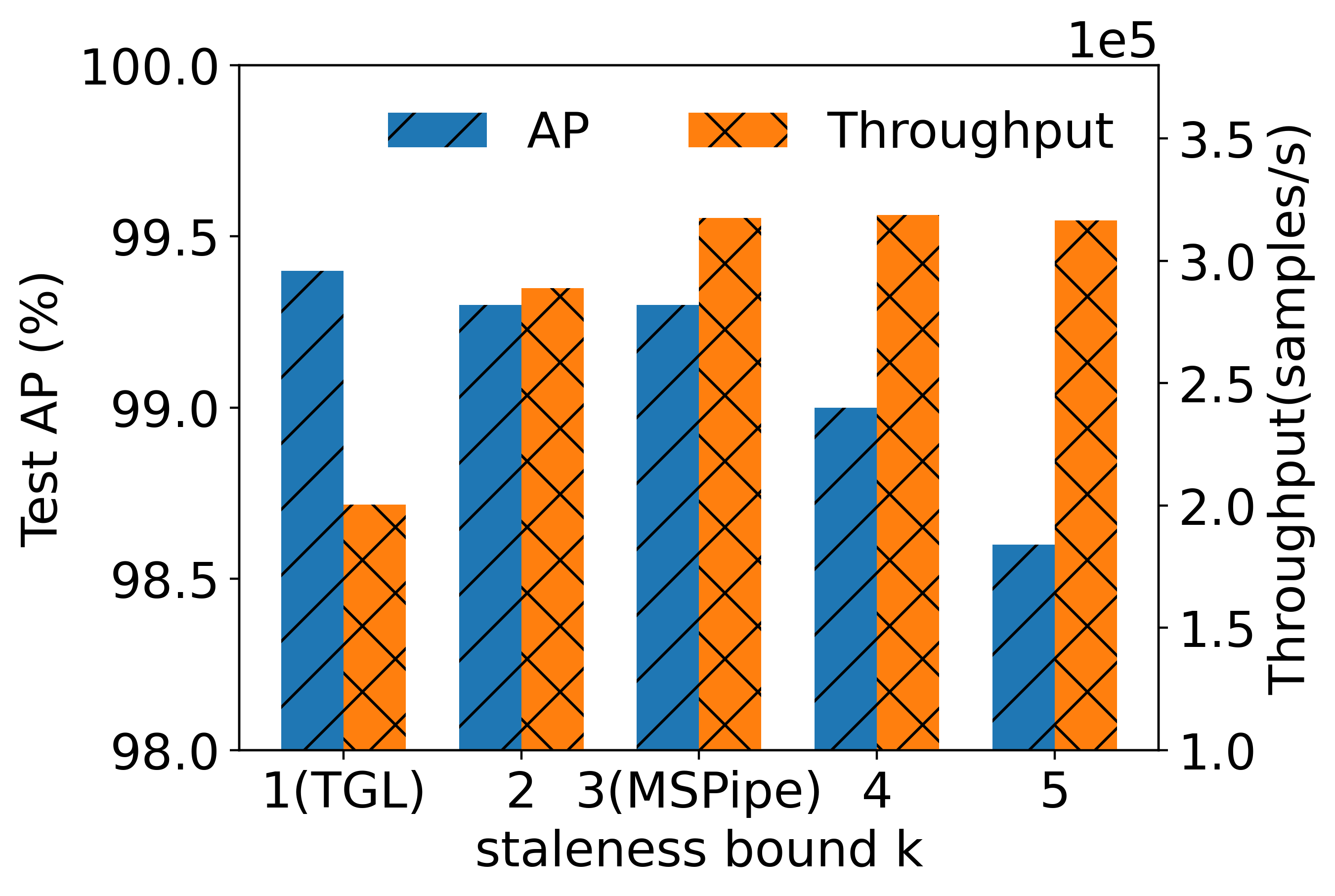}
\end{minipage}%
}
\subfigure[GDELT]{
\begin{minipage}[t]{.48\linewidth} 
\includegraphics[width=\linewidth]{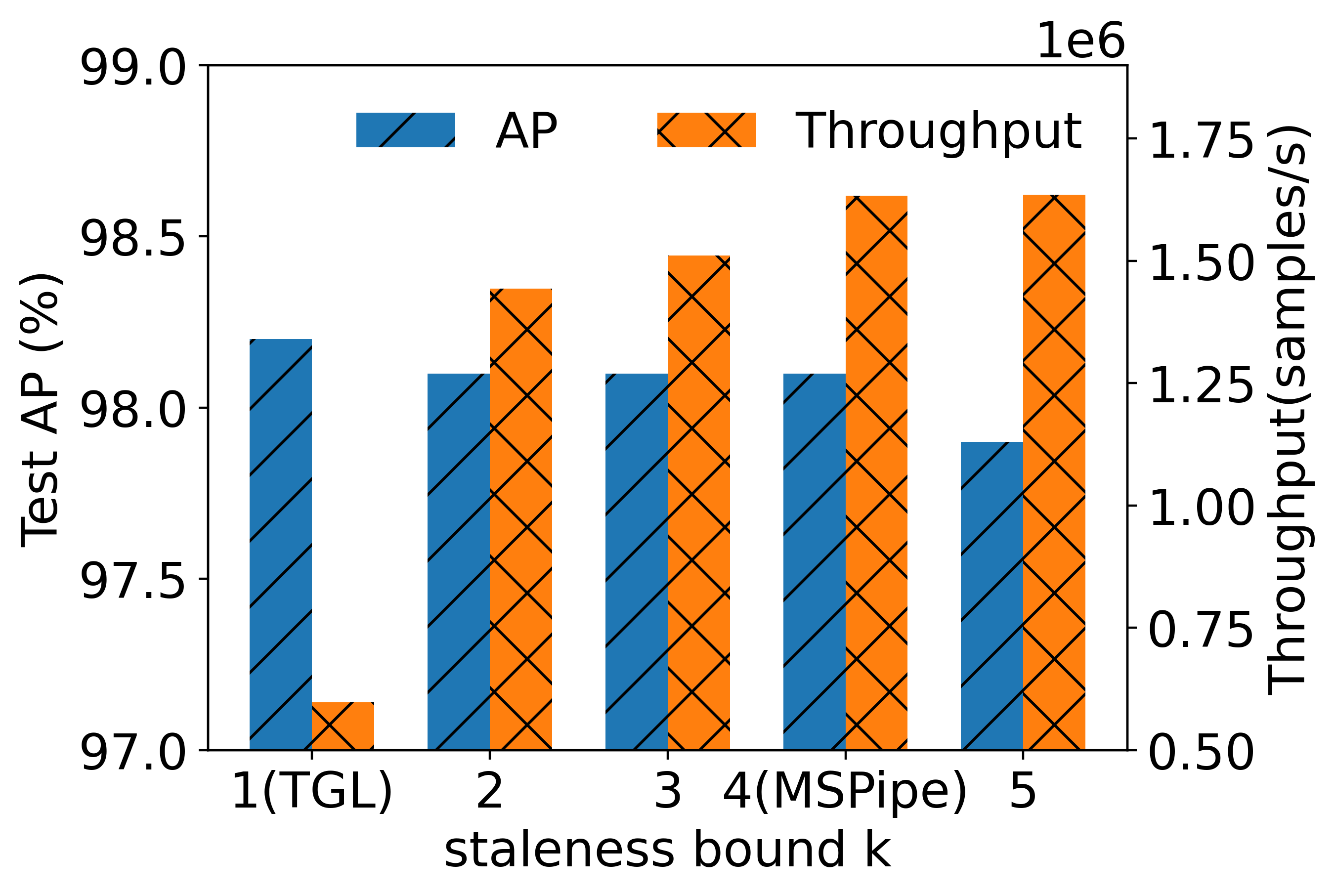}
\end{minipage}%
}
\vspace{-3mm}
\caption{Throughput and AP on different staleness bound}
\label{fig:optimal_staleness}
\vspace{-2mm}
\end{figure}

\subsection{Stall-free Minimal Staleness Bound} \label{sec:exp_minimal}
To further validate that MSPipe can find the minimal staleness bound without delaying the MTGNN training stage, we conduct a comparative analysis of accuracy and throughput between the minimal staleness bound computed by MSPipe and other different staleness bounds $k$.  The results, depicted in Figure~\ref{fig:optimal_staleness},
consistently demonstrate that MSPipe achieves the highest throughput while maintaining the best accuracy compared to other staleness bound options. Additionally, the computed minimal staleness bounds for various datasets range from 2 to 4, providing further evidence for the necessity of accurately determining the minimal staleness bound rather than relying on random selection. Note that $k=1$ represents the baseline method of TGL without applying staleness.

\subsection{Staleness Mitigation Mechanism} \label{sec:exp_mitigate}
\begin{figure*}[t]
\begin{minipage}{.30\linewidth} 
        \centering
    \includegraphics[width=0.92\linewidth]{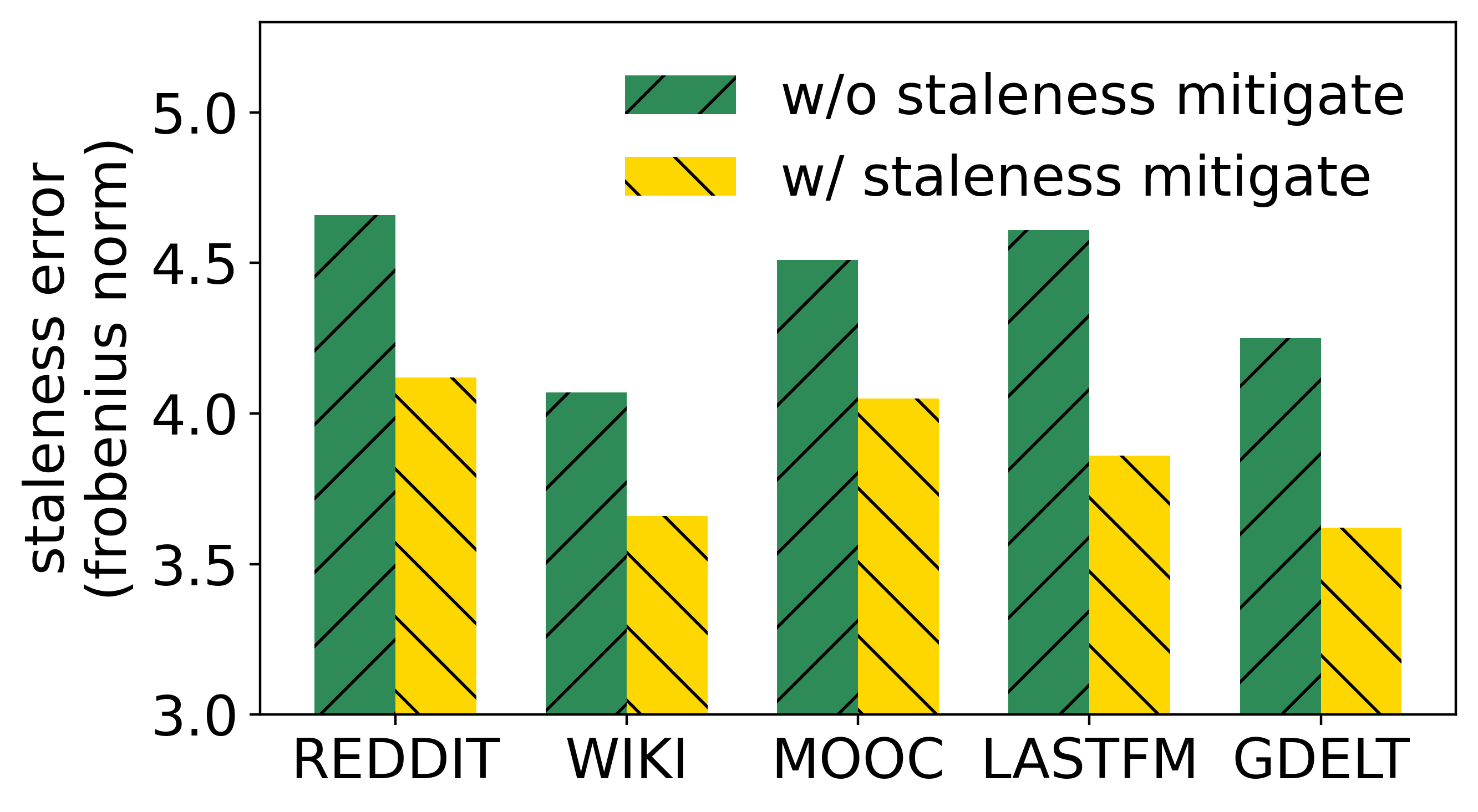}
    \vspace{-3mm}
    \caption{Staleness error comparison on TGN}
    \label{fig:staleness_error}
    \end{minipage}
\hspace{4pt}
    \begin{minipage}{.32\linewidth} 
        \centering
    \includegraphics[width=0.86\linewidth]{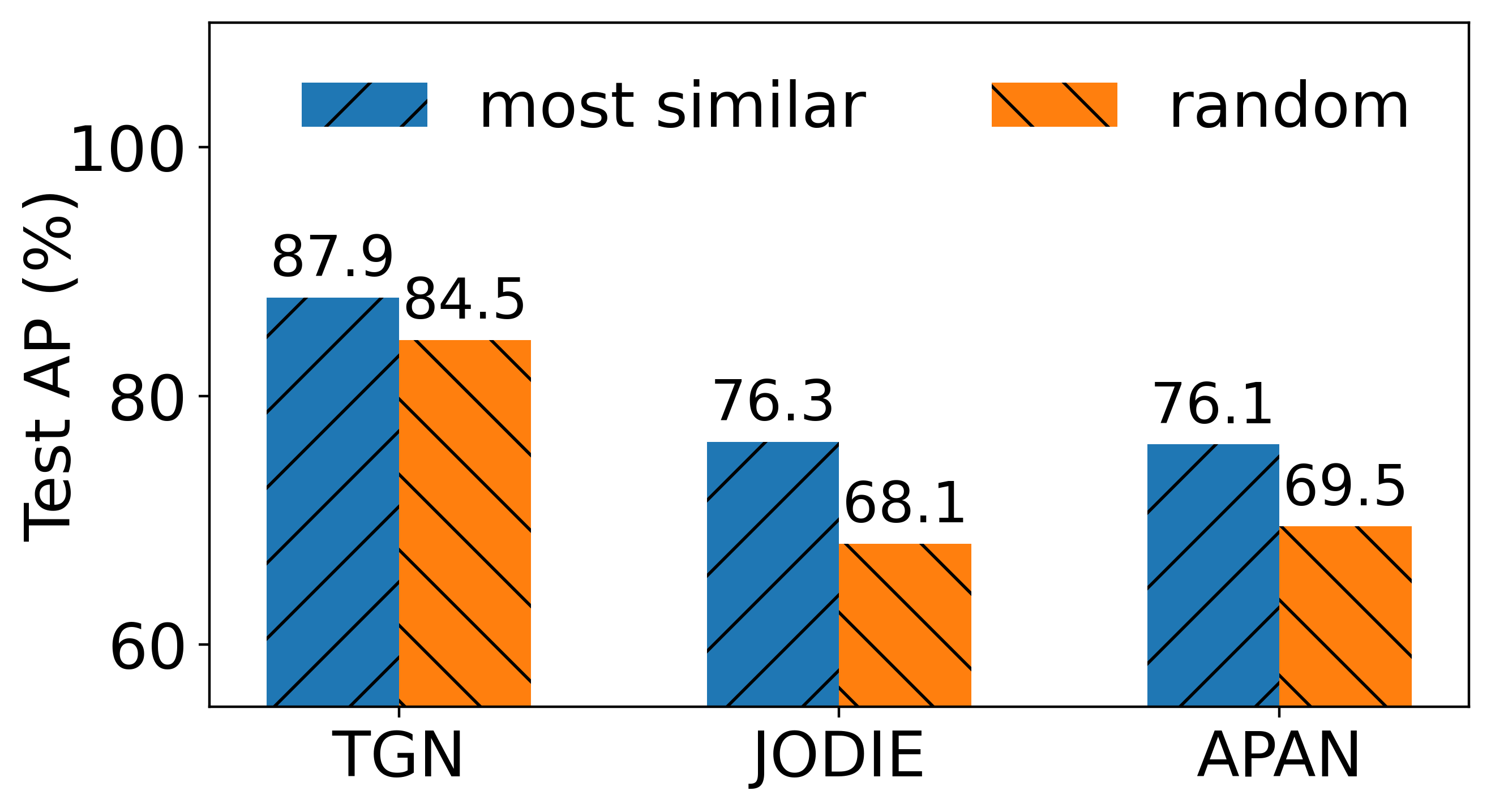}
    \vspace{-3mm}
    \caption{Staleness mitigation with most similar or random nodes on LastFM}
    \label{fig:random_similar}
    \end{minipage}
\hspace{4pt}
    \begin{minipage}{.30\linewidth} 
        \centering
    \includegraphics[width=0.8\linewidth]{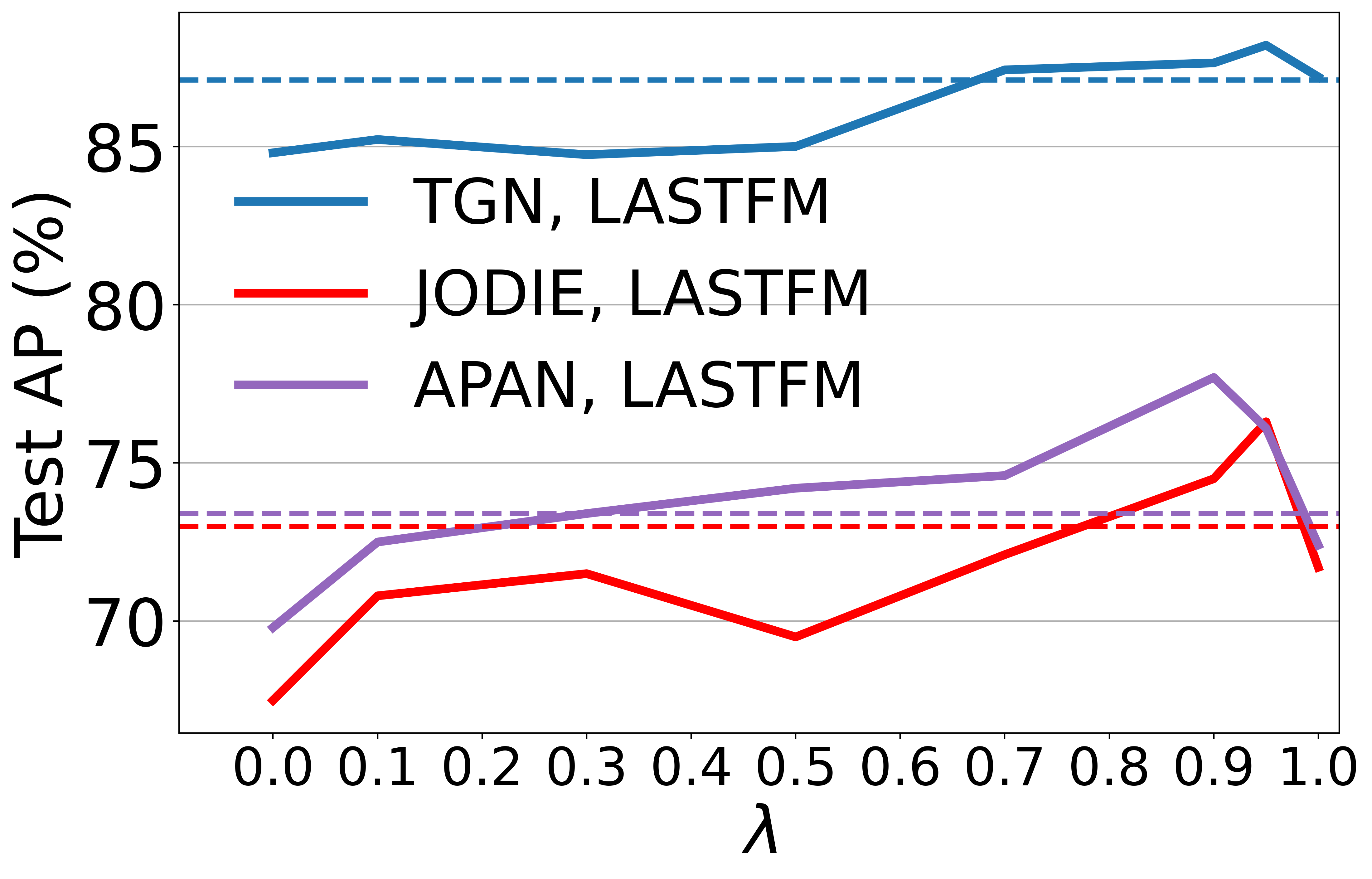}
    \vspace{-3mm}
    \caption{Hyperpara-meter analysis}
    \label{fig:lambda}
    \end{minipage}
\vspace{-2mm}
\end{figure*}

\textbf{Error reduction.} To better understand the accuracy enhancement and convergence speed-up achieved by MSPipe-S, we conduct a detailed analysis of the intermediate steps involved in our staleness mitigation mechanism. Specifically, we refer to Theorem~\ref{thm:convergent}, where we assume the existence of a bounded difference $\epsilon_s$ between the stale node memory vector $\tilde{s}^{(i)}_{v}$ and the precise node memory vector $s^{(i)}_{v}$. To assess the effectiveness of our staleness mitigation mechanism, we compare the mitigated staleness error $\big \Vert \hat{s}^{(i)}_{v} - s^{(i)}_{v} \big \Vert_{F}$ obtained after applying our mechanism with the original staleness error $\big \Vert \tilde{s}^{(i)}_{v} - s^{(i)}_{v} \big \Vert_{F}$. As shown in Figure~\ref{fig:staleness_error}, MSPipe-S consistently reduces the staleness error across all datasets, validating the theoretical guarantee and the effectiveness in enhancing accuracy.

\textbf{Benefit of using most-similar neighbors.}
We further investigate our staleness mitigation mechanism by comparing using the most similar and active nodes for staleness mitigation with utilizing random active nodes, on the LastFM dataset. In Figure~\ref{fig:random_similar}, we observe that our proposed most similar mechanism leads to better model performance, while a random selection from the active nodes would even degrade model accuracy. This can be attributed to the fact that similar nodes possess resemblant representations, enabling the stale node to acquire more updated information. Further details regarding the comparison of memory similarity between the most similar nodes and random nodes can be found in
Appendix~\ref{sec:app_c3_fullexp}.

\renewcommand\arraystretch{0.95}
\begin{table}[t]
\caption{Additional memory overhead of TGN when applying staleness. 
Other models can be found in Appendix~\ref{sec:app_mem}.
}
\vspace{-1mm}
\resizebox{1\linewidth}{!}{
\begin{tabular}{llllll}
\toprule
Overhead\textbackslash Dataset & REDDIT & WIKI & MOOC & LastFM & GDELT \\
\midrule
Addition & 48.8MB & 38.4MB & 42.9MB & 38.1MB & 1.17GB \\
Upperbound & 51.4MB & 34.3MB & 44.3MB & 44.3MB & 1.35GB \\
GPU Mem (40GB) portion & 0.12\% & 0.10\% & 0.11\% & 0.09\% & 2.92\% \\
\bottomrule
\end{tabular}}
\label{tab:mem_gpu}
\vspace{-3mm}
\end{table}

\textbf{Hyperparameter analysis.} We examine the effect of hyperparameter $\lambda$ on test accuracy, as depicted in Figure~\ref{fig:lambda}. 
The dashed horizontal lines in the figure denote the AP from the TGL baseline for comparison.
We find that mitigating staleness with a larger $\lambda$ ($>0.8$) results in better model performance than TGL's results, indicating that we should retain more of the original stale memory representations and apply a small portion of mitigation from their similar ones. Notably, setting $\lambda$ to 1 causes MSPipe-S to revert to the standard MSPipe configuration, thereby omitting the staleness mitigation strategy entirely.

\subsection{GPU memory and utilization} \label{sec:exp_gpu_mem}
We present an analysis of the memory overhead associated with MSPipe, as staleness-based strategies generally require additional memory to enhance training throughput. Unlike other asynchronous training frameworks~\cite{chen2022sapipe,li2018pipe, wan2022pipegcn, peng2022sancus} that introduce staleness during DNN or GNN parameter learning, MSPipe only introduces staleness within the memory module to break temporal dependencies. Each subgraph is executed sequentially, resulting in no additional hidden states during MTGNN computation. The extra memory consumption in MSPipe comes from the prefetched subgraph, including node/edge features and memory vectors. We provide a detailed analysis to determine the upper bound of this additional memory overhead, assuming maximum neighbor size for all nodes (i.e., $\mathcal{N}=10$).
Additionally, we measure the actual memory consumption using \textit{torch.cuda.memory\_summary()} API in experiments across all datasets. Table~\ref{tab:mem_gpu} shows that the observed additional memory usage in MSPipe aligns with our analyzed upper bound. Moreover, we compare the additional memory cost with the GPU memory size in Table~\ref{tab:mem_gpu}, demonstrating that it constitutes a relatively small proportion (up to 2.92\%) of the modern GPU's capacity.

\begin{figure}[t]
    \centering
    \includegraphics[width=0.8\linewidth]{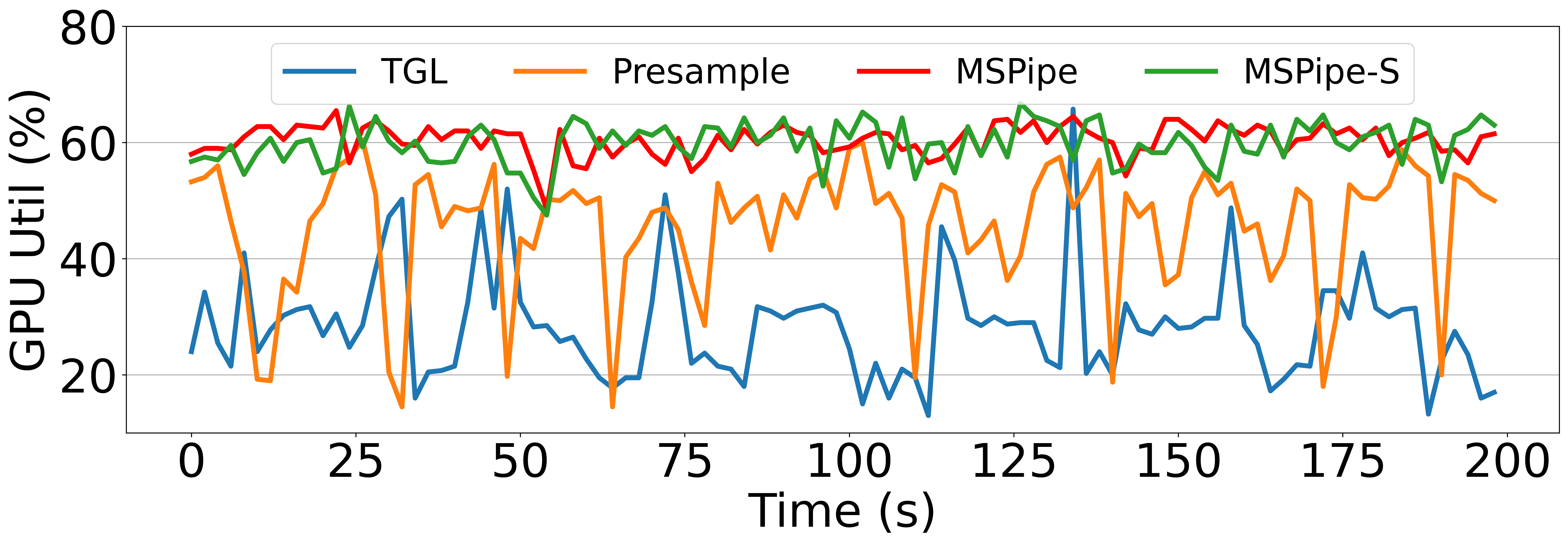}
    \vspace{-3mm}
    \caption{GPU utilization of different methods when training TGN with LastFM dataset.}
    \label{fig:gpu_util}
    \vspace{-3mm}
\end{figure}

Figure~\ref{fig:gpu_util} presents the GPU utilization during the training of the TGN model using the LastFM dataset. The plot showcases the average utilization of 4 A100 GPUs, with a smoothing interval of 2 seconds. The utilization data was collected throughout the training process across multiple epochs, excluding the validation stage. In Figure~\ref{fig:gpu_util}, both MSPipe and MSPipe-S demonstrate consistently high GPU utilization, outperforming the baseline methods that exhibit significant fluctuations. This notable improvement can be attributed to the minimal staleness and pipeline scheduling mechanism introduced in MSPipe, ensuring uninterrupted execution of the MTGNN training stage. In contrast, the TGL and Presample methods require the GPU to wait for data preparation, resulting in decreased GPU utilization and overall performance degradation.

\section{Related Works}
\textbf{Sampling-based mini-batch
training}
has become 
the norm for static GNN and TGNN training~\cite{gandhi2021p3, hamilton2017inductive, waleffe2023mariusgnn, yang2022gnnlab, ying2018graph}, which samples a subset of neighbors of target nodes to generate a subgraph, as input to GNN. The bottlenecks mainly lie in subgraph sampling and feature fetching due to the neighbor explosion problem~\cite{chen2018fastgcn, yan2018spatial}. ByteGNN~\cite{zheng2022bytegnn} and SALIENT~\cite{kaler2022accelerating} adopt pre-sampling and pre-fetching to hide sampling and feature fetching overhead in multi-layer static GNN training. 
These optimizations may not address the bottleneck in TGNN training, where maintaining node memories in sequential order incurs overhead while lightweight sampling and feature fetching are sufficient for single TGNN layer~\cite{rossi2021tgn, zhang2023tiger, wang2021apan, kumar2019jodie}.

\textbf{Asynchronous Distributed Training.} \label{sec:related}
Many studies advocate asynchronous training with staleness for DNN and static GNN models.
For distributed DNN training, previous works ~\cite{li2018pipe, chen2022sapipe, recht2011hogwild, ho2013more, dai2018toward, barkai2019gap}
adopt stale weight gradients on large model parameters to eliminate communication overhead, while GNN models typically have much smaller sizes.
For static GNN training, PipeGCN~\cite{wan2022pipegcn} and Sancus~\cite{peng2022sancus} introduce staleness in node embeddings under the full-graph training paradigm.
Although these methods are effective for static GNNs, their effect is limited when applied to MTGNNs, from three aspects: \textbf{1)} they focus on full graph training and apply staleness between multiple GNN layers to overlap the significant communication overhead with computation. In MTGNN training, the communication overhead is relatively small due to subgraph sampling and the presence of only one GNN layer. \textbf{2)} all previous GNN training frameworks simply introduce a pre-defined staleness bound without explicitly analyzing the relationship between model quality and training throughput, potentially leading to sub-optimal parallelization solutions; \textbf{3)} the unique challenges arising from the temporal dependency caused by memory fetching and updating in MTGNN training have not been adequately addressed by these frameworks.
Therefore, these asynchronous training frameworks for DNN and static GNN are not suitable for accelerating MTGNNs. A detailed analysis of the related work can be found in Appendix~\ref{sec:app_related_work}.


\section{Conclusion}
We present MSPipe, a general and efficient memory-based TGNN training framework that improves training throughput while maintaining model accuracy. MSPipe addresses the unique challenges posed by temporal dependency in MTGNN. MSPipe strategically identifies the minimal staleness bound to adopt and proposes an online scheduler to dynamically control the staleness bound without stalling the pipeline. Moreover, MSPipe employs a lightweight staleness mitigation strategy and provides a comprehensive theoretical analysis for MTGNN training. Extensive experiments validate that MSPipe attains significant speed-up over state-of-the-art TGNN training frameworks while maintaining high model accuracy.

\begin{acks}
We would like to thank the anonymous reviewers and area chairs for their helpful comments. This work is supported in part by grants from Hong Kong RGC under the contracts HKU 17207621, 17203522 and C7004-22G (CRF)
\end{acks}

\bibliographystyle{ACM-Reference-Format}
\balance
\bibliography{reference}

\appendix

\section{Proofs} \label{sec:app_proof}
In this section, we provide the detailed proofs of the theoretical analysis.

\begin{lemma}
    If $f(\cdot)$ is $\beta$-smooth, then we have,
    $$f(y) \leq f(x) + \big \langle \nabla f(x), y-x \big \rangle+\frac{\beta}{2}\Vert y-x \Vert^{2}$$
    Proof. 
    \begin{align*}
        &\big |f(y) - f(x) - \big \langle \nabla f(x), y-x \big \rangle\big | \\
        =& \Big |\int_{0}^{1}\big \langle \nabla f(x) + t(y-x), y-x \big \rangle dt - \big \langle \nabla f(x), y-x \big \rangle\Big | \\
        \leq& \int_{0}^{1}\big |\big \langle \nabla f(x) + t(y-x) - \nabla f(x), y-x \big \rangle\big |dt \\
        \leq& \int_{0}^{1} \Vert \nabla f(x) + t(y-x) - \nabla f(x)\Vert \cdot \Vert y-x \Vert dt \\
        \leq& \int_{0}^{1} t\beta \Vert y-x \Vert^{2} dt \\
        =& \frac{\beta}{2} \Vert y-x \Vert^{2}
    \end{align*}
\end{lemma}

\begin{lemma}
    if $\mathcal{L}(\cdot)$ is $\rho$-Lipschitz smooth, then we have
    $$\big \Vert \nabla \tilde{\mathcal{L}}(W) - \nabla \mathcal{L}(W)\big \Vert_F \leq \rho \epsilon_s$$
    where $\nabla \tilde{\mathcal{L}}\left(W_t\right)$ denote the gradient when stale memoys are used.
\end{lemma}
\textit{Proof.} By the assumption that there is a bounded difference between the stale node memory vector $\tilde{S_i}$ and the exact node memory vector $S_i$ with the staleness bound $\epsilon_s$, we have:
\begin{align*}
    \big \Vert S - \tilde{S} \big \Vert_F \leq \epsilon_s
\end{align*}
By smoothness of $\mathcal{L}(\cdot)$, we have
\begin{align*}
    &\big \Vert \nabla \mathcal{L}(S, W) - \nabla \mathcal{L}(\tilde{S}, W) \big \Vert_F \\
    =& \big \Vert \nabla \tilde{\mathcal{L}}(W) - \nabla \mathcal{L}(W) \big \Vert_F \\
    \leq& \rho \epsilon_s
\end{align*}

\textbf{Learning Algorithms.} In the $t^{th}$ step, we have
\begin{equation}
    \label{equ:sub1}
    W_{t+1}-W_t=-\eta_t \nabla \tilde{\mathcal{L}}\left(W_t\right)
\end{equation}

, where $\nabla \tilde{\mathcal{L}}\left(W_t\right)$ denote the gradient when stale memoys are used and $\eta_t$ is the learning rate.

By Lemma 1 and the $L_f$-smoothness of $\mathcal{L}$ , we have
\begin{align}
    \mathcal{L}(W_{t+1})-\mathcal{L}(W_t) \leq  
    \big \langle W_{t+1}-W_t, \nabla \mathcal{L}(W_t)\big \rangle+\frac{L_f}{2}\|W_{t+1}-W_t\|_F^2
\end{align}

Use Eqn.~\ref{equ:sub1} to substitute, we have
\begin{align} \label{equ:sub_2}
    \mathcal{L}\left(W_{t+1}\right)-\mathcal{L}\left(W_t\right) \leq 
    \underbrace{-\eta_t \langle \nabla \tilde{\mathcal{L}}(W_t), \nabla \mathcal{L}(W_t) \rangle}_{\circled{1}} + \underbrace{\frac{L_f \eta_t^2}{2} \Vert \nabla \tilde{\mathcal{L}}(W_t) \Vert_F^2}_{\circled{2}}
\end{align}

We bound the terms step by step and let $\delta_t = \nabla \tilde{\mathcal{L}}\left(W_{t}\right)- \nabla \mathcal{L}\left(W_t\right)$ to subsitute in Equ.~\ref{equ:sub_2}.

First, For $\circled{1}$, we have
\begin{align*}
    &-\eta_t \big \langle \nabla \tilde{\mathcal{L}}(W_t), \nabla \mathcal{L}(W_t) \big \rangle \\
    =& -\eta_t\big \langle\delta_t+\nabla \mathcal{L}(W_t), \nabla \mathcal{L}(W_t) \big \rangle \\
    =& -\eta_t \Big [\big \langle\delta_t, 
    \nabla \mathcal{L}(W_t)\big \rangle+ \big \Vert \nabla \mathcal{L}(W_t) \big \Vert_F^2 \Big ]
\end{align*}
For \circled{2}, we have
\begin{align*}
    &\frac{L_f \eta_t^2}{2} \big \Vert \nabla \tilde{\mathcal{L}}(W_t) \big \Vert_F^2 \\
    =& \frac{L_f \eta_t^2}{2} \big \Vert \delta_t + \nabla \mathcal{L}(W_t) \big \Vert_F^2 \\
    =& \frac{L_f \eta_t^2}{2} \Big (\big \Vert \delta_t \big \Vert_F^2 + 2\big \langle \delta_t, \nabla \mathcal{L}(W_t) \big \rangle + \big \Vert \nabla \mathcal{L}(W_t) \big \Vert_F^2 \Big )
\end{align*}

Combining both \circled{1} and \circled{2} together and by the choice of learning rate $\eta_t = \frac{1}{L_f}$, we have
\begin{align*}
    \mathcal{L}(W_{t+1})-\mathcal{L}(W_t) \leq - \Big (\eta_t-\frac{L_f}{2} \eta_t^2 \Big) \big \Vert \nabla \mathcal{L}(W_t) \big \Vert_F^2 + \frac{L_f \eta_t^2}{2} \big \Vert \delta_t \big \Vert_F^2
\end{align*}

By Lemma 2. we have $\|\delta_t\|_F^2 \leq \rho \epsilon_s$
\begin{align}
    \label{equ:sub_3}
    \mathcal{L}(W_{t+1})-\mathcal{L}(W_t) \leq - \Big (\eta_t-\frac{L_f}{2} \eta_t^2 \Big) \big \Vert \nabla \mathcal{L}(W_t) \big \Vert_F^2 + \frac{L_f \eta_t^2}{2} \rho \epsilon_s
\end{align}

Rearrange Eqn.~\ref{equ:sub_3} and let $c = \frac{L_f \rho \epsilon_s}{2}$, we have,
\begin{align}
    \Big (\eta_t-\frac{L_f}{2} \eta_t^2 \Big) \big \Vert \nabla \mathcal{L}(W_t) \big \Vert_F^2 \leq \mathcal{L}(W_{t})-\mathcal{L}(W_{t+1}) + \eta_t^2 c
\end{align}

Telescope sum from $t = 1...T$, we have
\begin{align}
    &\sum_{t=1}^T \Big (\eta_t-\frac{L_f \eta_t^2}{2} \Big)\|\nabla \mathcal{L}(W_t)\|_F^2 
 \leq \mathcal{L}(W_0)-\mathcal{L}(W_T)+\sum_{t=1}^T \eta_t^2 c \\
    &\min _{1 \leq t \leq T}\| \nabla \mathcal{L}(W_t)\|_F^2 \leq \frac{\mathcal{L}(W_0)-\mathcal{L}(W_T)}{\sum_{t=1}^T \big (\eta_t-\frac{L_f \eta_t^2}{2} \big)}+\frac{\sum_{t=1}^T \eta_t^2 c}{\sum_{t=1}^T \big (\eta_t-\frac{L_f \eta_t^2}{2} \big)} \label{equ:sub_4}
\end{align}

Substitute Equ.~\ref{equ:sub_4} with $\eta_t = min \{ \frac{1}{\sqrt{t}}, \frac{1}{L_f}\}$ and $\mathcal{L}(W^\ast) \leq \mathcal{L}(W_T)$, we have
\begin{align*}
    &\min _{1 \leq t \leq T}\|\nabla \mathcal{L}(W_t)\|_F^2 \\
    \leq& \Big (2 \big (\mathcal{L}(W_0)-\mathcal{L}(W^\ast) \big) + \frac{c}{L_f} \Big)\frac{1}{\sqrt{T}} \\
    \leq& \Big (2 \big (\mathcal{L}(W_0)-\mathcal{L}(W^\ast) \big ) + \frac{\rho \epsilon_s}{2} \Big) \frac{1}{\sqrt{T}}
\end{align*}
Therefore, the convergence rate of MSPipe is $O(T^{-\frac{1}{2}})$, which maintains the same convergence rate as vanilla sampling-based GNN training methods ($O(T^{-\frac{1}{2}})$ ~\cite{chen2017stochastic, cong2020minimal, cong2021importance}).

\section{MORE DISCUSSION ON THE RELATED WORK} \label{sec:app_related_work}
As discussed before, the key design space of the memory-based TGNN model lies in memory updater and memory aggregator functions. JODIE~\cite{kumar2019jodie} updates the memory using two mutually recursive RNNs and applies MLPs to predict the future representation of a node. Similar to JODIE, TGN~\cite{rossi2021tgn} and APAN~\cite{wang2021apan} use RNN as the memory update function while incorporating an attention mechanism to capture spatial and temporal information jointly. APAN further optimizes inference speed by using asynchronous propagation. A recent work TIGER~\cite{zhang2023tiger} improves TGN by introducing an additional memory module that stores node embeddings and proposes a restarter for warm initialization of node representations.

Moreover, some researchers focus on optimizing the inference speed of MTGNN models: ~\cite{Zhou2022ModelArchitectureCF} propose a model-architecture co-design to reduce computation complexity and external memory access. TGOpt~\cite{wang2023tgopt} leverages redundancies to accelerate inference of the temporal attention mechanism and the time encoder. 

There are several static GNN training schemes with staleness techniques, PipeGCN~\cite{wan2022pipegcn} and Sancus~\cite{peng2022sancus}, as we have discussed the difference in Section~\ref{sec:related}, we would like to emphasize and detail the difference between those works and MSPipe:
\begin{itemize}[leftmargin=*]
    \item \textbf{Dependencies and Staleness}: PipeGCN~\cite{wan2022pipegcn} and Sancus~\cite{peng2022sancus} aim to eliminate inter-layer dependencies in multi-layer GNN training to enable communication-computation overlap. In contrast, MSPipe is specifically designed to tackle temporal dependencies within the memory module of MTGNN training. The dependencies and staleness in MTGNN training pose unique challenges that require distinct theoretical analysis and system designs.
    \item \textbf{The choice of staleness bound}: Previous staleness-based static GNN methods randomly choose a staleness bound for acceleration, which may lead to suboptimal system performance and affect model accuracy. MSPipe strategically decides the minimal staleness bound that can reach the highest throughput without sacrificing the model accuracy.
    \item \textbf{Bottlenecks}: In full-graph training scenarios, such as PipeGCN~\cite{wan2022pipegcn} and Sancus~\cite{peng2022sancus}, the main bottleneck lies in communication between graph partitions on GPUs. Due to limited GPU memory, the graph is divided into multiple parts, leading to increased communication time during full graph training. Therefore, these methods aim to optimize the communication-computation overlap to improve training throughput. In contrast, in MTGNN training, the main bottleneck stems from maintaining the memory module on the CPU and the associated challenges of updating and synchronizing it with CPU storage across multiple GPUs~\cite{zhong2023gnnflow}. MSPipe focuses on addressing this specific bottleneck. Furthermore, unlike full graph training where the entire graph structure needs to be stored in the GPU, MTGNN adopts a sampling-based subgraph training approach. As a result, the communication overhead in MTGNN is significantly smaller than full graph training.
    \item \textbf{Training Paradigm and Computation Patterns}: PipeGCN~\cite{wan2022pipegcn} and Sancus~\cite{peng2022sancus} are tailored for full-graph training scenarios, which differ substantially from MTGNN training in terms of training paradigm, computation patterns, and communication patterns. MTGNNs typically involve sample-based subgraph training, which presents unique challenges and constraints not addressed by full graph training approaches. Therefore, the full graph training works cannot support MTGNN training.
    \item \textbf{Multi-Layer GNNs vs Single-Layer MTGNNs}: PipeGCN~\cite{wan2022pipegcn} and Sancus~\cite{peng2022sancus} lies on the assumption that the GNN have multiple layers (e.g., GCN~\cite{kipf2016semi}, GAT~\cite{ying2018graph}) and they break the dependencies among multiple layers to overlap communication with computation. While memory-based TGNNs only have one layer with a memory module~\cite{zhou2022tgl, dgb_neurips_D&B_2022, rossi2021tgn, kumar2019jodie, wang2021apan}, which makes their methods lose efficacy for MTGNNs.
\end{itemize}
\section{Training Time Breakdown}
\label{sec:breakdown}
\subsection{Profiling setups}
We use TGL~\cite{zhou2022tgl}, the SOTA MTGNN training framework, on a server equipped with 4 A100 GPUs for profiling, which is the same as the experiment testbed introduced in the section~\ref{sec:4}. The local batch size for the REDDIT, WIKI, MOOC, and LastFM datasets is set to 600, while for the GDELT dataset, it is set to 4000. All the breakdown statistics are averaged over 100 epochs. All these hyperparameters are the same as the experiments. We firmly believe that, by leveraging TGL's highly optimized performance, we can evaluate bottlenecks and areas for improvement, further justifying the need for our proposed MSPipe framework.

\renewcommand\arraystretch{0.7}
\begin{table}[htbp]
\centering
\caption{Training time breakdown of JODIE model}
\resizebox{\linewidth}{!}{
\begin{tabular}{llllll}
\toprule
\textbf{Dataset} & \textbf{Sample} & \textbf{\makecell{Fetch\\feature}} & \textbf{\makecell{Fetch\\memory}} & \textbf{\makecell{Train\\MTGNN}} & \textbf{\makecell{Update\\memory}} \\ 
 \midrule
REDDIT~\cite{kumar2019jodie} & 4.14\% & 8.05\% & 7.36\% & 50.11\% & 30.34\% \\
WIKI~\cite{kumar2019jodie} & 2.20\% & 1.10\% & 4.95\% & 46.70\% & 45.05\% \\
MOOC~\cite{kumar2019jodie} & 3.41\% & 1.02\% & 5.80\% & 51.05\% & 38.71\% \\
LASTFM~\cite{kumar2019jodie} & 4.29\% & 1.14\% & 6.19\% & 44.95\% & 43.43\% \\
GDELT~\cite{zhou2022tgl} & 3.25\% & 8.56\% & 9.34\% & 38.75\% & 40.11\% \\
\bottomrule
\end{tabular}}
\label{tab:breakdown_jodie}
\end{table}

\renewcommand\arraystretch{0.7}
\begin{table}[htbp]
\centering
\caption{Training time breakdown of APAN model}
\resizebox{\linewidth}{!}{
\begin{tabular}{llllll}
\toprule
\textbf{Dataset} & \textbf{Sample} & \textbf{\makecell{Fetch\\feature}} & \textbf{\makecell{Fetch\\memory}} & \textbf{\makecell{Train\\MTGNN}} & \textbf{\makecell{Update\\memory}} \\ 
 \midrule
REDDIT~\cite{kumar2019jodie} & 12.94\% & 5.75\% & 15.18\% & 39.14\% & 27.00\% \\
WIKI~\cite{kumar2019jodie} & 6.52\% & 0.87\% & 9.13\% & 42.61\% & 40.87\% \\
MOOC~\cite{kumar2019jodie} & 10.60\% & 0.83\% & 8.32\% & 45.11\% & 35.14\% \\
LASTFM~\cite{kumar2019jodie} & 11.12\% & 1.02\% & 12.26\% & 41.77\% & 33.83\% \\
GDELT~\cite{zhou2022tgl} & 14.34\% & 3.25\% & 20.31\% & 23.95\% & 38.15\% \\
\bottomrule
\end{tabular}}
\label{tab:breakdown_apan}
\end{table}

\subsection{Overlap the memory update stage with MTGNN training stage}
We have identified an opportunity to overlap the execution of the memory update stage with the MTGNN training stage. Although we have implemented this overlapping, the memory update overhead remains significant, as reported in Table~\ref{tab:breakdown},~\ref{tab:breakdown_jodie}, and~\ref{tab:breakdown_apan}. There are two main reasons for this:

$1.$ The MTGNN training stage cannot fully overlap with the memory update stage due to the dependency on the memory updater for updating the memory within the MTGNN training stage, as discussed in Section~\ref{sec:preliminary}. Additionally, the computational overhead of the memory updater may outweigh that of the embedding modules~\cite{wang2021apan, kumar2019jodie}. Consequently, the available time for the memory update stage to overlap with the MTGNN training stage becomes further limited.

$2.$ The MTGNN training process can be decomposed into three steps: the memory updater computes the updated memory, the MTGNN layer computes the embeddings, and the loss and backward steps are performed (including all-reduce). The latter two stages can indeed be parallelized with the memory update stage, which we have already implemented in our experiments, aligning with TGL~\cite{zhou2022tgl}. However, even with these overlaps, the memory update stage still accounts for up to 31.7\%, 45.0\%, and 40.9\% of the total time, as indicated in Table~\ref{tab:breakdown},~\ref{tab:breakdown_jodie}, and~\ref{tab:breakdown_apan} respectively, making it impossible to completely conceal the associated overhead.

\subsection{Breakdown statistics of JODIE and APAN}
 We provide the training time breakdowns for the JODIE and APAN models in Table~\ref{tab:breakdown_jodie} and Table~\ref{tab:breakdown_apan}, which reveal that memory operations, including memory fetching and updating, can account for up to 50.51\% and 58.56\% of the total training time, respectively. Notably, the significant overhead is primarily due to memory operations rather than the sampling and feature fetching stages, which distinguishes these models from static GNN models and the systems designed for static GNN models.

\subsection{GPU sampler analysis}\label{sec:app_gpu_sample}

\begin{table*}[]
\caption{Detailed training time breakdown of TGN model to illustrate the effect of the GPU sampler.}
\centering
\resizebox{0.7\linewidth}{!}{
\begin{tabular}{llllllll}
\toprule
\textbf{Dataset} & \textbf{Framework} & \textbf{\makecell{Avg\\Epoch(s)}} & \textbf{\makecell{Sample(s)}}  & \textbf{\makecell{Fetch\\feature (s)}} & \textbf{\makecell{Fetch\\memory(s)}} & \textbf{\makecell{Train\\MTGNN(s)}} & \textbf{\makecell{Update\\memory(s)}} \\
 \midrule
\multirow{2}{*}{REDDIT} & TGL & 7.31 & 0.69 & 0.92 & 0.42 & 3.43 & 1.85 \\
 & MSPipe-NoPipe & 7.05 & 0.44 & 0.88 & 0.41 & 3.42 & 1.90 \\
  \midrule
\multirow{2}{*}{WIKI} & TGL & 2.41 & 0.16 & 0.14 & 0.14 & 1.24 & 0.73 \\
 & MSPipe-NoPipe & 2.32 & 0.08 & 0.12 & 0.10 & 1.20 & 0.82 \\
  \midrule
\multirow{2}{*}{MOOC} & TGL & 4.31 & 0.42 & 0.13 & 0.11 & 2.29 & 1.37 \\
 & MSPipe-NoPipe & 4.20 & 0.31 & 0.31 & 0.21 & 2.13 & 1.41 \\
 \midrule
\multirow{2}{*}{LASTFM} & TGL & 13.10 & 1.50 & 1.19 & 1.11 & 5.64 & 3.65 \\
 & MSPipe-NoPipe & 12.64 & 1.04 & 1.20 & 1.05 & 6.12 & 3.23 \\
 \midrule
\multirow{2}{*}{GDELT} & TGL & 645.46 & 113.62 & 82.39 & 67.62 & 242.61 & 139.22 \\
 & MSPipe-NoPipe & 626.09 & 94.26 & 85.20 & 69.21 & 240.99 & 136.43 \\ 
\bottomrule
\end{tabular}}
\label{tab:gpu_sample}
\end{table*}

MSPipe utilizes a GPU sampler for improved resource utilization and faster sampling and we further clarify the remarkable speedup mainly comes from our pipeline mechanism not the GPU sampler. As shown in Table~\ref{tab:gpu_sample}, we conducted a detailed profiling of the sampling time using TGL and found that our sampler is 24.3\% faster than TGL's CPU sampler for 1-hop most recent sampling, which accounts for only 3.6\% of the total training time. Therefore, the performance gain is primarily attributed to our pipeline mechanism and resource-aware minimal staleness schedule but not to the acceleration of the sampler.

\subsection{Why does the memory update stage take longer time than memory fetching?}
The memory update takes a longer time for two reasons: \textbf{1)}In a multi-GPU environment, the memory module is stored in the CPU, allowing multiple GPUs to read simultaneously but not write simultaneously to ensure consistency and avoid conflicts; \textbf{2)} our memory fetching implementation, aligns with TGL, utilizes non-blocking memory copy APIs for efficient transfer of memory vectors from CPU to GPU with pinned memory. However, the lack of a non-blocking API equivalent for \textit{tensor.cpu()} can impact performance.

\section{Implementation details}
\label{sec:app_impl}
\subsection{Algorithm details}
We clarify that $\tau^{(j)}$ is the execution time of different stages, which can be collected in a few iterations of the profiling. The  $\tau^{(j)}$ and the staleness $k_i$ can be pre-calculated for all the graph data, which can be reused for future training. It’s simple and efficient to do the profiling, pre-calculation, and training with our open-source code provided in the anonymous link.

In the case of stages such as the GNN computation stage, the execution time is likely to be dependent on the number of sampled nodes or edges. This quantity not only varies across different batches but also depends on the underlying graph structure. While the training time of a static GNN can differ due to varying numbers of neighbors for each node and the utilization of random sampling, memory-based TGNNs typically employ a fixed-size neighbor sampling approach using the most recent temporal sampler. Specifically, the sampler selects a fixed number of the most recently observed neighbors to construct the subgraph. Consequently, as the timestamp increases, the number of neighboring nodes grows, and it becomes more stable, governed by the maximum number of neighbors per node constraint. Through our profiling analysis, we observed that the number of nodes in the subgraph converges after approximately 10-20 iterations, allowing the average execution time to effectively represent the true execution time.

\subsection{Multi-GPU server implementation}
We have provided a brief description of how MSPipe works in multi-GPU servers at Section~\ref{sec:preliminary} and Section~\ref{sec:3_1} and we have provided the implementation with the anonymous link in the abstract. We will give you a more detailed analysis of the implementation details here:
The graph storage is implemented with NVIDIA UVA so each GPU worker retrieves a local batch of events and performs the sampling process on GPU to generate sub-graphs. The memory module is stored in the CPU's main memory without replication to ensure consistency and exhibit the ability to store large graphs. Noted that, except for the GPU sample, the other stages align with TGL.
Here is a step-by-step overview:
\begin{enumerate}[leftmargin=*]
    \item Each GPU worker retrieves a local batch of events and performs the sampling process on the GPU to generate sub-graphs.
    \item Fetches the required features and node memory vectors from the CPU to the GPU for the subgraphs.
    \item Performs MTGNN forward and backward computations on each GPU. MSPipe implements Data Parallel training similar to TGL.
    \item The memory module is stored in the CPU's main memory without replication to ensure consistency. Each GPU transfers the updated memory vectors to the CPU and updates the corresponding elements, which ensures that the memory module remains consistent across all GPUs.
\end{enumerate}

\subsection{Stall-free minimal staleness scheduling} \label{sec:app_algo}
We propose a resource-aware online scheduling algorithm to decide the starting time of stages in each training iteration, as given in Algorithm~\ref{alg:online} 
\begin{algorithm}[H]
    \caption{Online Scheduling for MTGNN training pipeline} 
    \begin{algorithmic}[1]
    \small
        \State {\bfseries Input:} $E$ batches of events $\mathcal{B}_i$, Graph $\mathcal{G}$, minimum staleness iteration number $k_i$
        \State {\bfseries Global:} $i_{\text{upd}} \gets 0$\Comment{
       the latest iteration whose memory update is done}
        \For{$i \in {1, 2, ..., E}$ in parallel}
            \If{$lock(sample\_lock)$}
                \State $\mathcal G_{\text{sub}} \gets Sample(\mathcal G, \mathcal B_i)$\Comment{sample subgraph $\mathcal G_{\text{sub}}$ using a batch of events}
            \EndIf
            \If{$lock(feature\_lock \And pcie\_lock)$}
                \State $fetch\_feature(\mathcal G_{\text{sub}})$\Comment{feature fetching for the subgraphs}
            \EndIf
            \If{$lock(memory\_lock \And pcie\_lock$)}
                \While{$i - i_{\text{upd}} > k_i$ }
                    \State $wait()$ \Comment{delay memory fetching until staleness iteration number is smaller than $k_i$}
                \EndWhile
                \State $fetch\_memory(\mathcal G_{\text{sub}})$  \Comment{transfer  memory vectors for the subgraphs}
            \EndIf
            \If{$lock(gnn\_lock)$}
                \State $MTGNN(\mathcal G_{\text{sub}})$ \Comment{train the MTGNN model using the subgraphs}
            \EndIf
            \If{$lock(update\_lock)$}
                \State $update\_mem(\mathcal G_{\text{sub}}, \mathcal B_i)$ \Comment{generate new memory vectors and write back to CPU storage}
                \State $i_{\text{upd}} \gets i$ \Comment{update the last iteration with memory update done}
            \EndIf         
        \EndFor
    \end{algorithmic}
    \label{alg:online}
    \vspace{-1pt}
\end{algorithm}
To enable asynchronous and parallel execution of the stages, we utilize a thread pool and a CUDA stream pool. Each batch of data is assigned an exclusive thread and stream from the respective pools, enabling concurrent processing of multiple batches. Dedicated locks for each stage are used to resolve resource contention and enforce sequential execution (Equation~\ref{eq:start}). Figure~\ref{fig:schedule} provides a schematic illustration of our online scheduling. The schedule of the memory fetching stage ensures the minimal staleness iteration requirement (Lines 8-11). As illustrated in Figure~\ref{fig:schedule}, the scheduling effectively fills the bubble time while minimizing staleness and avoiding resource competence. At the end of each training iteration, new memory vectors are generated based on the staled historical memories and events in the current batch (Line 15). Finally, the latest iteration whose memory update stage has been completed is recorded, enabling other parallel threads that run other training iterations to track (Line 16). Note that the first few iterations before iteration $k$ will act as a warmup, which means they will not wait for the memory update $k$ iterations before.
\section{Full experiments} \label{sec:app_c}
We first provide the details of the experiments and discuss the experiment setting. Then we provide the full version of the experiment results, including the accuracy and throughput speedup, the convergence of the JODIE and APAN model, the distribution of $\Delta t$ in remaining datasets, and the analysis of the node memory similarity.

\subsection{Details of the Experiments}
\textbf{Datasets.} This paper employs several datasets, each with its unique properties and characteristics. The Reddit dataset captures the posting behavior of users on subreddits over one month, and the link feature is extracted through the conversion of post text into a feature vector. The Wikipedia dataset records the editing behavior of users on Wikipedia pages over a month, and the link feature is extracted through the conversion of the edit text into a 172-dimensional Linguistic Inquiry and Word Count (LIWC) feature vector. The MOOC dataset captures the online learning behavior of students in a MOOC course while
the LastFM dataset contains information about which songs were listened to by which users over one month. The GDELT dataset is a Temporal Knowledge Graph that records global events in multiple languages every 15 minutes, which covers events from 2016 to 2020 and consists of homogeneous dynamic graphs with nodes representing actors and temporal edges representing point-time events. Furthermore, it is important to highlight that \textit{TGNN training employs graph edges as training samples}, in contrast to static GNN training, which utilizes nodes as training samples. All the datasets are downloaded from the \href{https://github.com/amazon-science/tgl/blob/main/down.sh}{link} in TGL~\cite{zhou2022tgl} repository.

\subsection{Full version of the Experiment Results} \label{sec:app_c3_fullexp}
\subsubsection{The superior AP in LastFM} 

The reasons why our staleness mitigation strategy outperforms the AP of the baseline TGL in the LastFM dataset is due to the unique characteristics of the LastFM datasets:

$\bullet$ The LastFM dataset exhibits a larger average time gap ($\frac{t_{max} - t_{min}}{E}$, where $t_{max}$ and $t_{min}$)represent the largest and smallest timestamps, respectively, and $E$ denotes the number of events) compared to other datasets, as discussed by ~\cite{cong2023we}. Specifically, LastFM has an average time gap of 106, whereas Reddit's average time gap is 4, Wiki's average time gap is 17, MOOC's average time gap is 3.6, and GDELT's average time gap is 0.1.

$\bullet$ Consequently, even without staleness in the baseline method, the node memory in the LastFM graph tends to become significantly outdated~\cite{rossi2021tgn}, as discussed in Section 3.3. Our staleness mitigation strategy eliminates the outdated node representation by aggregating the memories of the recently active nodes with the highest similarity. This approach helps mitigate the impact of the large time gap present in LastFM datasets, ultimately leading to an improvement in AP compared to the baseline methods.

\subsubsection{Scalability results on all datasets} \label{sec:app_scalability}
\begin{figure*}[htbp]
\vspace{-2pt}
\subfigure[REDDIT]{
\begin{minipage}[t]{0.19\linewidth}
\includegraphics[width=\linewidth]{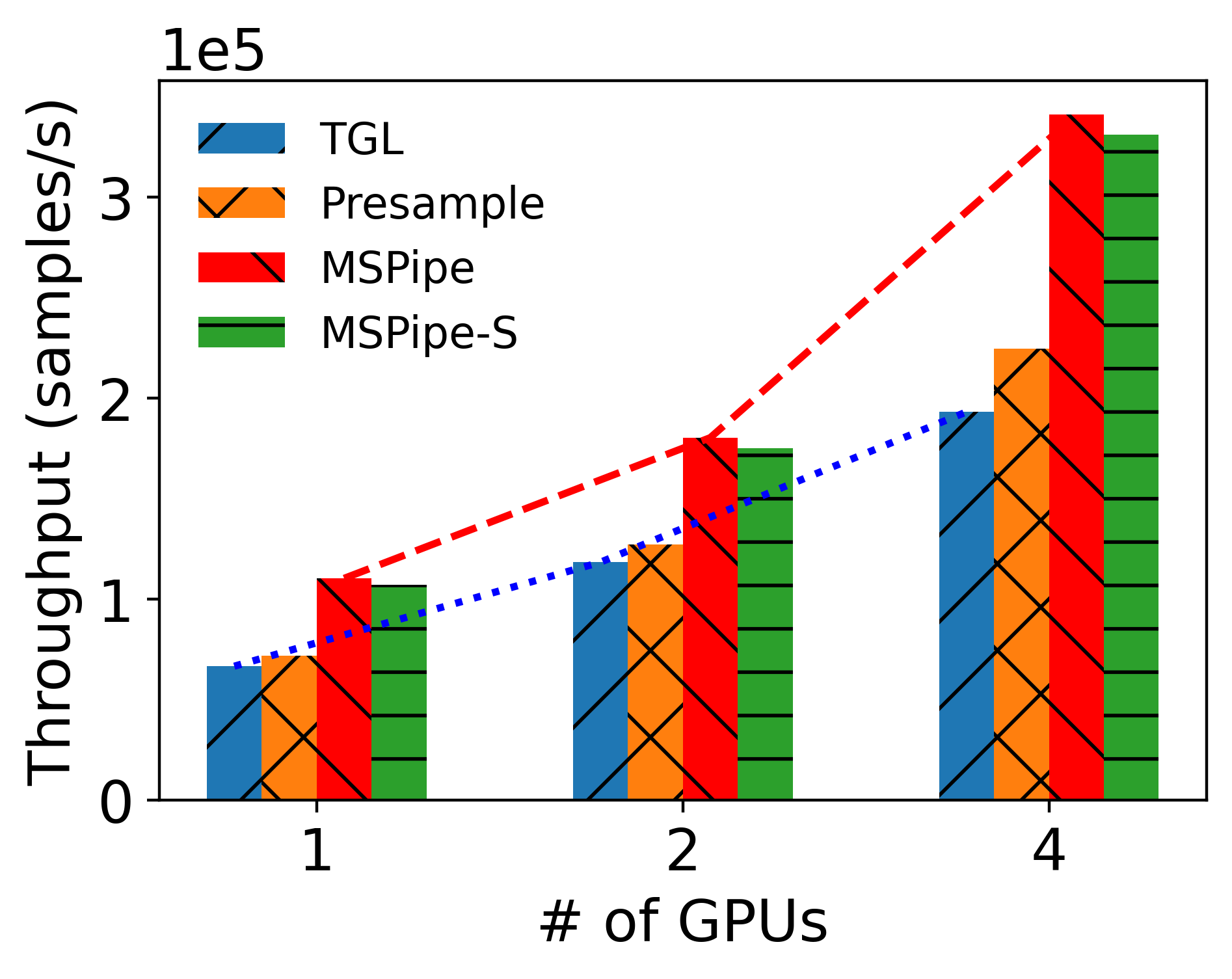}
\end{minipage}%
}%
\subfigure[WIKI]{
\begin{minipage}[t]{0.19\linewidth}
\includegraphics[width=\linewidth]{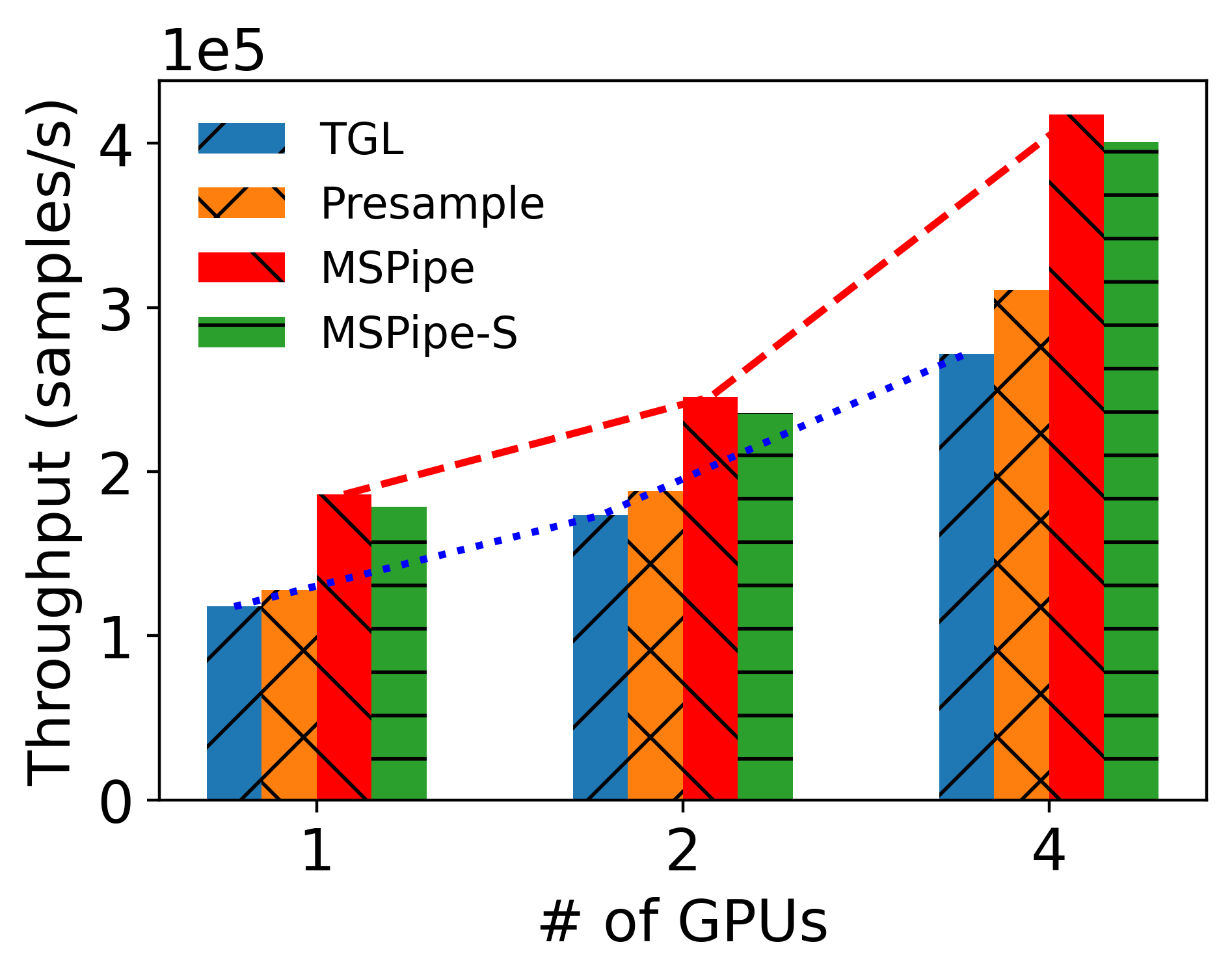}
\end{minipage}%
}%
\subfigure[MOOC]{
\begin{minipage}[t]{0.19\linewidth}
\includegraphics[width=\linewidth]{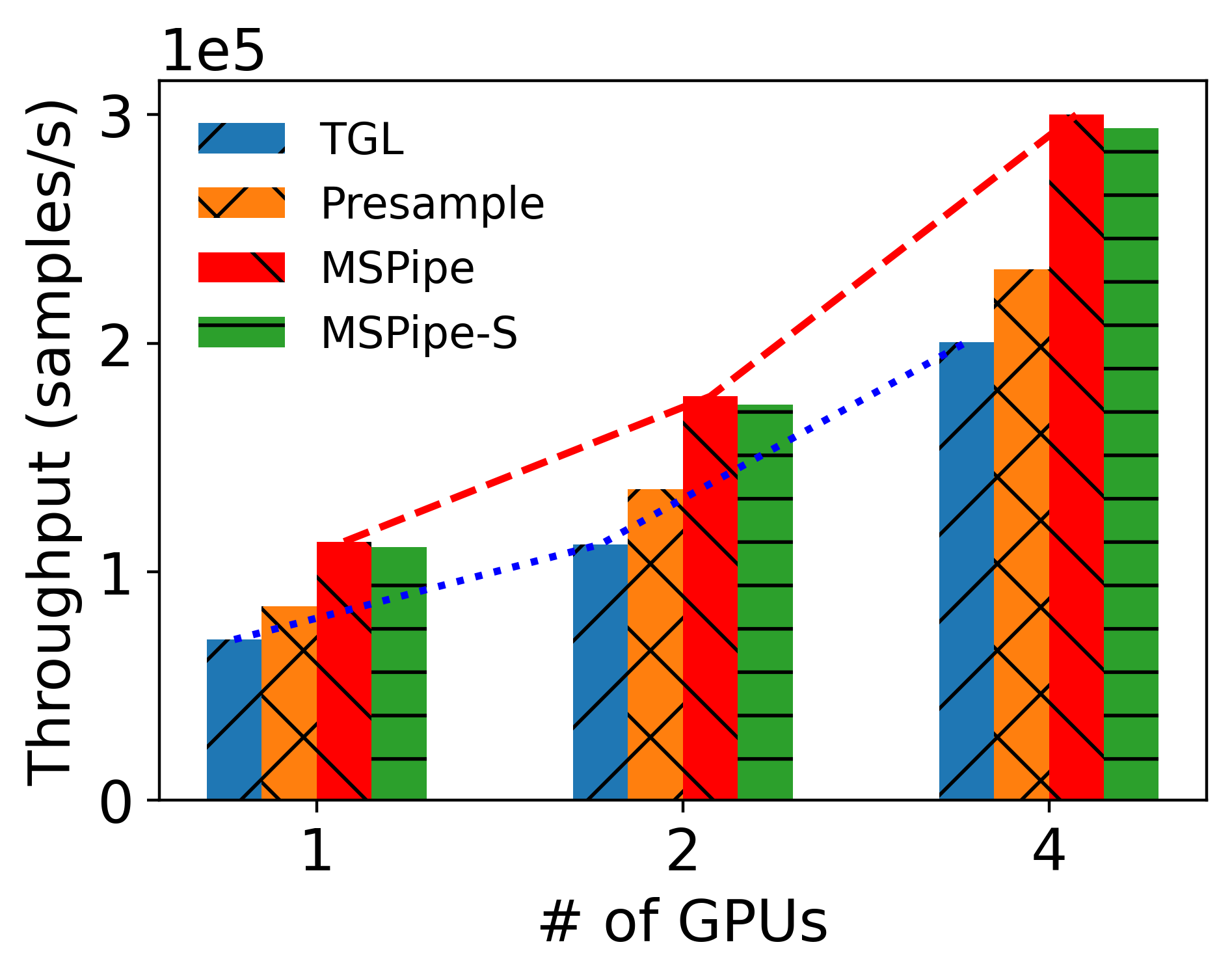}
\end{minipage}%
}%
\subfigure[LASTFM]{
\begin{minipage}[t]{0.19\linewidth}
\includegraphics[width=\linewidth]{figs/tgn_lastfm_scale_new.png}
\end{minipage}
}%
\subfigure[GDELT]{
\begin{minipage}[t]{0.19\linewidth}
\includegraphics[width=\linewidth]{figs/tgn_gdelt_scale_new.png}
\label{fig:app_scale_gdelt}
\end{minipage}
}%
\vspace{-3mm}
\caption{Scalability of training TGN.}
\label{fig:app_scalability}
\vspace{-3mm}
\end{figure*}

We further provide the full scalability results discussed in Section~\ref{sec:exp_expedited}. We show the training throughput with different numbers of GPUs of TGN models on five datasets in Figure~\ref{fig:app_scalability}. MSPipe not only achieves consistent speed-up but also demonstrates remarkable scaling efficiency, reaching up to 83.6\% on a single machine. Scaling efficiency is computed as the ratio of the speed-up achieved by utilizing 4 GPUs to the ideal speed-up. These results surpass those of other baseline methods. Furthermore, when scaling TGN training on GDELT to two machines equipped with eight GPUs (as shown in Figure~\ref{fig:app_scale_gdelt}), MSPipe continues to outperform the baselines and exhibits superior scalability, even without explicit optimization for inter-machine communication.

\begin{figure*}[t]
\subfigure[REDDIT]{
\begin{minipage}[t]{0.25\linewidth}
\includegraphics[scale=0.125]{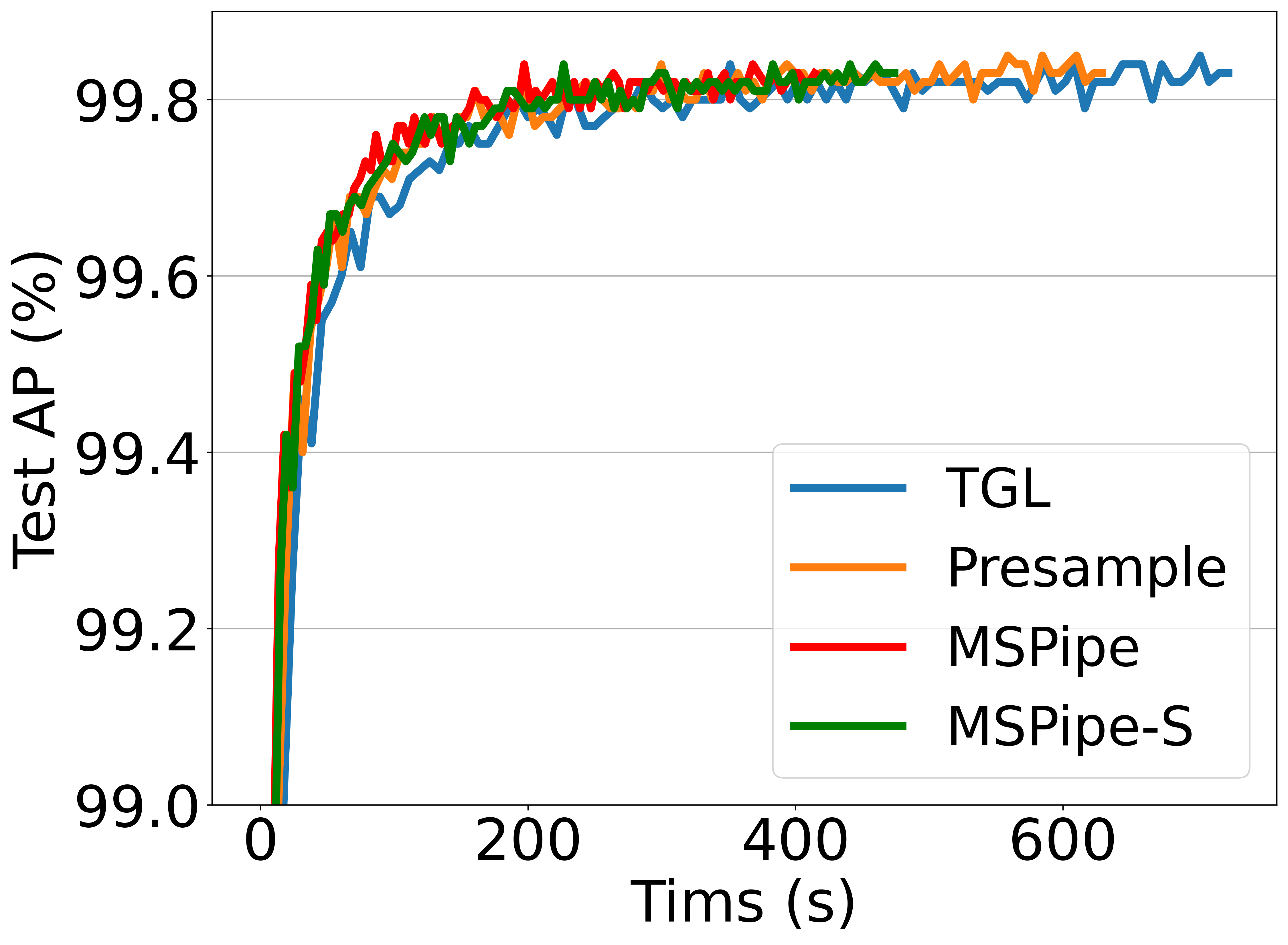}
\end{minipage}%
}%
\subfigure[WIKI]{
\begin{minipage}[t]{0.25\linewidth}
\includegraphics[scale=0.125]{figs/tgn_wiki_conv_ts_thin.png}
\end{minipage}%
}%
\subfigure[MOOC]{
\begin{minipage}[t]{0.25\linewidth}
\includegraphics[scale=0.125]{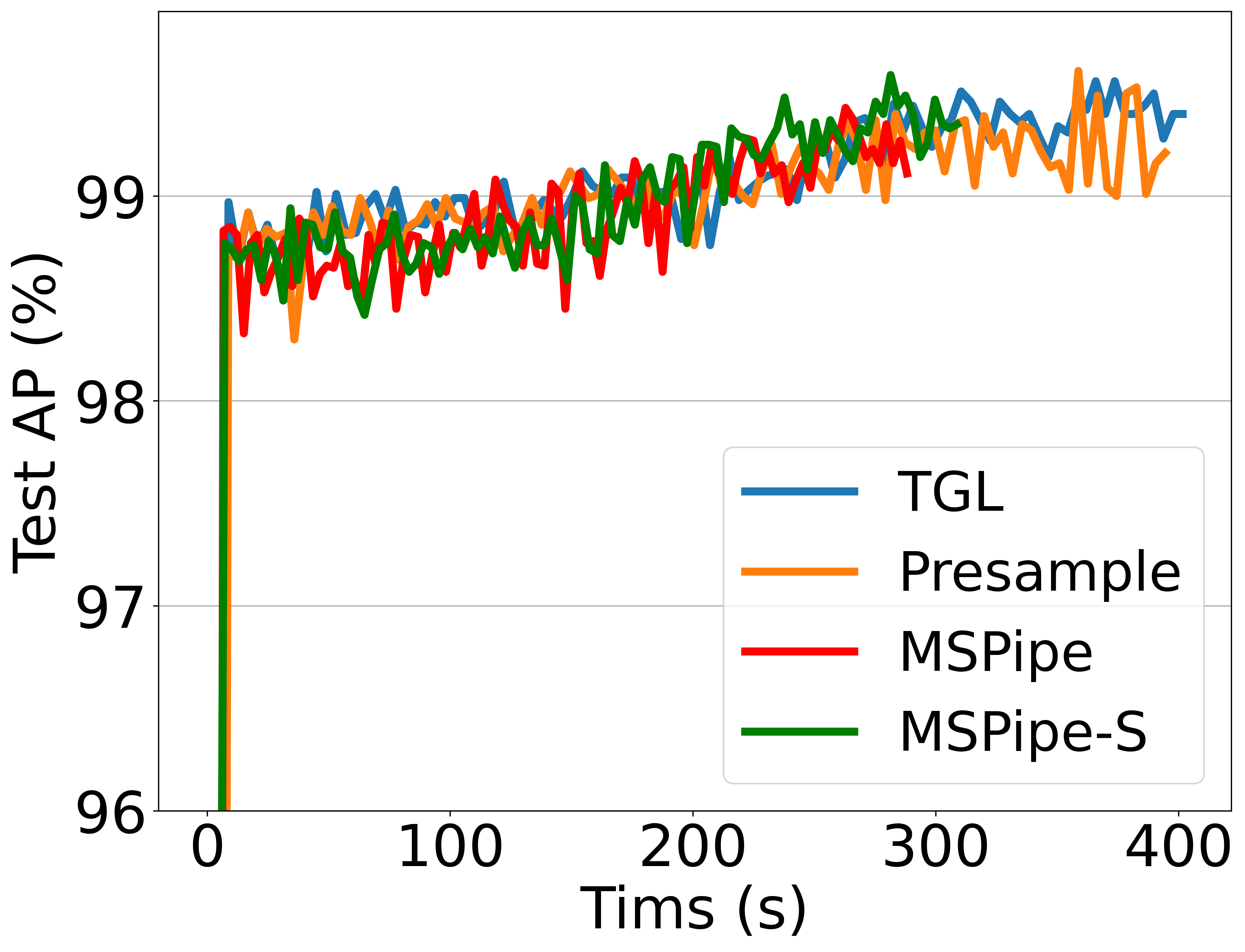}
\end{minipage}
}%
\subfigure[LASTFM]{
\begin{minipage}[t]{0.25\linewidth}
\includegraphics[scale=0.125]{figs/tgn_lastfm100_conv_ts_thin.png}
\end{minipage}
}%
\vspace{-3mm}
\caption{Convergence of TGN training. x-axis is the wall-clock training time, and y-axis is the test average precision.}
\label{fig:converge_tgn}
\end{figure*}

\begin{figure*}[htbp]
\vspace{-2pt}
\subfigure[REDDIT]{
\begin{minipage}[t]{0.25\linewidth}
\includegraphics[scale=0.125]{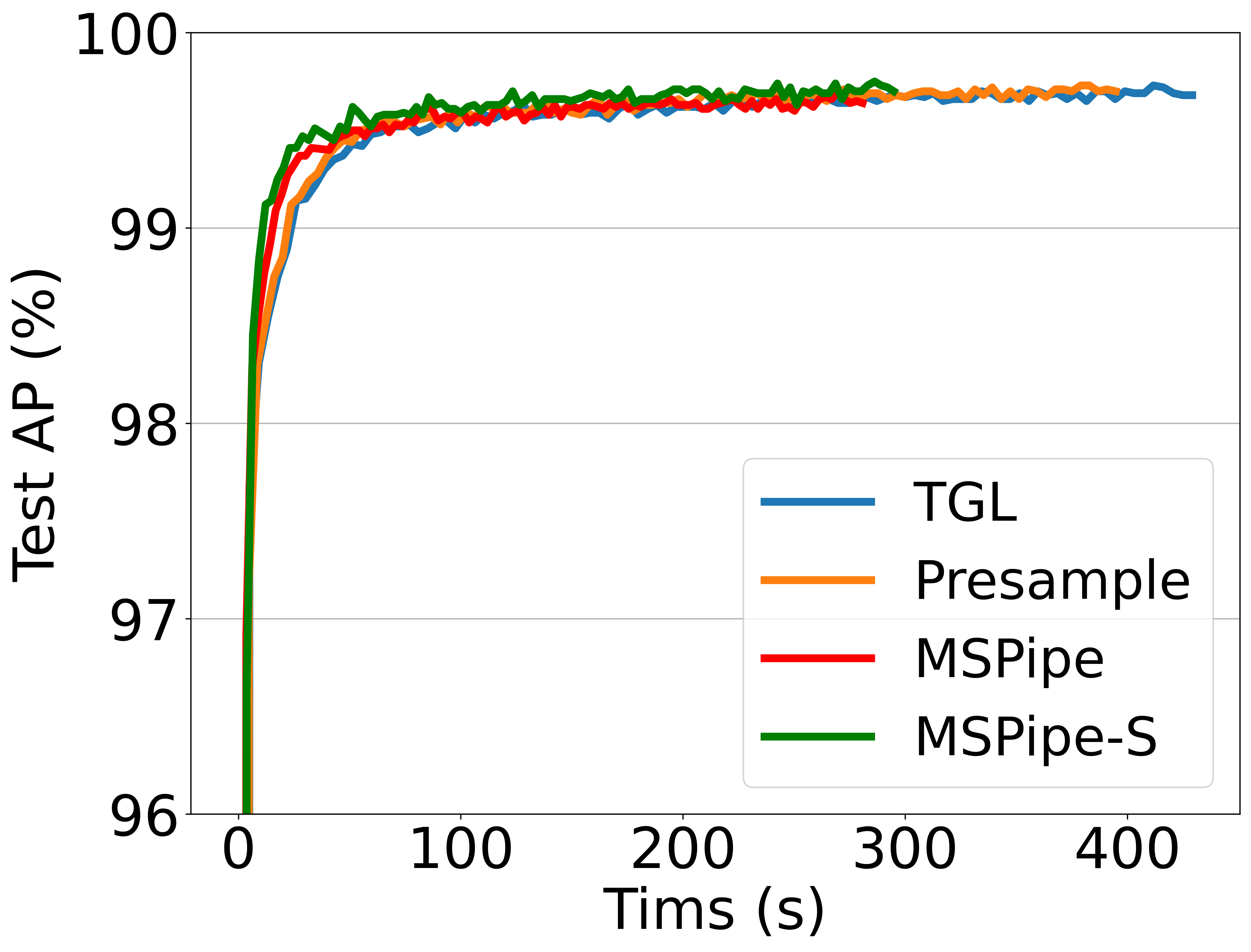}
\end{minipage}%
}%
\subfigure[WIKI]{
\begin{minipage}[t]{0.25\linewidth}
\includegraphics[scale=0.125]{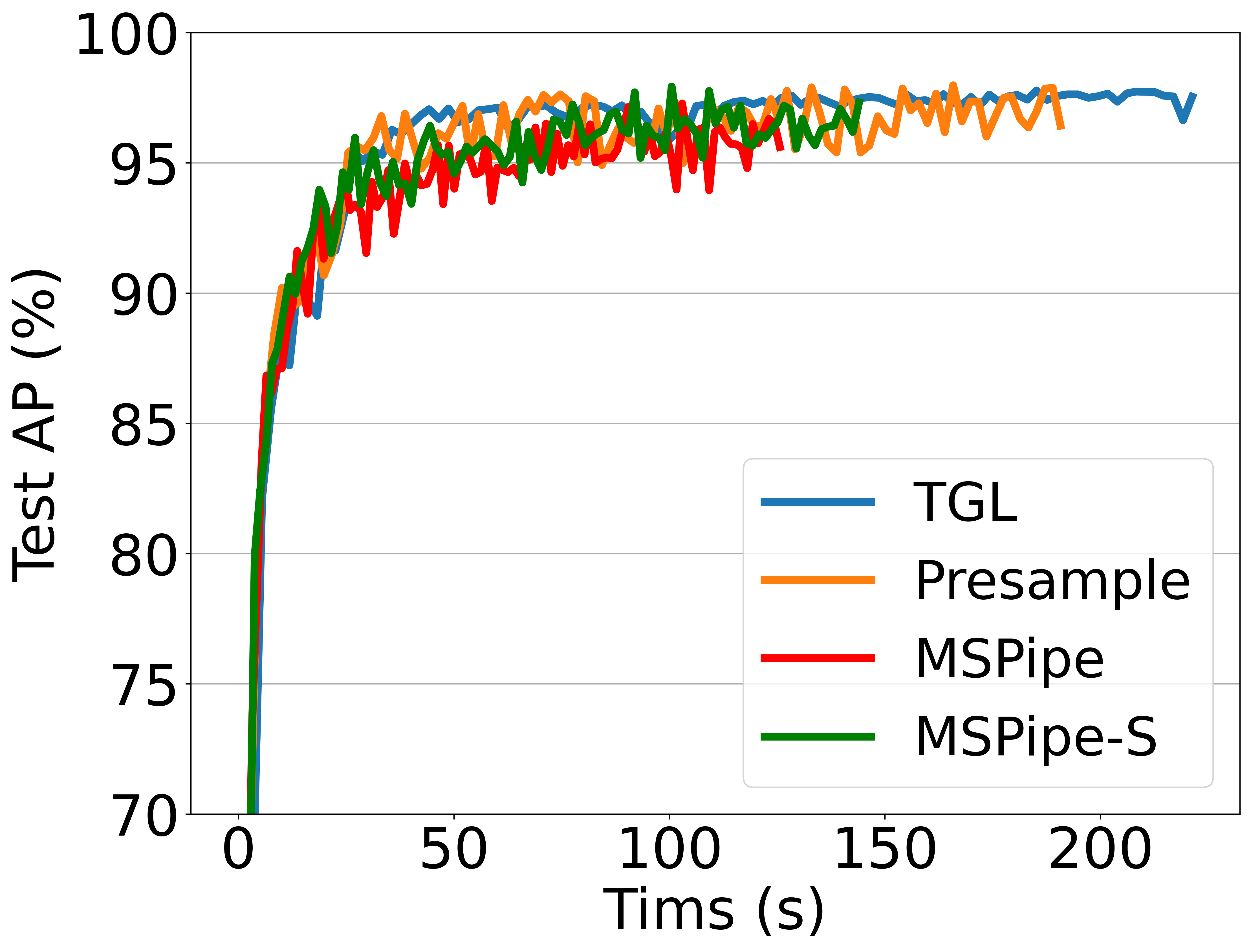}
\end{minipage}%
}%
\subfigure[MOOC]{
\begin{minipage}[t]{0.25\linewidth}
\includegraphics[scale=0.125]{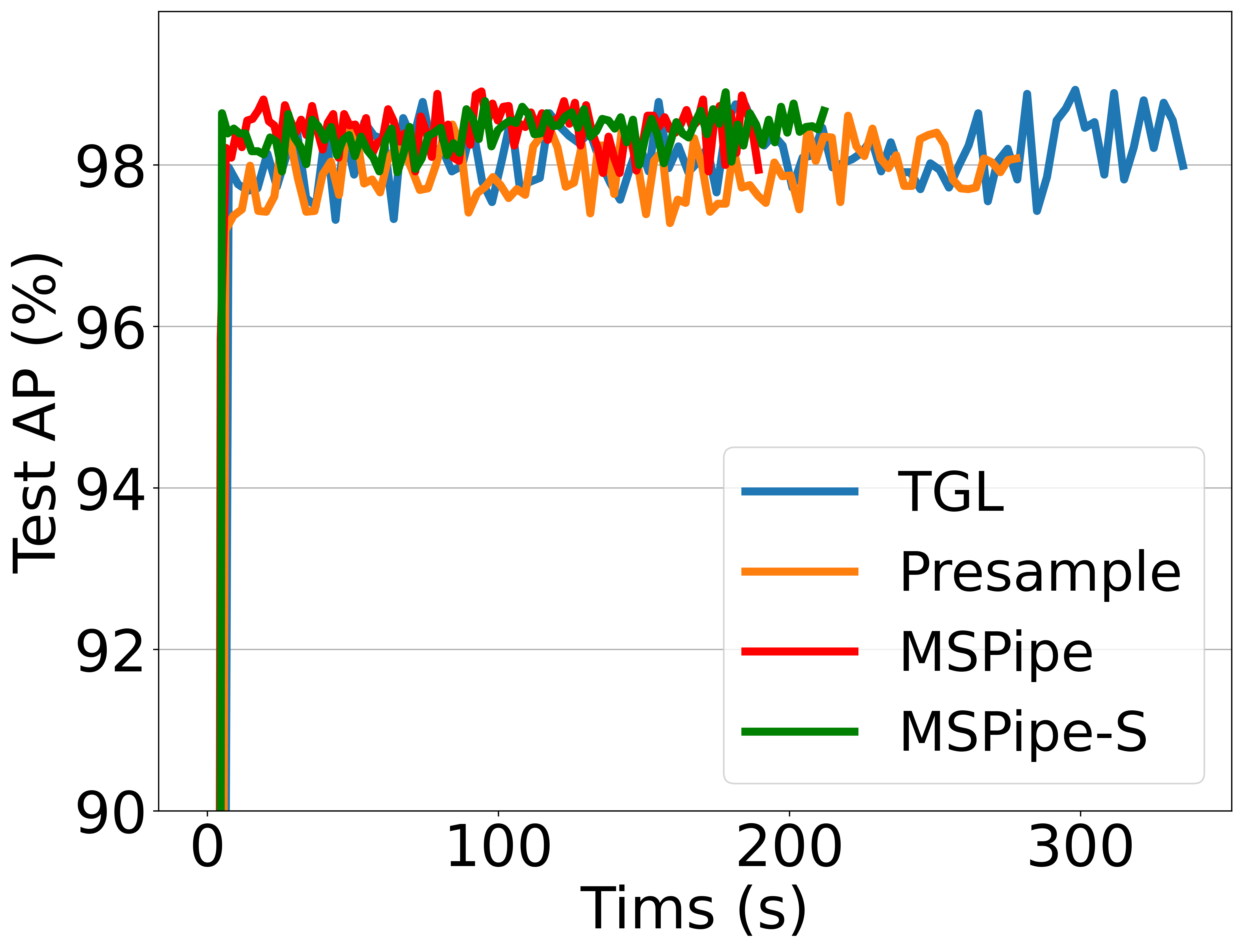}
\end{minipage}
}%
\subfigure[LASTFM]{
\begin{minipage}[t]{0.25\linewidth}
\includegraphics[scale=0.125]{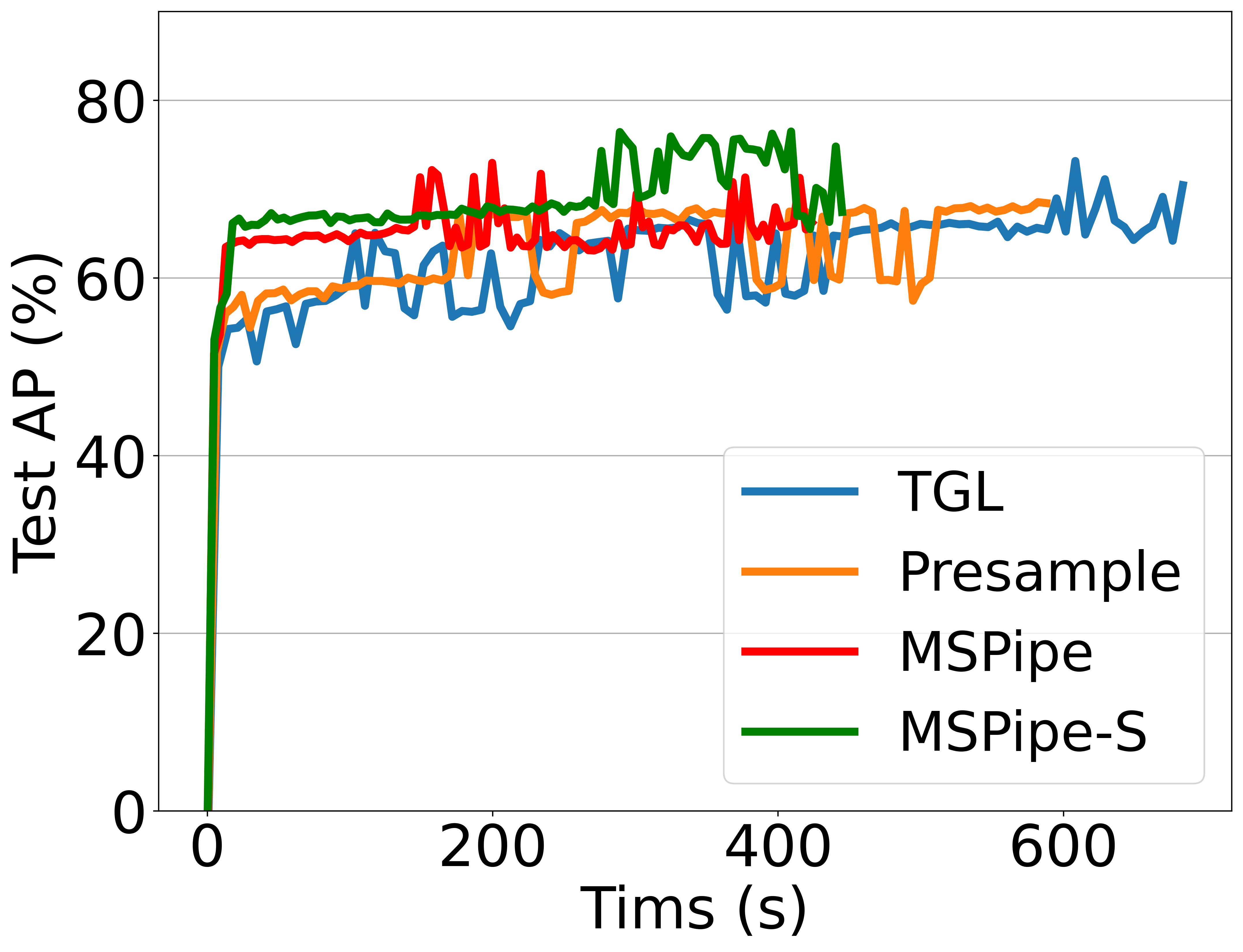}
\end{minipage}
}%
\vspace{-3mm}
\caption{Convergence of JODIE training. the x-axis is the wall-clock training time, and the y-axis is the test average precision}
\label{fig:converge_jodie}
\end{figure*}

\begin{figure*}[htbp]
\vspace{-2pt}
\subfigure[REDDIT]{
\begin{minipage}[t]{0.25\linewidth}
\includegraphics[scale=0.125]{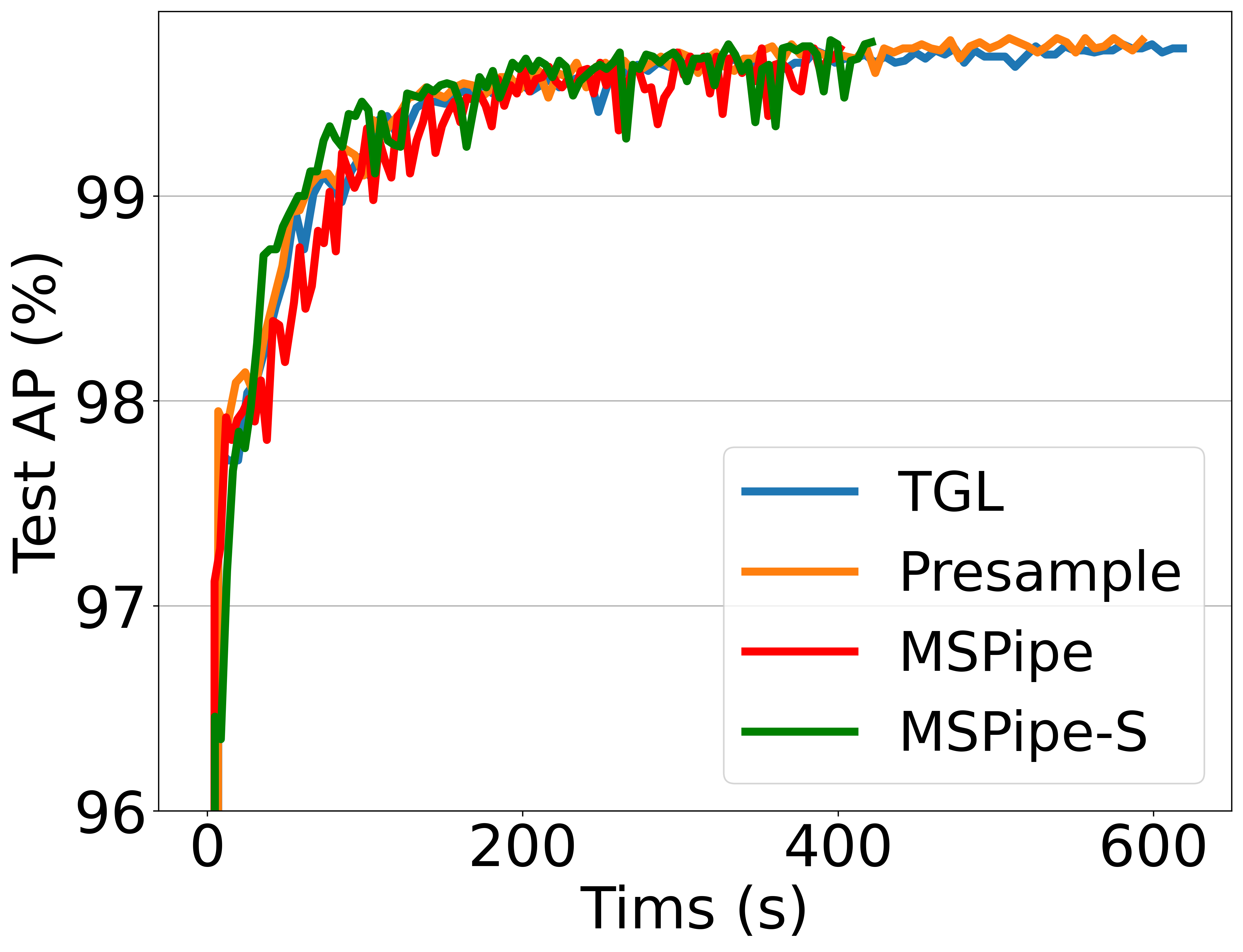}
\end{minipage}%
}%
\subfigure[WIKI]{
\begin{minipage}[t]{0.25\linewidth}
\includegraphics[scale=0.125]{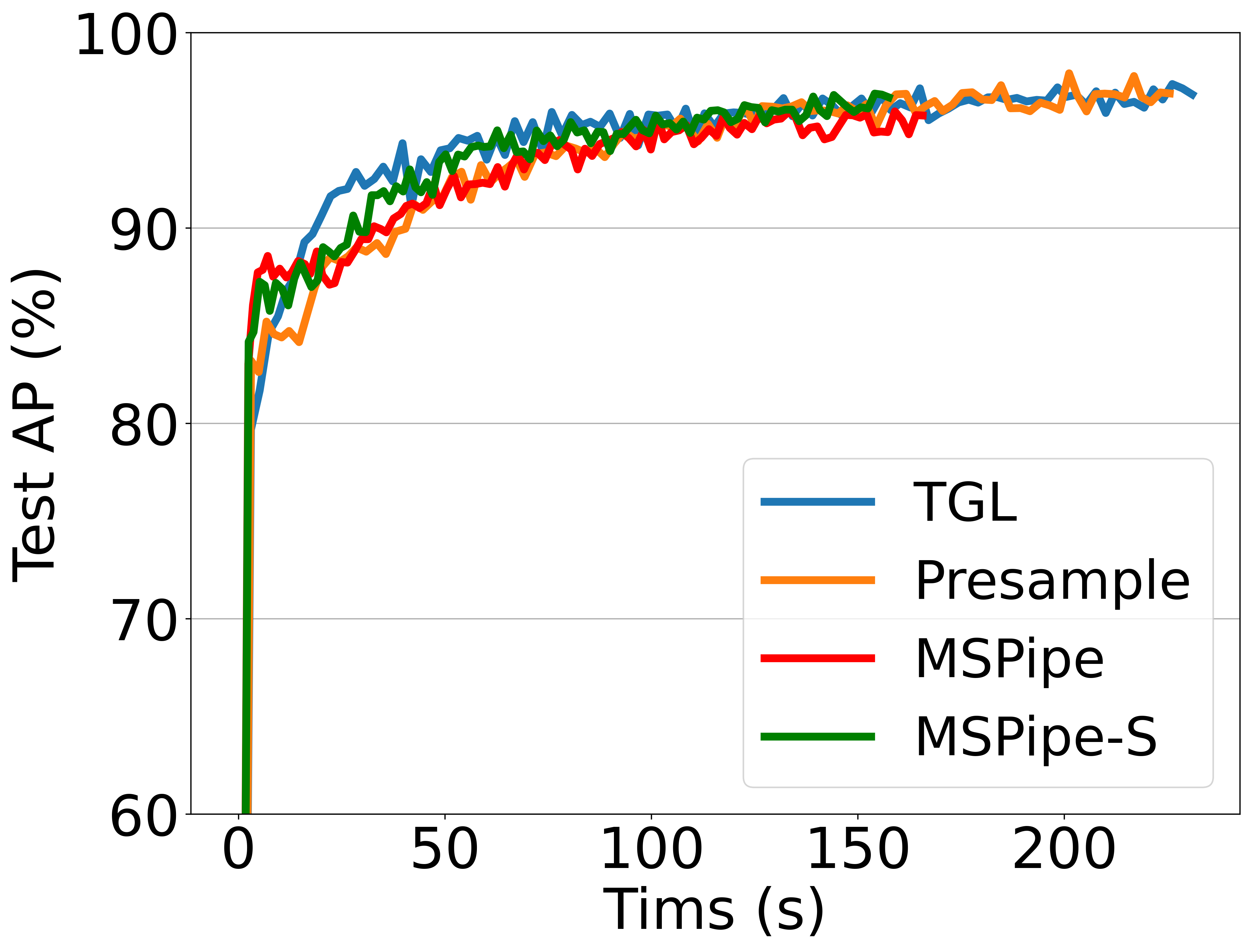}
\end{minipage}%
}%
\subfigure[MOOC]{
\begin{minipage}[t]{0.25\linewidth}
\includegraphics[scale=0.125]{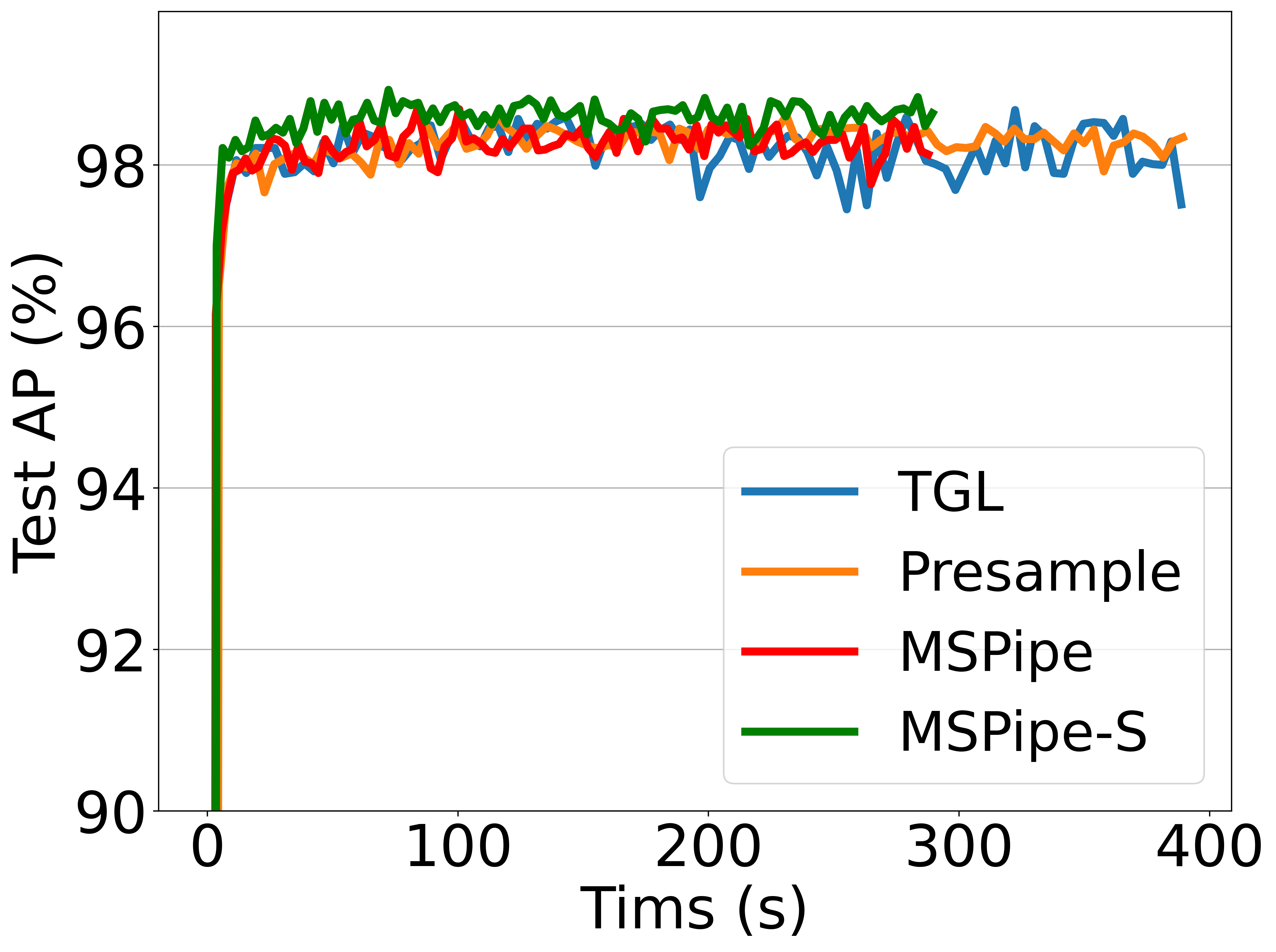}
\end{minipage}
}%
\subfigure[LASTFM]{
\begin{minipage}[t]{0.25\linewidth}
\includegraphics[scale=0.125]{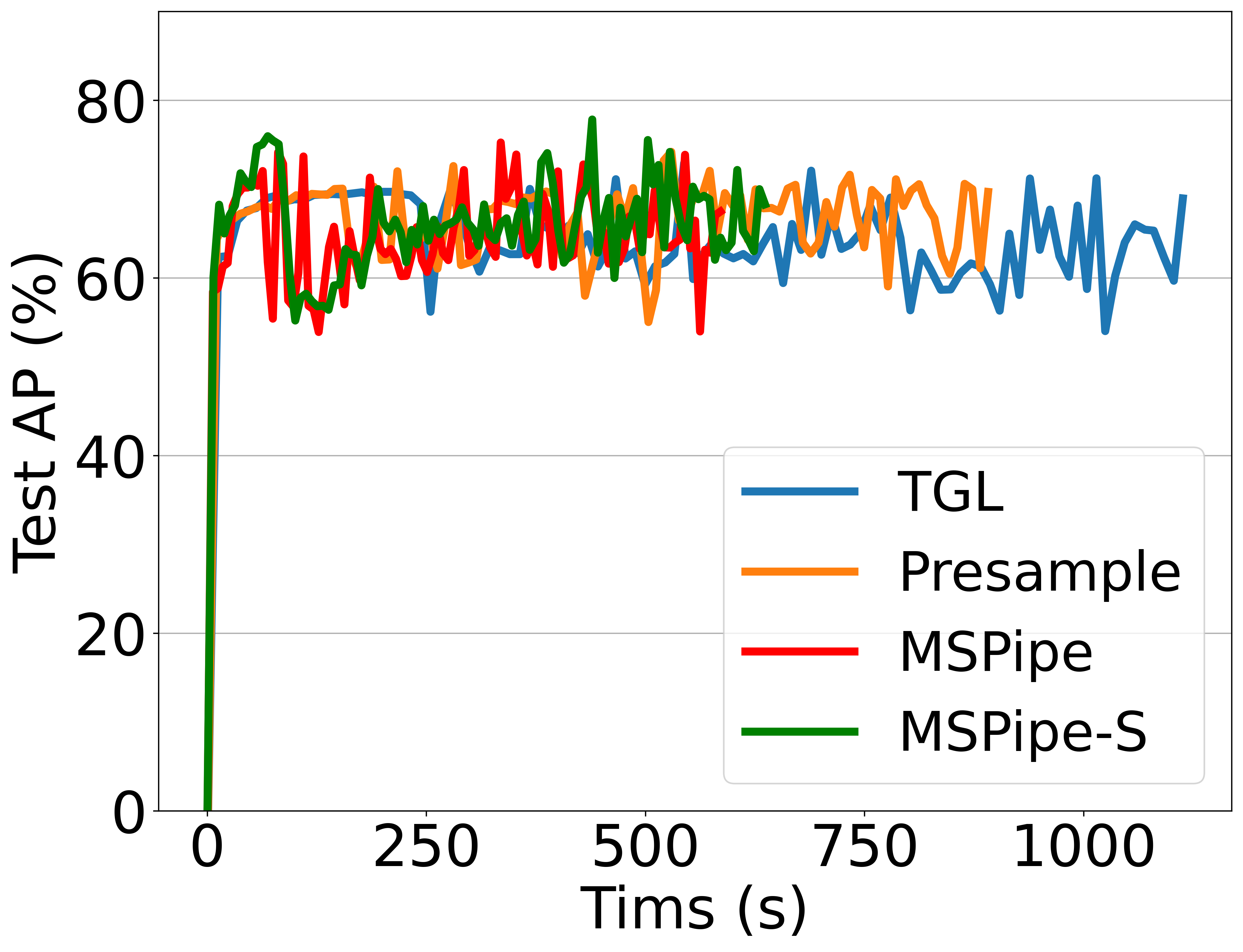}
\end{minipage}
}%
\vspace{-3mm}
\caption{Convergence of APAN training. the x-axis is the wall-clock training time, and the y-axis is the test average pricision}
\label{fig:converge_apan}
\end{figure*}

\begin{figure*}[htbp]
\vspace{-2pt}
\subfigure[REDDIT]{
\begin{minipage}[t]{0.25\linewidth}
\includegraphics[scale=0.23]{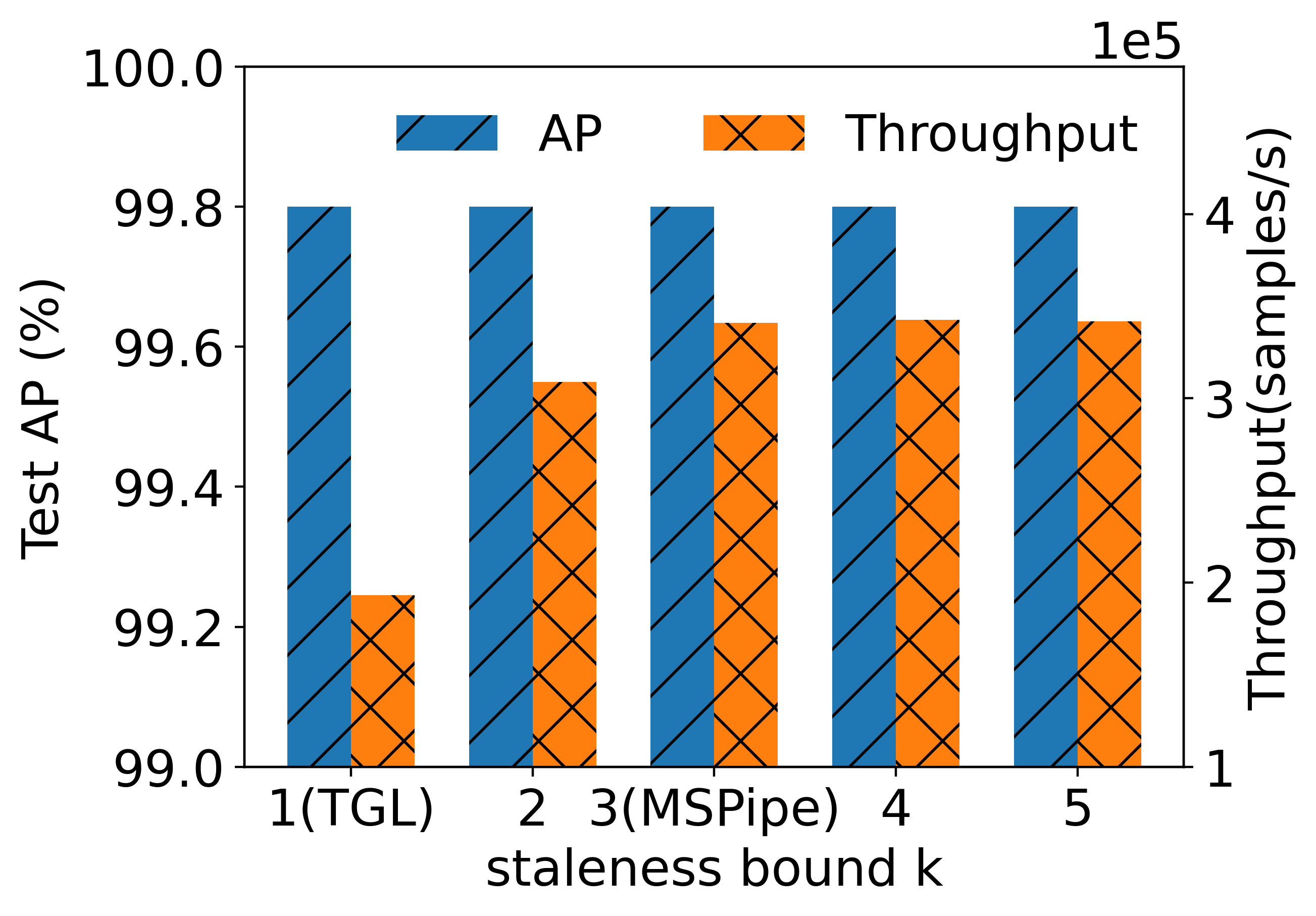}
\end{minipage}%
}%
\subfigure[WIKI]{
\begin{minipage}[t]{0.25\linewidth}
\includegraphics[scale=0.23]{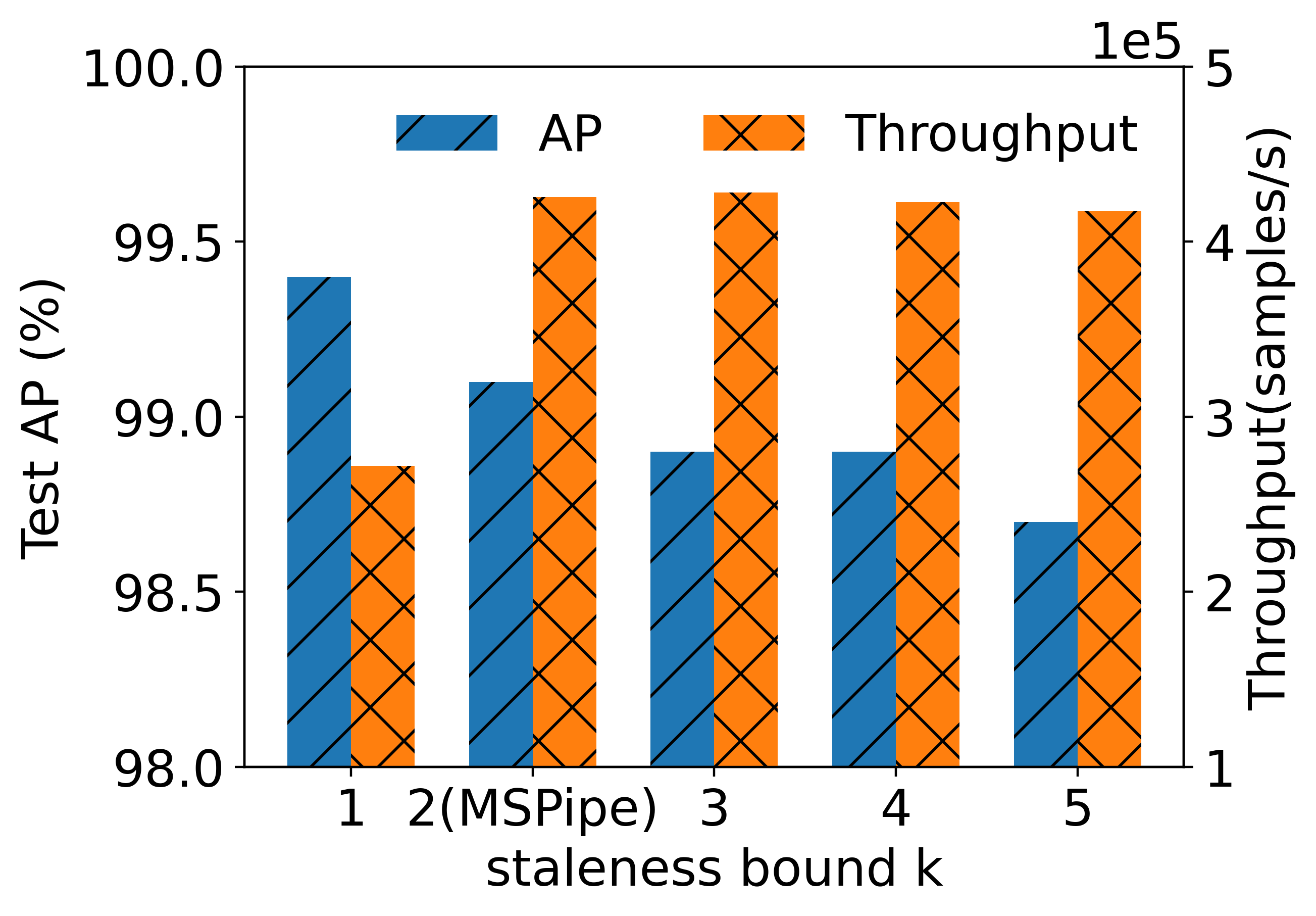}
\end{minipage}%
}%
\subfigure[LASTFM]{
\begin{minipage}[t]{0.25\linewidth}
\includegraphics[scale=0.23]{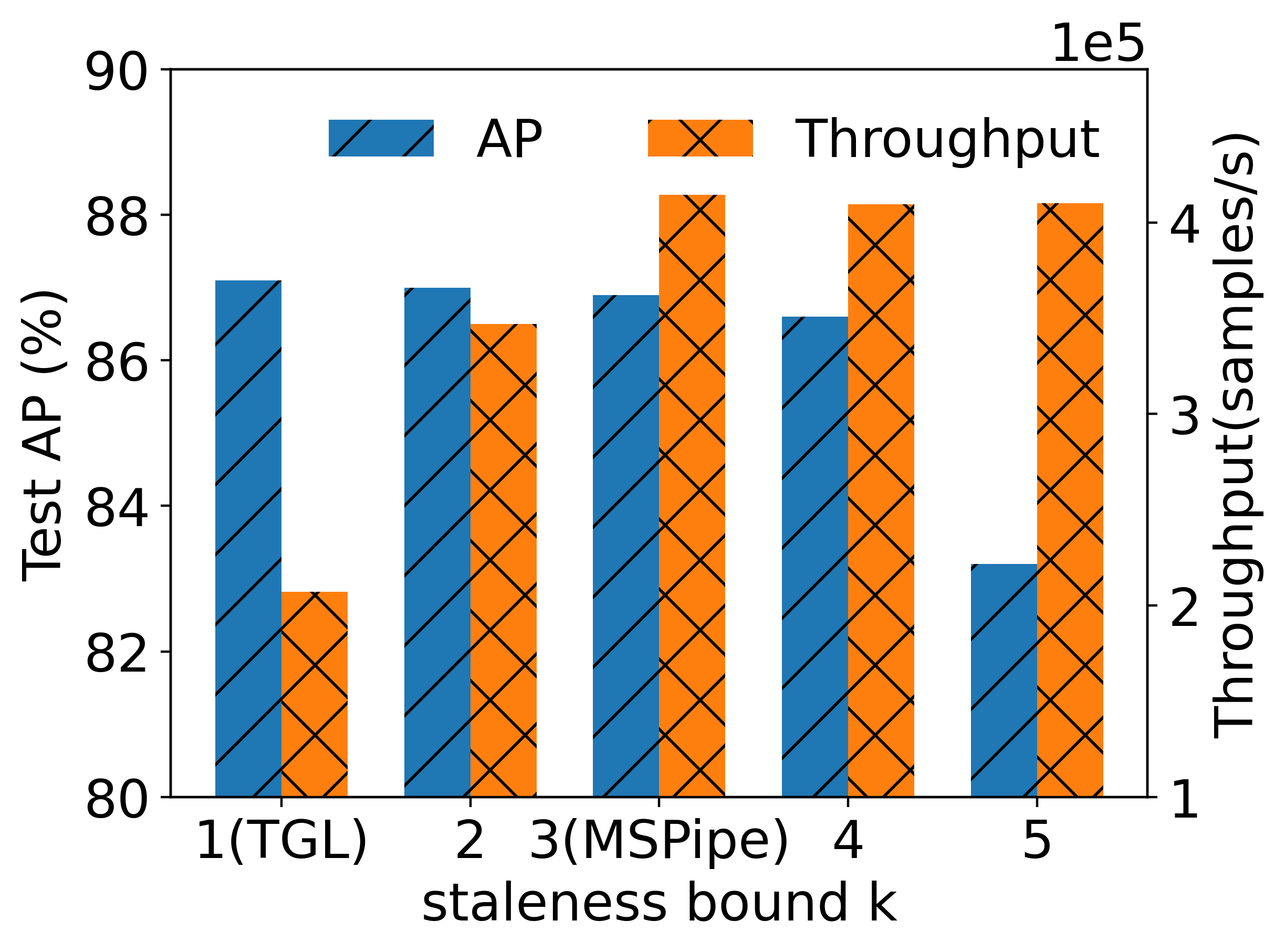}
\end{minipage}
}%
\subfigure[GDELT]{
\begin{minipage}[t]{0.25\linewidth}
\includegraphics[scale=0.23]{figs/gdelt_staleness_new.png}
\end{minipage}
}%
\caption{Staleness error comparison on TGN. MOOC and GDELT datasets are presented in Figure~\ref{fig:optimal_staleness}}
\label{fig:app_staleness_error}
\end{figure*}

\subsubsection{Convergence of the TGN, JODIE and APAN} \label{sec:app_convergence_result}
We further provide the full results discussed in Section~\ref{sec:exp_con_rate}. We show the convergence of TGN, JODIE, and APAN models on five datasets in Figure~\ref{fig:converge_tgn}, Figure~\ref{fig:converge_jodie} and Figure~\ref{fig:converge_apan}. We can see that the training curves of all models largely overlap with the baselines (TGL and Presample), demonstrating that MSPipe preserves the convergence rate. Notably, MSPipe-S achieves better performance than the other variants on the WIKI and LastFM datasets.

\subsubsection{Comparison between different staleness bound} \label{sec:app_staleness_error}
Furthermore, we present a comprehensive comparison of various staleness bounds across multiple datasets including REDDIT, WIKI, LASTFM, and GDELT, using the TGN model, in order to validate the efficacy of MSPipe. The results consistently demonstrate that MSPipe outperforms other staleness-bound options in terms of both throughput and accuracy across all datasets. As shown in Fig~\ref{fig:stale_count}, the number of staleness $k_i$ will soon converge to a steady minimal staleness value. To represent this minimal staleness bound, we utilize a fixed value that corresponds to the steady state. This choice allows us to showcase the minimal staleness bound effectively.

\begin{figure*}[t]
\vspace{-2pt}
\subfigure[REDDIT]{
\begin{minipage}[t]{0.25\linewidth}
\centering
        \includegraphics[width=0.90\linewidth]{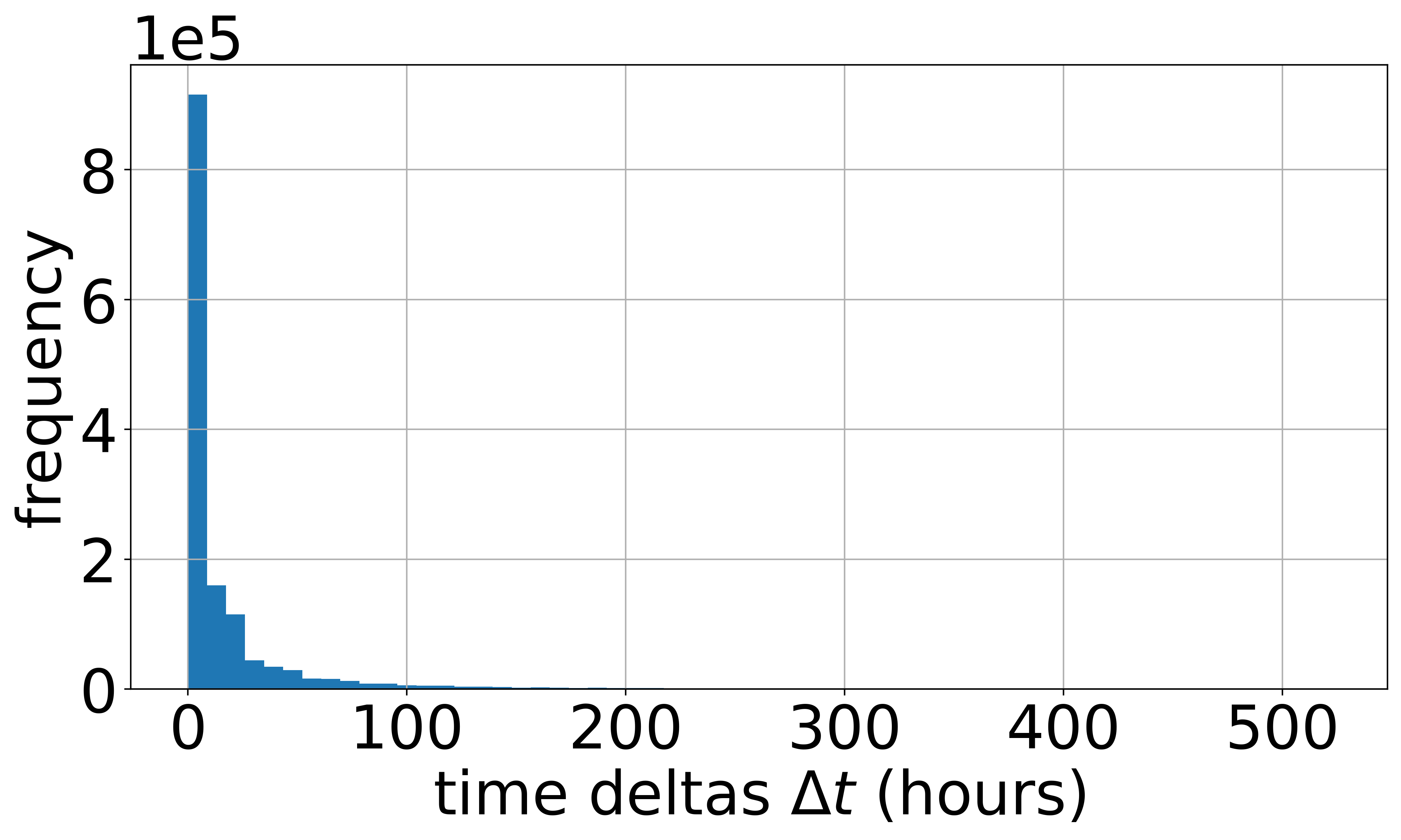}
\end{minipage}%
}%
\subfigure[MOOC]{
\begin{minipage}[t]{0.25\linewidth}
\centering
        \includegraphics[width=0.90\linewidth]{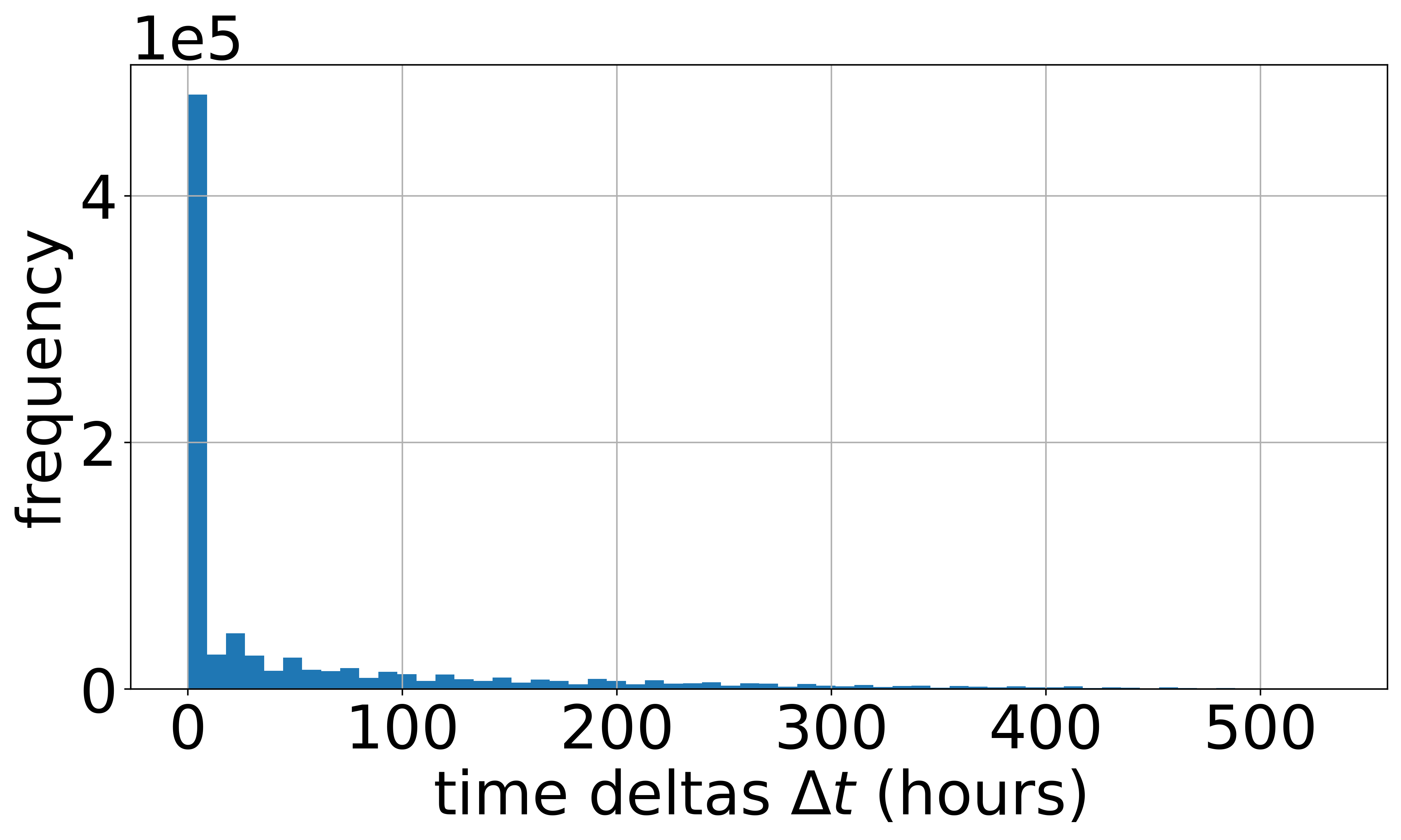}
\end{minipage}%
}%
\subfigure[LASTFM]{
\begin{minipage}[t]{0.25\linewidth}
\includegraphics[width=0.90\linewidth]{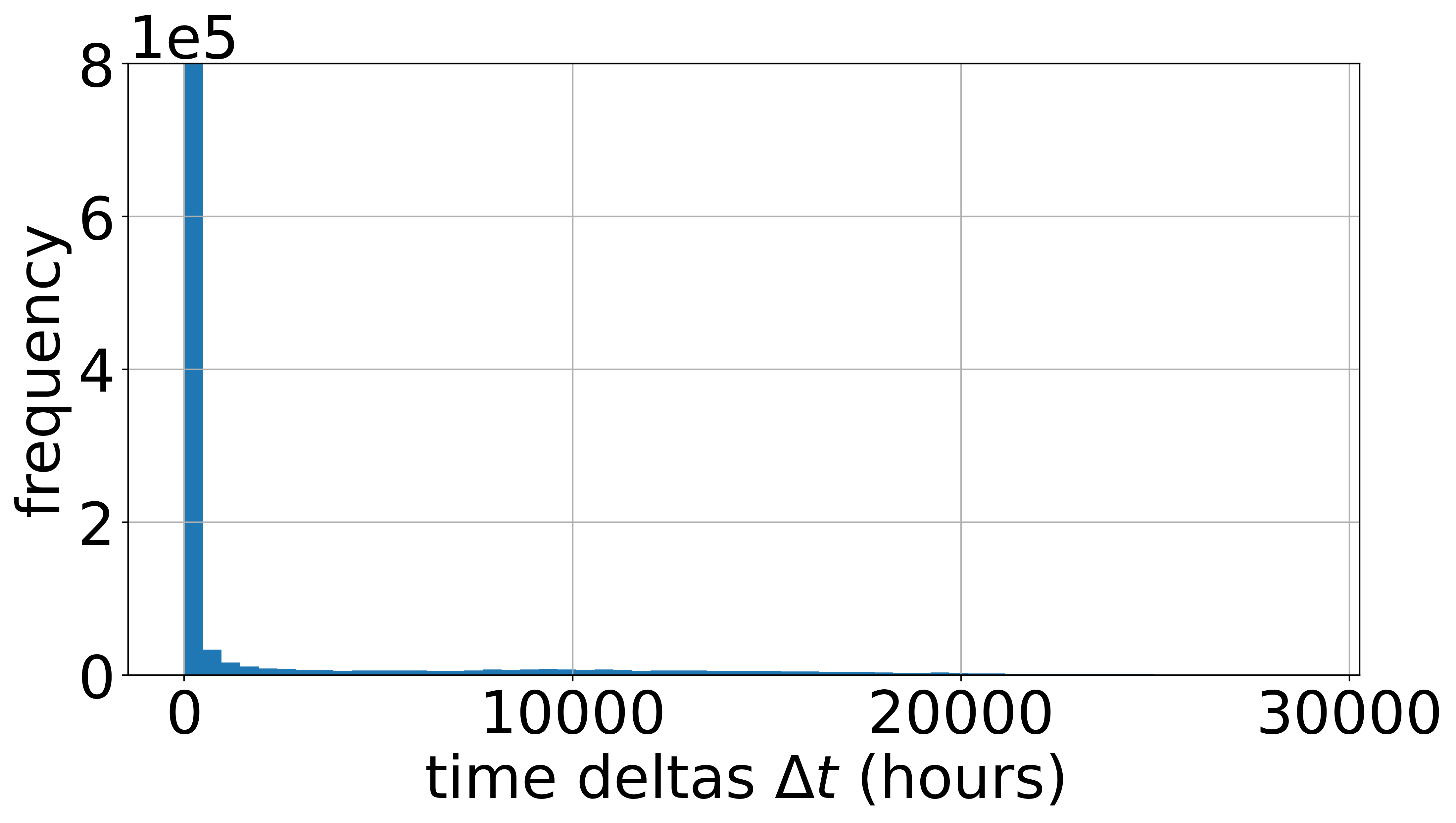}
\end{minipage}
}%
\subfigure[GDELT]{
\begin{minipage}[t]{0.25\linewidth}
\includegraphics[width=0.90\linewidth]{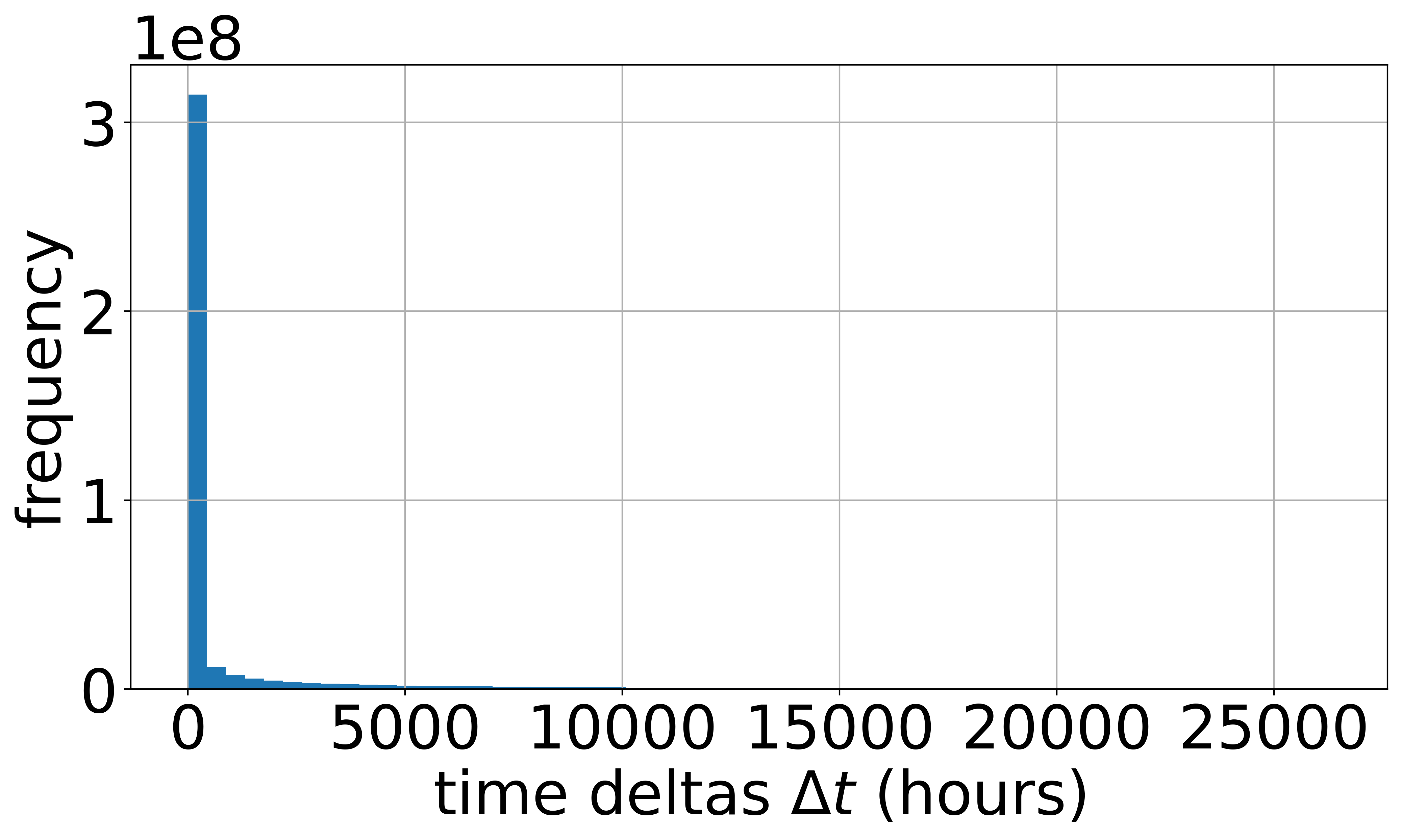}
\end{minipage}
}%
\caption{Distribution of $\Delta t$ on different datasets. The WIKI dataset is presented in Figure~\ref{fig:wiki}}

\label{fig:delta_ts_all}
\end{figure*}

\subsubsection{The distribution of $\Delta t$ on other datasets.} \label{sec:app_delta_t_of_different_data}
We introduce $\Delta t$ as the duration since a node $v$'s memory was last updated, which differs from the $\Delta t$ in the MTGNN inference system~\cite{wang2023tgopt, Zhou2022ModelArchitectureCF}. The $\Delta t$ defined in TGOpt~\cite{wang2023tgopt} and Zhou {\em et al.}~\cite{Zhou2022ModelArchitectureCF} are designed for the time-encoder, which is computed by the difference between current events' timestamp and their historical events' timestamps with their neighbors. We further post the distribution of $\Delta t$ of the remaining datasets in Figure~\ref{fig:delta_ts_all} and observed that the $\Delta t$ in all datasets follow the power-law distribution, indicating that most $\Delta t$ values are small and that most node memories are not stale or constant. This observation provides insights into the occurrence patterns of nodes in different dynamic graphs. Our similarity-based staleness mitigation mechanism focuses on compensating for memory vectors with stale $\Delta t$ values in the long tail of the
distributions.

\subsubsection{Analysis of the node memory similarity.} \label{sec:app_node_memory_similar}
\begin{figure}
\vspace{-2mm}
\begin{minipage}{.55\linewidth} 
\centering
    \includegraphics[width=0.9\linewidth]{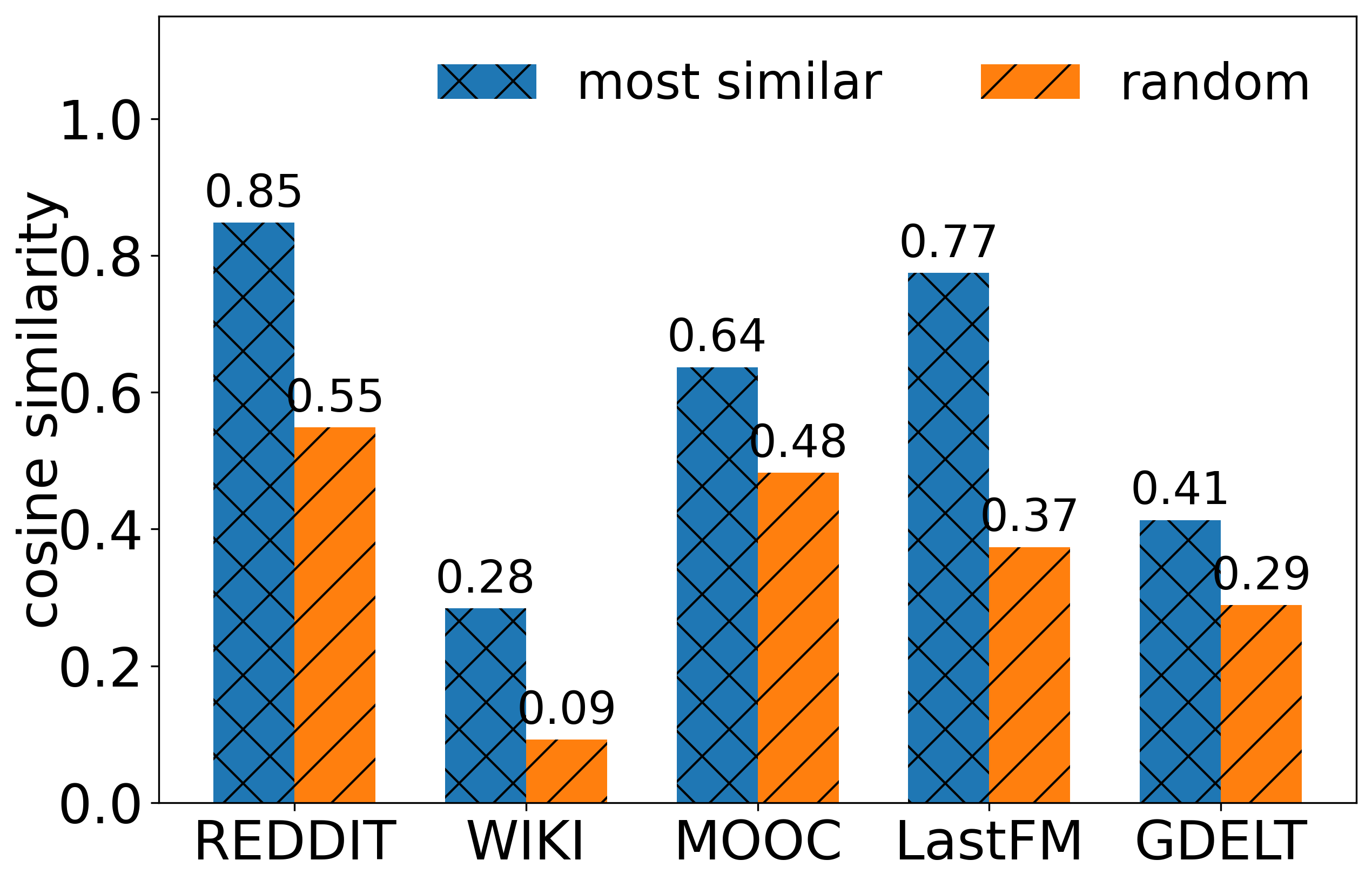}
    \vspace{-3mm}
    \caption{The cosine similarity of the memory vectors between the target nodes (with staled node memory) and their most similar nodes or random nodes}
    \label{fig:mem_sim}
\end{minipage}%
\hspace{2pt}
\begin{minipage}{.43\linewidth} 
\centering
    \includegraphics[width=0.95\linewidth]{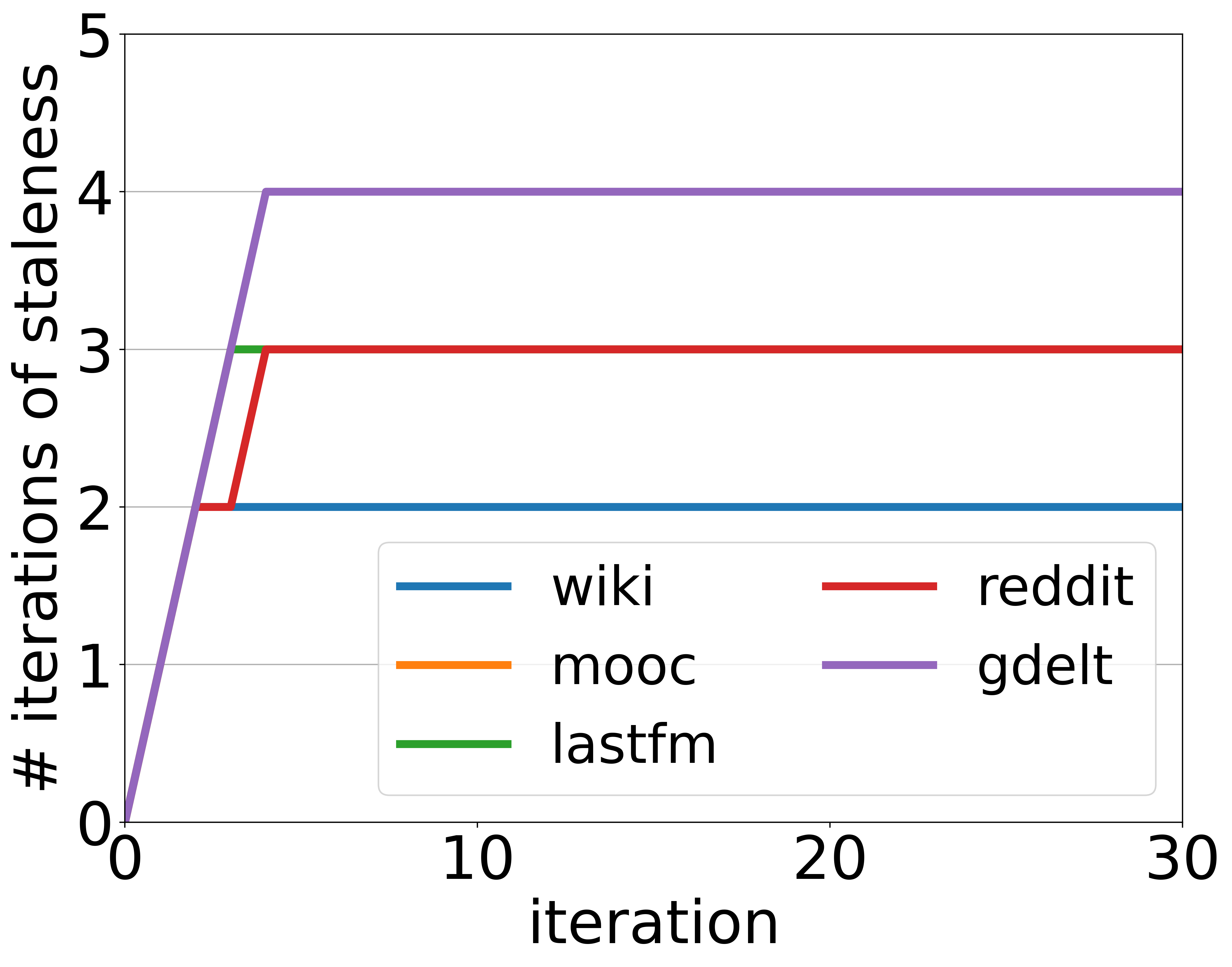}
    \caption{The minimal number of staleness $k_i$ in different iteration $i$}
    \label{fig:stale_count}
\end{minipage}
\vspace{-3mm}
\end{figure}

We compensate the stale node memory by finding their most similar and recently active nodes with the intuition that similar nodes have resembling representations that facilitate the stale node to obtain more updated information. The most similar nodes are computed by counting their common neighbors to get Jaccard similarity. As illustrated in Figure~\ref{fig:mem_sim}, our mechanism for identifying the most recent similar nodes can locate those with representations that are not only similar but also more recently updated than randomly selected nodes. We use cosine similarity as the evaluation metric for similarity.

\subsection{The variance of $k_i$ with respect to $i$}
We further evaluate the variance of $k_i$ when the $i$ changes. As shown in Figure~\ref{fig:stale_count}, the number of staleness $k_i$ will soon converge to a steadily minimal staleness value. This is because of the periodic manner of the MTGNN training as the computation time of different training stages is quite steady.

\renewcommand\arraystretch{0.86}
\begin{table*}[]
\caption{Batch size sensitive analysis. The best results are in \textbf{bold}, and the second-best are \underline{underlined}.}
\vspace{-2mm}
\begin{tabular}{clllllllllll}
\toprule
\multirow{2}{*}{\textbf{Batch size}} & \multicolumn{1}{c}{\multirow{2}{*}{\textbf{Scheme}}} & \multicolumn{2}{c}{\textbf{REDDIT}} & \multicolumn{2}{c}{\textbf{WIKI}} & \multicolumn{2}{c}{\textbf{MOOC}} & \multicolumn{2}{c}{\textbf{LASTFM}} & \multicolumn{2}{c}{\textbf{GDELT}} \\
 & \multicolumn{1}{c}{} & \multicolumn{1}{c}{\textbf{AP(\%)}} & \multicolumn{1}{c}{\textbf{Speedup}} & \multicolumn{1}{c}{\textbf{AP(\%)}} & \multicolumn{1}{c}{\textbf{Speedup}} & \multicolumn{1}{c}{\textbf{AP(\%)}} & \multicolumn{1}{c}{\textbf{Speedup}} & \multicolumn{1}{c}{\textbf{AP(\%)}} & \multicolumn{1}{c}{\textbf{Speedup}} & \multicolumn{1}{c}{\textbf{AP(\%)}} & \multicolumn{1}{c}{\textbf{Speedup}} \\
 \midrule
\multirow{5}{*}{\makecell{Batch 300\\(2000 for GDELT)}} & TGL & \textbf{99.8} & 1$\times$ & \textbf{99.5} & 1$\times$ & \textbf{99.4} & 1$\times$ & \textbf{88.1} & 1$\times$ & \textbf{98.5} & 1$\times$ \\
 & Presample & \textbf{99.8} & 1.26$\times$ & \textbf{99.5} & 1.08$\times$ & \textbf{99.4} & 1.04$\times$ & {\ul 88.0} & 1.51$\times$ & \textbf{98.5 }& 1.12$\times$ \\
 & MSPipe & \textbf{99.8} & \textbf{1.73$\times$} & 99.4 & \textbf{1.67$\times$} & \textbf{99.4} & \textbf{1.47$\times$} & 87.2 & \textbf{1.90$\times$} & 98.2 & \textbf{1.93$\times$} \\
 & MSPipe-S & \textbf{99.8} & {\ul 1.68$\times$} & \textbf{99.5} & {\ul 1.65$\times$} & \textbf{99.4} & {\ul 1.45$\times$} & {\ul 88.0} & {\ul 1.86$\times$} & \textbf{98.5} & {\ul 1.88$\times$} \\ \hline
\multirow{5}{*}{\makecell{Batch 900\\(6000 for GDELT)}} & TGL & \textbf{99.8} & 1$\times$ & \textbf{98.9} & 1$\times$ & \textbf{98.6} & 1$\times$ & {\ul 86.9} & 1$\times$ & {\ul 97.8} & 1$\times$ \\
 & Presample & \textbf{99.8} & 1.10$\times$ & \textbf{98.9} & 1.12$\times$ & \textbf{98.6} & 1.10$\times$ & {\ul 86.9} & 1.37$\times$ & {\ul 97.8} & 1.26$\times$ \\
 & MSPipe & \textbf{99.8} & \textbf{1.62$\times$} & 98.5 & \textbf{1.49$\times$} & \textbf{98.6} & \textbf{1.58$\times$} & 86.7 & \textbf{1.87$\times$} & 97.7 & \textbf{2.01$\times$} \\
 & MSPipe-S & \textbf{99.8} & {\ul 1.56$\times$} & \textbf{98.9} & {\ul 1.46$\times$} & \textbf{98.6} & {\ul 1.53$\times$} & \textbf{87.8} & {\ul 1.80$\times$} & \textbf{98.2} & {\ul 1.93$\times$} \\
 \midrule
\multirow{5}{*}{\makecell{Batch 1200\\(8000 for GDELT)}} & TGL & \textbf{99.8} & 1$\times$ & \textbf{98.5} & 1$\times$ & {\ul 98.3} & 1$\times$ & {\ul 85.8} & 1$\times$ & {\ul 97.1} & 1$\times$ \\
 & Presample & \textbf{99.8} & 1.34$\times$ & \textbf{98.5} & 1.37$\times$ & {\ul 98.3} & 1.32$\times$ & {\ul 85.8} & 1.56 & {\ul 97.1} & 1.28$\times$ \\
 & MSPipe & \textbf{99.8} & \textbf{1.64$\times$} & \textbf{98.5} & \textbf{1.48$\times$} & {\ul 98.3} & \textbf{1.69$\times$} & {\ul 85.8} & \textbf{1.92$\times$} & {\ul 97.1} & \textbf{1.99$\times$} \\
 & MSPipe-S & \textbf{99.8} & {\ul 1.59$\times$} & \textbf{98.5} & {\ul 1.45$\times$} & \textbf{98.8} & {\ul 1.62$\times$} & \textbf{86.2} & {\ul 1.84$\times$} & \textbf{98.1} & {\ul 1.90$\times$} \\
 \midrule  
\multirow{5}{*}{Batch 1600} & TGL & \textbf{99.8} & 1$\times$ & \textbf{98.4} & 1$\times$ & {\ul 97.9} & 1$\times$ & \textbf{84.4} & 1$\times$ &  &  \\
 & Presample & \textbf{99.8} & 1.38$\times$ & \textbf{98.4} & 1.39$\times$ & {\ul 97.9} & 1.33$\times$ & \textbf{84.4} & 1.51$\times$ &  &  \\
 & MSPipe & \textbf{99.8} & \textbf{1.66$\times$} & 98.1 & \textbf{1.58$\times$} & {\ul 97.9} & \textbf{1.71$\times$} & 82.7 & \textbf{1.97$\times$} &  &  \\
 & MSPipe-S & \textbf{99.8} & {\ul 1.58$\times$} & {\ul 98.3} & {\ul 1.53$\times$} & \textbf{98.7} & {\ul 1.64$\times$} & {\ul 84.2} & {\ul 1.88$\times$} &  &  \\
 \bottomrule
\end{tabular}
\label{tab:batchsize_sensitive}
\vspace{-3mm}
\end{table*}
\subsection{Batch size sensitivity analysis} \label{sec:app_batch_size_sens}
To further validate the effectiveness of MSPipe in different batch sizes, we conducted batch size sensitivity evaluations using the following local batch sizes: 300, 900, 1200, and 1600 for the small datasets, and 2000, 6000, and 8000 for the large dataset (used 600 and 4000 in the original experiments), illustrated in Table~\ref{tab:batchsize_sensitive}.

As demonstrated in Table~\ref{tab:batchsize_sensitive}, MSPipe consistently outperforms all baseline methods in varying batch sizes, achieving up to 2.01$\times$ speedup without compromising model accuracy. These results further validate the practicality of MSPipe. It is worth noting that for the same dataset, MSPipe tends to exhibit similar speedup among various batch sizes, indicating no direct correlation between batch size and speedup. 

\renewcommand\arraystretch{0.6}
\begin{table*}[t]
\caption{MSPipe compares with baseline methods using larger batch size.}
\begin{tabular}{lllllllllllll}
\toprule
\multicolumn{1}{c}{\multirow{2}{*}{\textbf{Scheme}}} & \multicolumn{3}{c}{\textbf{REDDIT}} & \multicolumn{3}{c}{\textbf{WIKI}} & \multicolumn{3}{c}{\textbf{MOOC}} & \multicolumn{3}{c}{\textbf{LastFM}} \\
\multicolumn{1}{c}{} & AP(\%) & Time(s) & Speedup & AP(\%) & Time(s) & Speedup & AP(\%) & Time(s) & Speedup & AP(\%) & Time(s) & Speedup \\
\midrule
\makecell{TGL\\batch 600}& 99.8 & 7.31 & 1$\times$ & \textbf{99.4} & 2.41 & 1$\times$ & \textbf{99.4} & 4.31 & 1$\times$ & \textbf{87.2 }& 13.10 & 1$\times$ \\
\midrule
\makecell{MSPipe\\batch 600} & 99.8 & \textbf{4.14} & \textbf{1.77$\times$} & \underline{99.1} & \textbf{1.57} & \textbf{1.54$\times$} & \underline{99.3} & \textbf{2.88} & \textbf{1.50$\times$} & \underline{86.9} & \textbf{6.55} & \textbf{1.87$\times$} \\
\midrule
\makecell{TGL\\batch 900} & 99.8 & 5.22 & 1.40$\times$ & 98.9 & 2.03 & 1.19$\times$ & 98.7 & 3.18 & 1.36$\times$ & 86.9 & 10.10 & 1.30$\times$ \\
\midrule
\makecell{TGL\\batch 1200} & 99.8 & \underline{4.48} & \underline{1.63$\times$} & 98.5 & \underline{1.83} & \underline{1.32$\times$} & 98.3 & \underline{2.99} & \underline{1.44$\times$} & 85.8 & \underline{8.43} & \underline{1.55$\times$} \\
\bottomrule
\end{tabular}
\label{tab:app_compare_with_large_batch}
\end{table*}

\subsection{Compare with Strawman method: increase batch size}
We conducted additional empirical comparisons between MSPipe and baseline methods using larger batch sizes. In Table~\ref{tab:app_compare_with_large_batch}, MSPipe consistently outperforms baseline methods with batch sizes increased by 1.5$\times$ and 2$\times$, achieving speedups of up to 57\% and 32\% respectively. While the TGN model experiences up to 1.4\% accuracy loss with larger batch sizes, MSPipe maintains high accuracy with a maximum accuracy loss of 0.3\%. It is worth emphasizing that MSPipe can be applied with larger batch sizes to further boost training throughput as shown in Table~\ref{tab:batchsize_sensitive}.

\subsection{Memory overhead analysis.} \label{sec:app_mem}

In MSPipe, we introduce staleness within the memory module to facilitate the pre-fetching of features and memory in subsequent iterations. However, unlike other asynchronous training frameworks~\cite{chen2022sapipe,li2018pipe, wan2022pipegcn, peng2022sancus}, where staleness is introduced during DNN or GNN parameter learning, our MTGNN training stage does not incorporate staleness. Each subgraph is executed sequentially, resulting in no additional hidden states during MTGNN computation.

Consequently, the additional memory consumption in MSPipe arises from the prefetched subgraph, which includes node/edge features and memory vectors. We can compute an upper bound for this memory consumption as follows:

Let the subgraph in each iteration have a batch size of $B$, node feature dimension of $H_n$, edge feature dimensions of $H_e$, node memory dimension of $M$, and an introduced staleness bound of $K$. During subgraph sampling, we use the maximum neighbor size of $\mathcal{N}$ (e.g. 10) to compute the memory consumption, which represents an upper bound. Within each subgraph, we have three nodes per sample, comprising a source node, destination node, and neg\_sample node. Hence, a single subgraph contains a total of $3B(\mathcal{N}+1)$ nodes, where $\mathcal{N}+1$ denotes the number of neighbors per node, inclusive of the target node itself. Additionally, each graph event involves both positive and negative links, resulting in two edges per event. Consequently, the total number of links per subgraph amounts to $2B(\mathcal{N}+1)$. The memory utilization of a subgraph encompasses node IDs, edge IDs, node features, edge features, and node memory states. With the introduction of a staleness bound of $K$, the GPU accommodates a maximum of $K$ additional subgraphs. Assuming a data format of Float32 (i.e., 4 bytes), the additional memory consumption for these subgraphs can be formulated as:
\renewcommand\arraystretch{1}
\begin{table}[t]
\caption{Additional memory overhead of JODIE when applying stalenes.}
\resizebox{1\linewidth}{!}{
\begin{tabular}{llllll}
\toprule
Overhead\textbackslash Dataset & REDDIT & WIKI & MOOC & LastFM & GDELT \\
\midrule
Addition & 54.6MB & 42.6MB & 43.9MB & 47.7MB & 0.98GB \\
Upperbound & 51.4MB & 34.3MB & 44.3MB & 44.3MB & 1.35GB \\
GPU Mem (40GB) portion & 0.14\% & 0.11\% & 0.11\% & 0.12\% & 2.45\% \\
\bottomrule
\end{tabular}}
\label{tab:mem_jodie}
\vspace{-3mm}
\end{table}

\begin{align*}
&4 \times K \times [3B(\mathcal{N}+1)H_n + 2B(\mathcal{N}+1)H_e \\
& \qquad \quad + 3B(\mathcal{N}+1)M + 3B(\mathcal{N}+1) + 2B(\mathcal{N}+1)] \\
=\ & 4 \times K \times 3B(\mathcal{N}+1)(H_n + \frac{2}{3}H_e + M + \frac{5}{3}) \\
=\ & 12KB(\mathcal{N}+1)(H_n + \frac{2}{3}H_e + M + \frac{5}{3})
\end{align*}
Moreover, we conduct empirical experiments on all the models/datasets with the \textit{torch.cuda.memory\_summary()} API. As observed in Table\ref{tab:mem_gpu},\ref{tab:mem_jodie} and\ref{tab:mem_apan}, the additional memory usage from MSPipe strictly resembles to our analyzed upper bound. Additionally, we compare the additional memory cost with the GPU memory size, demonstrating that the additional memory overhead is a relatively small proportion (up to 3.20\% for APAN) of the modern GPU's capacity.

\renewcommand\arraystretch{1}
\begin{table}[t]
\caption{Additional memory overhead of APAN when applying staleness.}
\resizebox{1\linewidth}{!}{
\begin{tabular}{llllll}
\toprule
Overhead\textbackslash Dataset & REDDIT & WIKI & MOOC & LastFM & GDELT \\
\midrule
Addition & 50.1MB & 32.7MB & 43.5MB & 55.2MB & 1.28GB \\
Upperbound & 51.4MB & 34.3MB & 44.3MB & 44.3MB & 1.35GB \\
GPU Mem (40GB) portion & 0.12\% & 0.08\% & 0.11\% & 0.14\% & 3.20\% \\
\bottomrule
\end{tabular}}
\label{tab:mem_apan}
\vspace{-3mm}
\end{table}

\end{document}